\documentclass[10pt,journal,compsoc]{IEEEtran}
\ifCLASSOPTIONcompsoc
  \usepackage[nocompress]{cite}
\else
  \usepackage{cite}
\fi

\ifCLASSINFOpdf
\else
\fi

\usepackage{amsmath}
\usepackage{amssymb}
\usepackage{graphicx}
\usepackage{subfigure}
\usepackage{booktabs}
\usepackage{algorithmic}
\usepackage{array}
\usepackage{stfloats}
\usepackage{url}
\usepackage{multirow, colortbl, xcolor, booktabs}
\usepackage{overpic}
\usepackage{amsfonts}
\usepackage{bm}
\usepackage{relsize}
\usepackage{multirow}
\usepackage{threeparttable}
\usepackage{makecell}
\usepackage{bbm}
\usepackage{soul}
\definecolor{airforceblue}{rgb}{0.36, 0.54, 0.66}
\definecolor{skyblue}{rgb}{0.53, 0.81, 0.92}
\definecolor{frenchblue}{rgb}{0.0, 0.45, 0.73}

\definecolor{americanrose}{rgb}{1.0, 0.01, 0.24}
\definecolor{myred}{rgb}{0.753, 0.314, 0.275}
\definecolor{myblue}{rgb}{0.0, 0.24, 0.95}
\definecolor{tbl_gray}{gray}{0.85}
\newcommand{\RNum}[1]{\uppercase\expandafter{\romannumeral #1\relax}}

\newcommand\MYhyperrefoptions{bookmarks=true,bookmarksnumbered=true,
pdfpagemode={UseOutlines},plainpages=false,pdfpagelabels=true,
colorlinks=true,linkcolor={americanrose},citecolor={myblue},urlcolor={myblue}}
\usepackage[\MYhyperrefoptions,pdftex]{hyperref}

\usepackage[capitalize]{cleveref}

\newlength\savedwidth

\begin{document}
\title{A Survey on Label-efficient Deep Image Segmentation: Bridging the Gap between Weak Supervision and Dense Prediction}
%
\author{Wei Shen, Zelin Peng, Xuehui Wang, Huayu Wang, Jiazhong Cen, Dongsheng Jiang, Lingxi Xie, \\Xiaokang Yang,~\IEEEmembership{Fellow,~IEEE,} and Qi Tian,~\IEEEmembership{Fellow,~IEEE}
\IEEEcompsocitemizethanks{\IEEEcompsocthanksitem W. Shen, Z. Peng, X. Wang, H. Wang, J. Cen and X. Yang are with MoE Key Lab of Artificial Intelligence, AI Institute, Shanghai Jiao Tong University, Shanghai, 200240, China.\protect\\
E-mail: \{wei.shen,xkyang\}@sjtu.edu.cn.
\IEEEcompsocthanksitem D. Jiang, L. Xie and Q. Tian are with Huawei Inc., China. (Corresponding author: Q. Tian.)\protect\\
E-mail: \{jiangdongsheng1,xielingxi,tian.qi1\}@huawei.com.
}%
\thanks{Manuscript received April 19, 2005; revised August 26, 2015.}}

\markboth{A SUBMISSION TO IEEE TRANSACTION ON PATTERN ANALYSIS AND MACHINE INTELLIGENCE}%
{Shell \MakeLowercase{\textit{et al.}}: Bare Demo of IEEEtran.cls for Computer Society Journals}

\IEEEtitleabstractindextext{%
\begin{abstract}
The rapid development of deep learning has made a great progress in image segmentation, one of the fundamental tasks of computer vision. However, the current segmentation algorithms mostly rely on the availability of pixel-level annotations, which are often expensive, tedious, and laborious. To alleviate this burden, the past years have witnessed an increasing attention in building label-efficient, deep-learning-based image segmentation algorithms. This paper offers a comprehensive review on label-efficient image segmentation methods. To this end, we first develop a taxonomy to organize these methods according to the supervision provided by different types of weak labels (including no supervision, inexact supervision, incomplete supervision and inaccurate supervision) and supplemented by the types of segmentation problems (including semantic segmentation, instance segmentation and panoptic segmentation). Next, we summarize the existing label-efficient image segmentation methods from a unified perspective that discusses an important question: how to bridge the gap between weak supervision and dense prediction -- the current methods are mostly based on heuristic priors, such as cross-pixel similarity, cross-label constraint, cross-view consistency, and cross-image relation. Finally, we share our opinions about the future research directions for label-efficient deep image segmentation.
\end{abstract}

}

\maketitle

\IEEEdisplaynontitleabstractindextext

\IEEEpeerreviewmaketitle


\IEEEraisesectionheading{
\vspace{-2mm}
\section{Introduction}\label{sec:introduction}}
\vspace{-2mm}
\IEEEPARstart{S}{egmentation} is one of the oldest and most widely studied tasks in computer vision. Its goal is to produce a dense prediction for a given image, \emph{i.e.}, assigning each pixel a pre-defined class label (semantic segmentation)~\cite{YYDS_Grabcut_TOG_2004,YYDS_DenseCRF_NeurIPS_2011} or associating each pixel to an object instance (instance segmentation)~\cite{YYDS_SDS_ECCV_2014}, or the combintation of both (panoptic segmentation)~\cite{YYDS_Panoptic_CVPR_2019}, which enables grouping semantically-similar pixels into high-level meaningful concepts, such as objects (person, cat, ball, \emph{etc}), and stuff (road, sky, water, \emph{etc}). 

The last decade has witnessed a tremendous success in segmentation~\cite{YYDS_Unet_MICCAI_2015,YYDS_Deeplabv1_ICLR_2015,YYDS_DilatedConv_ICLR_2016,YYDS_Deeplabv2_TPAMI_2017,deeplabv3,YYDS_Deeplabv3plus_ECCV_2018,YYDS_FCIS_CVPR_2017,YYDS_Maskrcnn_ICCV_2017,YYDS_PSPNet_CVPR_2017,YYDS_PANet_CVPR_2018,YYDS_Segnet_TPAMI_2017,YYDS_Yolact_ICCV_2019,YYDS_DANet_CVPR_2019,YYDS_SOLO_ECCV_2020,YYDS_OCNet_IJCV21} brought by deep convolutional neural networks (CNNs), especially fully convolutional networks (FCNs)~\cite{YYDS_FCN_CVPR_2015}, thanks to their strong dense representation learning ability. However, these deep learning based image segmentation methods thrives with dense labels, \emph{i.e.}, per-pixel annotations, which are expensive and laborious to obtain. 

Given widespread label scarcity in the real world, developing label-efficient deep image segmentation methods, which are based on the supervision from weak labels (weak supervision) to reduce the dependency on dense labels, becomes a growing trend, attracting more and more researchers' attention. As a result, there has been an explosive growth in the number of label-efficient deep image segmentation methods proposed in recent years, which makes it difficult for researchers to keep pace with the new progress. Therefore, a survey on these label-efficient deep image segmentation methods is urgently necessary. However, to the best of our knowledge, there are only a few related survey papers~\cite{Survey_SEMI_Weak_AIR20,DASS_survey}, which merely focus on one particular segmentation task with  the supervision from weak labels of limited types.

\begin{figure*}[!ht]
	\centering
	\begin{overpic}[width=\linewidth]{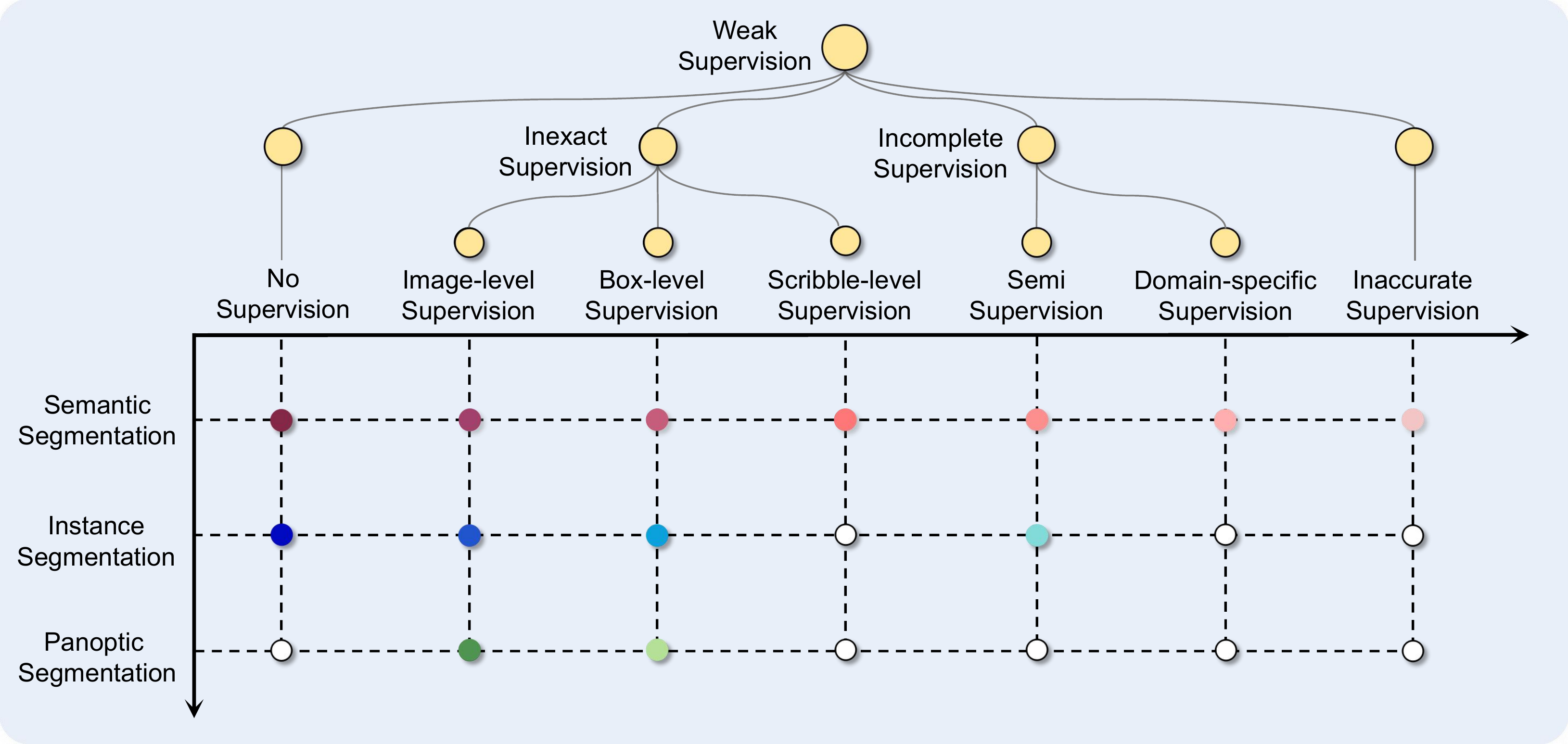}
		\put(19, 22){\tiny{MaskContrast\cite{con_maskcontrast}}}
		\put(19, 24){\tiny{STEGO~\cite{con_stego}}}
        \put(19, 14.4){\tiny{FreeSOLO~\cite{freesolo}}}
		\put(31, 18.7){\tiny{AffinityNet~\cite{WSSS_PSA_2018_CVPR}}}
		\put(31, 16.7){\tiny{SEAM~\cite{WSSS_SEAM_2020_CVPR}}}

		\put(91, 18.7){\tiny{ADELE\cite{WSSS_AEL_2022_CVPR}}}
		
		\put(31, 4){\tiny{JTSM\cite{Shen_WSPS_2021_CVPR}}}
		\put(67, 22){\tiny{PseudoSeg\cite{SSL_PseudoSeg}}}
		\put(67, 24){\tiny{CAC~\cite{SSL_CAC}}}
		
		\put(67, 11.5){\tiny{Mask$^X$ RCNN~\cite{PSIS_MaskXRCNN_CVPR_2018}}}
		\put(67, 9.5){\tiny{Shapeprop~\cite{PSIS_Shapeprop_CVPR_2020}}}

		\put(31, 11.5){\tiny{PRM\cite{WSIS_PRM_CVPR_2018}}}
		\put(31, 9.5){\tiny{Label-PEnet~\cite{WSIS_Labelpenet_ICCV_2019}}}
		
		\put(55, 18.7){\tiny{ScribbleSup\cite{WSSS_ScribbleSup_2016_CVPR}}}
		\put(43, 22){\tiny{BAP~\cite{WSSS_BAP_2021_CVPR}}}
		\put(43, 16.4){\tiny{SDI~\cite{WSIS_SDI_CVPR_2017}}}
		\put(43, 14.4){\tiny{BBTP~\cite{WSIS_BBTP_NIPS_2019}}}
		\put(43, 7.2){\tiny{WPS\cite{Li_WSPS_ECCV_2018}}}
		
		\put(79.3, 22){\tiny{BDL~\cite{UDA_BDL}}}
		\put(79.3, 24){\tiny{ProDA~\cite{UDA_ProDA}}}
	\end{overpic}
        \vspace{-6mm}
	\caption{The taxonomy of label efficient deep image segmentation methods according to the type categorization of weak supervision (upper half) and the type categorization of segmentation problems. The interactions with filled dots and hollow dots indicate the segmentation problems with the certain types of weak supervision have been explored and have not been explored, respectively. For the former, some typical works are provided.}
    \vspace{-2mm}
    \label{fig:taxonomy}
\end{figure*}

This paper aims at providing a comprehensive overview for recent label-efficient deep image segmentation methods. These methods focus on diverse problems. Here, a \emph{problem} is defined as a particular segmentation problem, \emph{i.e.}, semantic segmentation, instance segmentation and panoptic segmentation, with a certain type of weak supervision. To organize such methods for diverse problems, we need to address two questions in this survey: 1) How to build a taxonomy for these methods? 2) How to summarize the strategies used in these methods from a unified perspective? We notice that the types of weak labels are pivotal to determine the strategies to design these label-efficient image segmentation methods. Thus, we try to answer the above two questions from the perspective of weak supervision. Towards this end, we first provide the type categorization of weak supervision, which is hierarchical, as shown in~\cref{fig:taxonomy}. The types of weak supervision include

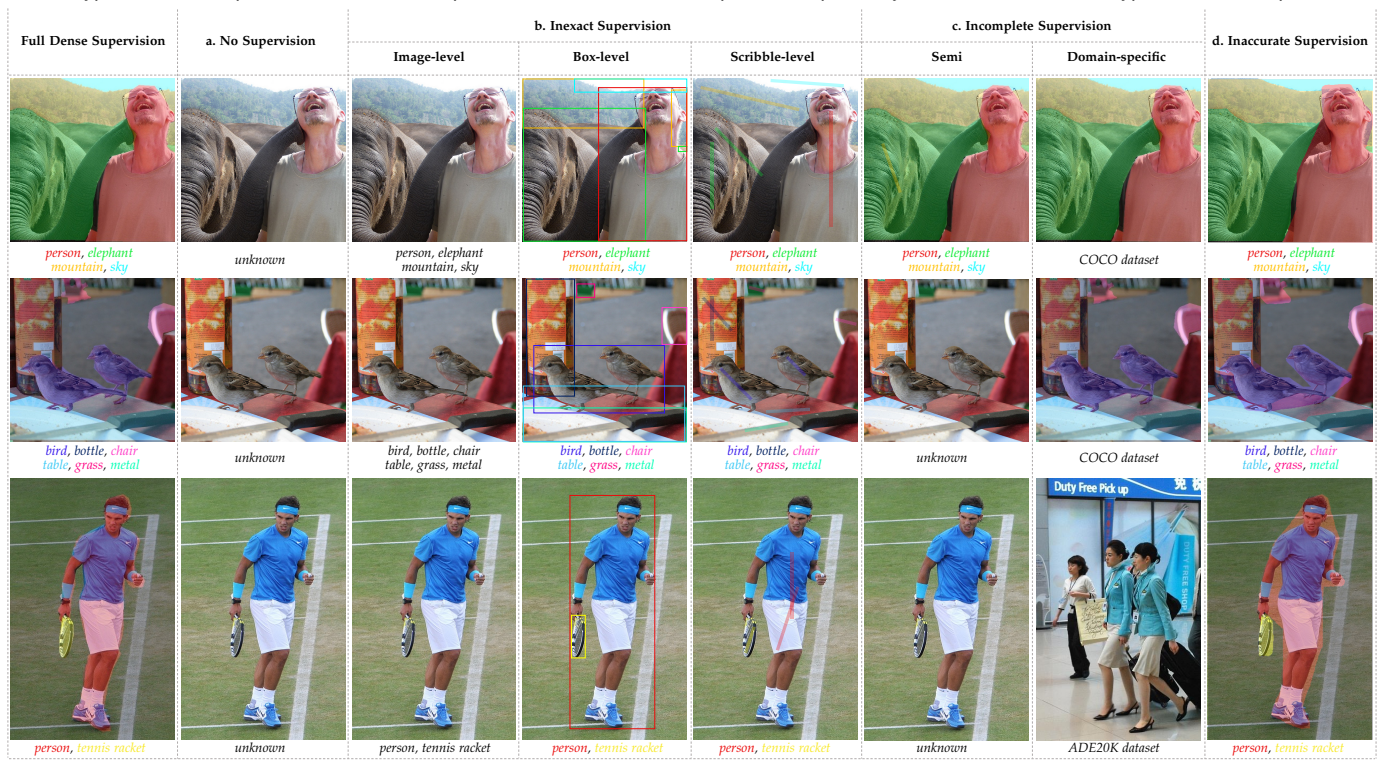
\begin{figure*}[!ht]
	\centering
	\begin{overpic}[width=\linewidth]{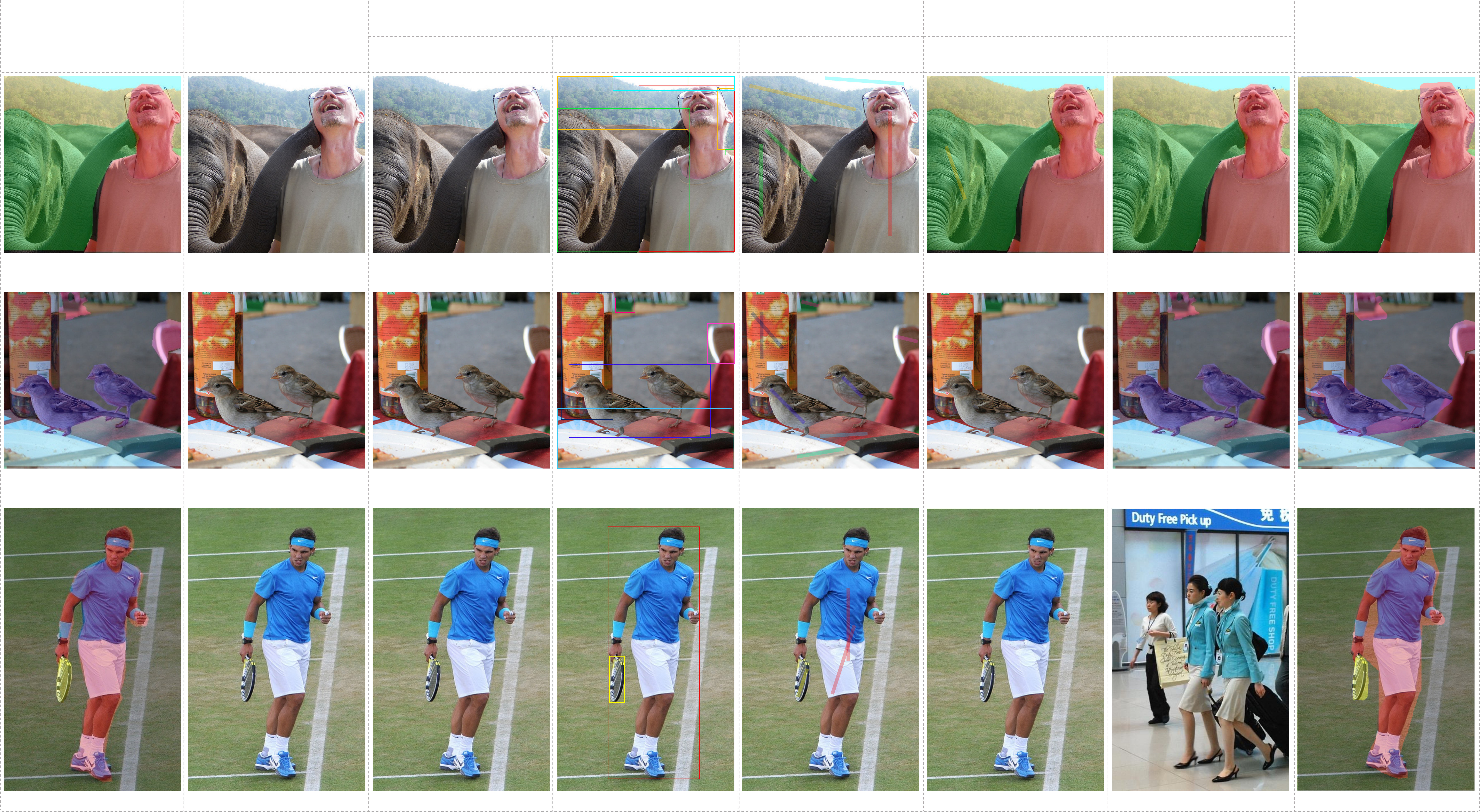}

	    \put(2.8,36.7){\tiny{\emph{\textcolor[RGB]{235,10,10}{person}, \textcolor[RGB]{10,235,50}{elephant}}}} \put(3.2,35.7){\tiny{\emph{\textcolor[RGB]{255,192,0}{mountain}, \textcolor[RGB]{0,251,255}{sky}}}}
        \put(2.8,22.2){\tiny{\emph{\textcolor[RGB]{50,20,235}{bird}, \textcolor[RGB]{0,32,96}{bottle}, \textcolor[RGB]{255,60,200}{chair}}}} \put(2.6,21.2){\tiny{\emph{\textcolor[RGB]{57,200,245}{table}, \textcolor[RGB]{255,0,130}{grass}, \textcolor[RGB]{0,255,181}{metal}}}}
        \put(2,0.5){\tiny{\emph{\textcolor[RGB]{235,10,10}{person}, \textcolor[RGB]{248,233,43}{tennis racket}}}}
        
        \put(16.6,36.2){\tiny{\emph{unknown}}}
        \put(16.6,21.7){\tiny{\emph{unknown}}}
        \put(16.6,0.5){\tiny{\emph{unknown}}}
        
	    \put(28.4,36.7){\tiny{\emph{person, elephant}}} \put(28.8,35.7){\tiny{\emph{mountain, sky}}}
        \put(27.8,22.2){\tiny{\emph{bird, bottle, chair}}} \put(27.6,21.2){\tiny{\emph{table, grass, metal}}}
        \put(27.2,0.5){\tiny{\emph{person, tennis racket}}} 
        
	    \put(40.6,36.7){\tiny{\emph{\textcolor[RGB]{235,10,10}{person}, \textcolor[RGB]{10,235,50}{elephant}}}} \put(41.0,35.7){\tiny{\emph{\textcolor[RGB]{255,192,0}{mountain}, \textcolor[RGB]{0,251,255}{sky}}}}
        \put(40.4,22.2){\tiny{\emph{\textcolor[RGB]{50,20,235}{bird}, \textcolor[RGB]{0,32,96}{bottle}, \textcolor[RGB]{255,60,200}{chair}}}} \put(40.2,21.2){\tiny{\emph{\textcolor[RGB]{57,200,245}{table}, \textcolor[RGB]{255,0,130}{grass}, \textcolor[RGB]{0,255,181}{metal}}}}
        \put(39.8,0.5){\tiny{\emph{\textcolor[RGB]{235,10,10}{person}, \textcolor[RGB]{248,233,43}{tennis racket}}}}
        
	    \put(52.8,36.7){\tiny{\emph{\textcolor[RGB]{235,10,10}{person}, \textcolor[RGB]{10,235,50}{elephant}}}} \put(53.2,35.7){\tiny{\emph{\textcolor[RGB]{255,192,0}{mountain}, \textcolor[RGB]{0,251,255}{sky}}}}
        \put(52.6,22.2){\tiny{\emph{\textcolor[RGB]{50,20,235}{bird}, \textcolor[RGB]{0,32,96}{bottle}, \textcolor[RGB]{255,60,200}{chair}}}} \put(52.4,21.2){\tiny{\emph{\textcolor[RGB]{57,200,245}{table}, \textcolor[RGB]{255,0,130}{grass}, \textcolor[RGB]{0,255,181}{metal}}}}
        \put(52.0,0.5){\tiny{\emph{\textcolor[RGB]{235,10,10}{person}, \textcolor[RGB]{248,233,43}{tennis racket}}}}
        
	    \put(65.4,36.7){\tiny{\emph{\textcolor[RGB]{235,10,10}{person}, \textcolor[RGB]{10,235,50}{elephant}}}} \put(65.8,35.7){\tiny{\emph{\textcolor[RGB]{255,192,0}{mountain}, \textcolor[RGB]{0,251,255}{sky}}}}
	    \put(66.4,21.7){\tiny{\emph{unknown}}}
        \put(66.4,0.5){\tiny{\emph{unknown}}}

        \put(78.3,36.2){\tiny{\emph{COCO dataset}}}
        \put(78.3,21.7){\tiny{\emph{COCO dataset}}}
        \put(77.5,0.5){\tiny{\emph{ADE20K dataset}}}
        
	    \put(90.6,36.7){\tiny{\emph{\textcolor[RGB]{235,10,10}{person}, \textcolor[RGB]{10,235,50}{elephant}}}} \put(91.0,35.7){\tiny{\emph{\textcolor[RGB]{255,192,0}{mountain}, \textcolor[RGB]{0,251,255}{sky}}}}
        \put(90.4,22.2){\tiny{\emph{\textcolor[RGB]{50,20,235}{bird}, \textcolor[RGB]{0,32,96}{bottle}, \textcolor[RGB]{255,60,200}{chair}}}} \put(90.2,21.2){\tiny{\emph{\textcolor[RGB]{57,200,245}{table}, \textcolor[RGB]{255,0,130}{grass}, \textcolor[RGB]{0,255,181}{metal}}}}
        \put(89.5,0.5){\tiny{\emph{\textcolor[RGB]{235,10,10}{person}, \textcolor[RGB]{248,233,43}{tennis racket}}}}
        
        \put(1,52.2){\tiny{\textbf{Full Dense Supervision\label{fig:weak_label_dense}}}}
        \put(14.5,52.2){\tiny{\textbf{a. No Supervision\label{fig:weak_label_no}}}}
        \put(38.5,53.3){\tiny{\textbf{b. Inexact Supervision}\label{fig:weak_label_coarse}}}
        \put(28.2,51.0){\tiny{\textbf{Image-level}}}
        \put(41.3,51.0){\tiny{\textbf{Box-level}}}
        \put(52.8,51.0){\tiny{\textbf{Scribble-level}}}
        \put(69,53.3){\tiny{\textbf{c. Incomplete Supervision\label{fig:weak_label_incomplete}}}}
        \put(67.5,51){\tiny{\textbf{Semi}}}
        \put(77.4,51){\tiny{\textbf{Domain-specific}}}
        \put(88.0,52.2){\tiny{\textbf{d. Inaccurate Supervision}\label{fig:weak_label_noisy}}}
	\end{overpic}
        \vspace{-6mm}
	\caption{Examples for each type of weak supervision compared with the full dense supervision. }
        \vspace{-5mm}
        \label{fig:weak_label}
\end{figure*}
\vspace{-3mm}

\vspace{-1mm}
\begin{enumerate}
    \item No supervision: No annotations are provided for any of training images (Fig.~\ref{fig:weak_label} (a));
    \item Inexact supervision: Annotations are provided for all training images, but the annotation for each image is not as exact as desired, which does not fully cover all pixels' labels (Fig.~\ref{fig:weak_label} (b)). Inexact supervision can be categorized into (i) image-level supervision, (ii) box-level supervision and (iii) scribble-level supervision;
    \item Incomplete supervision: Full per-pixel annotations are provided for only a subset of training images (Fig.~\ref{fig:weak_label} (c)) . Incomplete supervision can be categorized into (i) semi supervision, if the rest training images are not fully annotated or not annotated and (ii) domain-specific supervision, if the rest training images are from a different domain;
    \item Inaccurate supervision: Per-pixel annotations are provided for all training images, but there are annotation errors, \emph{i.e.}, noisy annotations. (Fig.~\ref{fig:weak_label} (d)).
\end{enumerate}
\vspace{-2mm}

With this hierarchical type categorization of weak supervision, we can build a taxonomy for label-efficient deep image segmentation methods. As shown in~\cref{fig:taxonomy}, this taxonomy is built mainly according to the types of weak supervision supplemented by the types of segmentation problems: The horizontal and vertical axes show different types of weak supervision and segmentation tasks, respectively; Each intersection indicates the problem of the corresponding segmentation task with the corresponding weak supervision, where the interactions with filled dots and hollow dots indicate the problems have been explored and have not been explored, respectively; For each intersection with colored filled dots, \emph{i.e.}, a problem has been explored, some representative works are given.

\begin{table*}[t]
	\centering
	\renewcommand\arraystretch{1.0}
	\caption{Representative works of label-efficient deep image segmentation.}
        \vspace{-3mm}
	\label{tab:overview}
	\footnotesize
	\setlength{\tabcolsep}{3.5pt}{
	\resizebox{\textwidth}{!}{
		\begin{tabular}{c|c|c|c|c}
			\toprule
			\textbf{Supervision} & \textbf{Task} & \textbf{Problem} & \textbf{Method} & \textbf{Strategy to bridge the supervision gap} \\
			\midrule\midrule
			\multirow{5}{*}{\makecell[c]{No \\ Supervision}} &
			\multirow{4}{*}{\makecell[c]{Semantic \\ segmentation}} &  \multirow{4}{*}{\makecell[c]{Unsupervised \\ semantic segmentation}} & SegSort~\cite{clu_segsort}   & Prototype learning according to \textbf{cross-pixel similarity} \\
			& & & MaskContrast~\cite{con_maskcontrast}   & Pixel-wise contrastive learning to keep \textbf{cross-view consistency}\\
			& & & IIC~\cite{clu_iic}   & Mutual information maximization to keep \textbf{cross-view consistency} through a Siamese structure\\
			& & & STEGO~\cite{con_stego}  & Feature correspondence distillation among image collections to mine \textbf{cross-image relation}\\
            \cline{2-5}
            & \makecell[c]{Instance \\ segmentation} & \makecell[c]{Unsupervised \\ instance segmentation} & FreeSOLO~\cite{freesolo} & Pseudo mask generation from learned dense features according to \textbf{cross-pixel similarity} \\
            \hline
			\multirow{20}{*}{\makecell[c]{Inexact \\ Supervision}} &
			\multirow{10}{*}{\makecell[c]{Semantic \\ segmentation}} &
			\multirow{6}{*}{\makecell[c]{Semantic segmentation \\with image-level supervision}} & MDC~\cite{WSSS_RDC_2018_CVPR} & Seed area expanding by ensemble according to \textbf{cross-label constraint}   \\
			& & & SeeNet~\cite{WSSS_SE_2018_NEURIPS}& Seed area refinement guided by saliency maps according to \textbf{cross-pixel similarity}  \\
			& & & SEAM~\cite{WSSS_SEAM_2020_CVPR}   & Seed area refinement by enforcing \textbf{cross-view consistency} under affine transformations   \\
			& & & GWSM~\cite{WSSS_Group_2021_AAAI}   & Seed area refinement by capturing \textbf{cross-image relation} using a graph neural network\\
			& & & AffinityNet~\cite{WSSS_PSA_2018_CVPR} & Pseudo mask generation by semantic information propagation according to \textbf{cross-pixel similarity} \\
			& & & CIAN~\cite{WSSS_CIAN_2020_AAAI}   & Pseudo mask generation by feature learning adapted with \textbf{cross-image relation}  \\
            \cline{3-5}
            
		    & & \makecell[c]{Semantic segmentation \\with box-level supervision}& BAP\cite{WSSS_BAP_2021_CVPR}& Background removal within boxes based on \textbf{cross-pixel similarity} with pixels outside boxes\\ 
			\cline{3-5}

			& & \makecell[c]{Semantic segmentation \\with scribble-level supervision}& ScribbleSup\cite{WSSS_ScribbleSup_2016_CVPR} & Graph-based label propagation according to \textbf{cross-pixel similarity}  \\
			\cline{2-5}
			
			&
			\multirow{5}{*}{\makecell[c]{Instance \\ segmentation}} &
            \multirow{3}{*}{\makecell[c]{Instance segmentation \\ with image-level supervision}}& PRM~\cite{WSIS_PRM_CVPR_2018} & Instance-level seed area generation according to  \textbf{cross-label constraint}   \\
            &  & &  IRNet~\cite{WSIS_IRNet_CVPR_2019} & Self-training based instance-level pseudo mask generation by pairwise affinity according to \textbf{cross-label constraint} \\
			&  & &  Label-PEnet~\cite{WSIS_Labelpenet_ICCV_2019} & End-to-end instance-level pseudo mask generation according to \textbf{cross-label constraint} \\
			\cline{3-5}
			
			&  &\multirow{2}{*}{\makecell[c]{Instance segmentation \\ with box-level supervision}}& SDI~\cite{WSIS_SDI_CVPR_2017} & Self-training based pseudo mask generation from given boxes according to  \textbf{cross-label constraint}  \\
			&  & & BBTP~\cite{WSIS_BBTP_NIPS_2019} & End-to-end mask prediction by a projection loss according to \textbf{cross-label constraint} \\
			\cline{2-5}
			
			&
			\multirow{4}{*}{\makecell[c]{Panoptic\\ segmentation}} & 
			\makecell[c]{Panoptic segmentation \\ with image-level supervision} & JTSM~\cite{Shen_WSPS_2021_CVPR}  &   Unified feature representation learning under the multiple instance learning framework according to \textbf{cross-label constraint}.  \\
			\cline{3-5}
			& &\makecell[c]{Panoptic segmentation \\ with box-level supervision} & WPS\cite{Li_WSPS_ECCV_2018} & Seed area generation by \textbf{cross-label constraint} and pseudo instance mask locating by \textbf{cross-pixel similarity} \\
			\hline
			
			\multirow{12}{*}{\makecell[c]{Incomplete \\ Supervision}} &
			\multirow{6}{*}{\makecell[c]{Semantic \\ segmentation}} &	
            \multirow{3}{*}{\makecell[c]{Semi-supervised \\semantic segmentation}} &  
            AdvSemSeg~\cite{SSL_Hung2018}   & Pseudo mask quality control by adversarial learning implicitly according to \textbf{cross-image relation}\\
            & & & PseudoSeg~\cite{SSL_PseudoSeg}   & Pseudo mask regularization by enforcing \textbf{cross-view consistency} between weak and strong augmentations   \\
			& & & CAC~\cite{SSL_CAC}   & Pseudo mask regularization by enforcing \textbf{cross-view consistency} under different contexts \\
			\cline{3-5}
			
			&  &\multirow{3}{*}{\makecell[c]{Domain adaptive \\ semantic segmentation}}& BDL~\cite{UDA_BDL}  & Domain alignment by adversarial learning implicitly according to \textbf{cross-image relation} \\
			& & & DACS~\cite{UDA_DACS} & Domain alignment by domain mixing according to \textbf{cross-pixel similarity} \\
			&  & & ProDA~\cite{UDA_ProDA} & Pseudo mask regularization by enforcing \textbf{cross-view consistency} between prototype assignments\\
			\cline{2-5}
			&
			\multirow{4}{*}{\makecell[c]{Instance \\ segmentation}} &
            \multirow{4}{*}{\makecell[c]{Semi-supervised \\ instance segmentation}} & Mask$^X$ RCNN~\cite{PSIS_MaskXRCNN_CVPR_2018}   & Parameter transfer from detection to segmentation by \textbf{cross-label constraint}  \\
			& & & Shapeprop~\cite{PSIS_Shapeprop_CVPR_2020}  & Class-agnostic shape activation map learning by sailency propagation according to \textbf{cross-pixel similarity} \\
			&  & & ContraskMask~\cite{PSIS_ContrastMask_CVPR_2022} &  Foreground and background separation by pixel-wise contrastive learning according to \textbf{cross-pixel similarity} \\
			&  & & ShapeMask~\cite{PSIS_Shapemask_ICCV_2019} &  Common shape prior discovery by mask clustering according to \textbf{cross-image relation} \\
			\hline
			
			\makecell[c]{Inaccurate\\Supervision} &
			\makecell[c]{Semantic \\ segmentation} &
			\makecell[c]{Semantic segmentation\\ with noisy supervision} & ADELE~\cite{WSSS_AEL_2022_CVPR}   & Robustness boosting by enforcing multi-scale \textbf{cross-view consistency}   \\
			\bottomrule
		\end{tabular}}
            \vspace{-5mm}
	}
\end{table*}

Since a common challenge of these diverse problems lie in the big supervision gap between the weak labels and dense prediction, we can summarize the strategies for these problems from a unified perspective: how to bridging this supervision gap? This requires some heuristic priors, \emph{e.g.}, 1) \underline{\textbf{cross-label constraint}}: there exists natural constraints between weak labels and dense labels, such as an image-level category label indicates at least one pixel's label should be the same as this image-level category label; 2) \underline{\textbf{cross-pixel similarity}}: pixels with highly-similar cues, such as color, brightness and texture, probably belong to the same semantic region in an image; 3) \underline{\textbf{cross-view consistency}}: different views of the same image show consistency in both dense representation and predictions; and 4) \underline{\textbf{cross-image relation}}: the pixels from objects of the same category across different images have semantic relations, to generate pseudo dense supervision from weak labels. From this perspective, it is intriguing to see that similar strategies to employ the above priors are used for different segmentation problems, as summarized in ~\cref{tab:overview}. 

The remainder of this paper is organized as follows. We first give some foundations of label-efficient image segmentation methods in Sec.~\ref{sec:foundations}, including mathematical definitions, methodology overview, functions of the heuristic priors in the methodology, datasets and evaluation metrics for label-efficient image segmentation problems.
Then we review the existing label-efficient image segmentation methods according to our taxonomy: segmentation with no supervision in~\cref{sec:no supervision}, segmentation with inexact supervision in~\cref{sec:coarse supervision}, segmentation with incomplete supervision in~\cref{sec:incomplete supervision}, and segmentation with inaccurate supervision in~\cref{sec:noisy supervision}. In the final section, we give our conclusion and discuss several research directions and challenges.
\vspace{-3mm}

\vspace{-1mm}
\section{Foundations}
\vspace{-1mm}
\label{sec:foundations}
\subsection{Problem Definition}\label{sec:problem_definition}

In this section, we give mathematical definitions for different label-efficient image segmentation problems from a unified perspective.
Given a pre-defined set of $C$ semantic
classes encoded by $\mathcal{C}=\{0,1,\dots,C-1\}$, the task of segmentation aims to predict a dense label map $\mathbf{Y}\in\{\mathcal{C}\times\mathbb{N}\}^{H{\times}W}$ for an image $\mathbf{X}\in\{\mathbb{R}^3\}^{H{\times}W}$, where the entry of $\mathbf{Y}$ at spatial location $i\in\mathcal{I}$ is a label tuple $\mathbf{y}_i=(c_i,\iota_i)\in\mathcal{C}\times\mathbb{N}$ for the pixel in $\mathbf{X}$ at the same spatial location. Here, $H,W$ are the height and width of the image, respectively, $\mathcal{I}$ is the set of locations on a 2D lattice with size of $H{\times}W$, $\mathbb{N}$ is the space of nature numbers, $c_i$ represents the semantic class of the pixel at spatial location $i$, and $\iota_i$ represents its instance id (for semantic segmentation, $\iota_i\equiv0$). 

This goal is usually achieved by training a segmentation model on a training set $\mathcal{T}=\{(\mathbf{X}^{(n)},\mathbf{Y}^{(n)})|n\in\mathcal{N}=\{0,1,\ldots,N-1\}\}$ consisting of $N$ images, where $\mathbf{Y}^{(n)}\in\{\mathcal{C}\times\mathbb{N}\}^{H{\times}W}$ is the full dense label map for $n$-th image, \emph{i.e.}, each pixel at spatial location $i$ of image $\mathbf{X}^{(n)}$ is annotated by a label tuple $\mathbf{y}^{(n)}_i$. However, as we stated before, the full dense label $\mathbf{Y}^{(n)}$ is expensive and difficult to obtain. Alternatively, label-efficient segmentation models are trained based on weak labels which cannot cover full supervision signals but are much cheaper and easier to obtain. The definitions of label-efficient image segmentation problems are then determined by the types of the supervision from weak labels, which can be formulated from a unified perspective regarding the format of the training set. 

We first define some notations to help us give our formulations: Let $\mathbf{b}$ be the vertex coordinates of a bounding box on image $\mathbf{X}$, then we denote the spatial location set of pixels within bounding box $\mathbf{b}$ by $\mathcal{I}_{\mathbf{b}}$. Similarily, we use $\mathcal{I}_l\subseteq\mathcal{I}$ to denote the spatial location set of pixels which are labeled. In~\cref{tab:math_def}, we summarize the mathematical definitions for segmentation with the supervision of different types. Notably, in the column of ''Remark``, we give some descriptions about the formulations and show the relation between each weak supervision and the full dense supervision.
\vspace{-3mm}

\begin{table*}[t]
    \centering
    \caption{The mathematical definitions for image segmentation with different types of supervision.}\label{tab:math_def}
    \vspace{-3mm}
    \begin{tabular}{c|c|c|c}
    \toprule
    \multicolumn{2}{c|}{Supervision} & Training Set & \makecell[c]{Remark} \\
    \midrule\midrule
    \multicolumn{2}{c|}{\makecell[c]{Full Dense \\ Supervision}} & $\mathcal{T}=\{(\mathbf{X}^{(n)},\mathbf{Y}^{(n)})|n\in\mathcal{N}\}$ & N/A \\ 
    \hline
    \multicolumn{2}{c|}{\makecell[c]{No\\Supervision}} & $\mathcal{T}=\{\mathbf{X}^{(n)}|n\in\mathcal{N}\}$ & N/A \\ 
    \hline
    \multirow{5}{*}{\makecell[c]{Inexact\\Supervision}} & \makecell[c]{Image-\\level} & $\mathcal{T}=\{(\mathbf{X}^{(n)},\mathcal{C}^{(n)})|n\in\mathcal{N}\}$, $\mathcal{C}^{(n)}\subseteq\mathcal{C}$ & $\forall c\in \mathcal{C}^{(n)}, \exists i, c_i^{(n)}=c$ \\
    \cline{2-4}
                                  & \makecell[c]{Box-\\level} & $\mathcal{T}=\{(\mathbf{X}^{(n)},\mathcal{B}^{(n)})|n\in\mathcal{N}\}$, $\mathcal{B}^{(n)}=\{(\mathbf{b}^{(n,m)},\mathbf{y}^{(n,m)})\}_{m=1}^{M^{(n)}}$ & $\forall m, \exists i \in \mathcal{I}^{(n)}_{\mathbf{b}^{(n,m)}}, \mathbf{y}_i^{(n)}=\mathbf{y}^{(n,m)}$ \\
    \cline{2-4}
                                  & \makecell[c]{Scribble-\\level} & $\mathcal{T}=\{(\mathbf{X}^{(n)},\bar{\mathbf{Y}}^{(n)})|n\in\mathcal{N}\}$& $\bar{\mathbf{y}}_i^{(n)}={\mathbf{y}}_i^{(n)}$, iff $i\in\mathcal{I}^{(n)}_l\subseteq\mathcal{I}^{(n)}$\\
    \hline
    \multirow{4}{*}{\makecell[c]{Incomplete\\Supervision}} & \multirow{2}{*}{\makecell[c]{Semi}} & \multirow{2}{*}{\makecell[c]{$\mathcal{T}=\{(\mathbf{X}^{(n)},\mathbf{Y}^{(n)})|n\in\mathcal{N}_l\subset\mathcal{N}\}\bigcup\{\mathbf{X}^{(n)}|n\in\mathcal{N} \backslash \mathcal{N}_l\}$}} & \multirow{2}{*}{\makecell[c]{N/A}} \\
    &&& \\
    \cline{2-4}
                                  & \makecell[c]{Domain-\\specific} & $\mathcal{T}=\{(\mathbf{X}^{(n)},\mathbf{Y}^{(n)})|n\in\mathcal{N}_s\subset\mathcal{N}\}\bigcup\{\mathbf{X}^{(n)}|n\in \mathcal{N}_t \nsubseteq \mathcal{N}\}$ & $s$: source; $t$: target  \\
    \hline
    \multicolumn{2}{c|}{\makecell[c]{Inaccurate\\Supervision}} & $\mathcal{T}=\{(\mathbf{X}^{(n)},\tilde{\mathbf{Y}}^{(n)})|n\in\mathcal{N}$\}, & $\exists i,j \in \mathcal{I}^{(n)}, \tilde{\mathbf{y}}^{(n)}_i = {\mathbf{y}}^{(n)}_i, \tilde{\mathbf{y}}^{(n)}_j \neq {\mathbf{y}}^{(n)}_j $ \\
    \bottomrule
    
    \end{tabular}
    \vspace{-5mm}
    \label{tab:math_def}
\end{table*}


\vspace{-1mm}
\subsection{Methodology Overview}\label{sec:methodology_overview}
\vspace{-1mm}
Prior to discussing each specific label-efficient image segmentation method, we first summarize a general methodology to introduce some significant concepts involved in these label-efficient image segmentation methods. Since the main difficulty in label-efficient image segmentation is the lack of dense labels, a naive solution is first generating (dense) \textbf{pseudo labels} based on weak labels for each training image, then training a segmentation model with the pseudo labels. The above process is called \textbf{self-training}~\cite{WSSS_PSA_2018_CVPR,WSSS_BoxSup_2015_ICCV,WSIS_IRNet_CVPR_2019}. To improve the quality of the generated pseudo labels during self-training, \textbf{regularization} is usually applied to them. Formally, following the notations defined in Sec.~\ref{sec:problem_definition}, let $\mathcal{T}=\{(\mathbf{X}^{(n)},{\mathcal{Y}}^{(n)})|n\in\mathcal{N}\}$ be the given training set for label-efficient image segmentation, where ${\mathcal{Y}}^{(n)}$ denote any type of supervision, \emph{e.g.}, it can be $\mathcal{C}^{(n)}$ for image-level supervision and it can be ${\mathbf{Y}}^{(n)}$ if $n\in\mathcal{N}$ or $\emptyset$ if $n\in\mathcal{N} \backslash \mathcal{N}_l$ for semi supervision. To achieve segmentation with weak supervision, we first need a deep network $f_{\bm{\theta}}$ parameterized by $\bm{\theta}$ to produce a dense feature map $\mathbf{Z}^{(n)}=f_{\bm{\theta}}(\mathbf{X}^{(n)})$ for each training image $\mathbf{X}^{(n)}$. Then, a function $g_{\bm{\omega}}$ parameterized by $\bm{\omega}$ is applied to the dense feature map $\mathbf{Z}^{(n)}$ to generate a pseudo label map $\mathbf{P}^{(n)}=g_{\bm{\omega}}(\mathbf{Z}^{(n)})$. Such a pseudo label generation process is supervised by ${\mathcal{Y}}^{(n)}$, \emph{i.e.}, $\bm{\theta}$ and $\bm{\omega}$ are optimized by minimizing a loss function $\ell_{\texttt{GEN}}(\mathbf{P}^{(n)}, {\mathcal{Y}}^{(n)})$ whose form is depended on how to design the mapping between $\mathbf{P}^{(n)}$ and ${\mathcal{Y}}^{(n)}$. To further improve the quality of $\mathbf{P}^{(n)}$, a regularization loss $\ell_{\texttt{REG}}(\mathbf{P}^{(n)})$ is also usually used together with the generation loss $\ell_{\texttt{GEN}}$. In sum, the self-training process for label-efficient image segmentation includes the following steps: 1) Learn a pseudo label map for each training image $\mathbf{X}^{(n)}$ by minimizing the pseudo label generation loss $\ell_{\texttt{GEN}}(\mathbf{P}^{(n)}, {\mathcal{Y}}^{(n)})$ and optionally the pseudo label generation loss $\ell_{\texttt{REG}}(\mathbf{P}^{(n)})$: $$(\bm{\theta}^\ast,\bm{\omega}^\ast)=\arg\min_{\bm{\theta},\bm{\omega}}\ell_{\texttt{GEN}}(\mathbf{P}^{(n)}, {\mathcal{Y}}^{(n)})+\ell_{\texttt{REG}}(\mathbf{P}^{(n)}).$$ The optimized pseudo label map is obtained by $\hat{\mathbf{Y}}^{(n)}=g_{\bm{\omega}^\ast}(f_{\bm{\theta}^\ast}(\mathbf{X}^{(n)}))$. 2) Train a segmentation model based on the training set with the optimized pseudo labels: $\{(\mathbf{X}^{(n)},\hat{\mathbf{Y}}^{(n)})|n\in\mathcal{N}\}$.

Another methodology to address label-efficient image segmentation is \textbf{end-to-end training}~\cite{WSSS_ORL_2018_ECCV,WSIS_BBTP_NIPS_2019,WSSS_RRM_2020_AAAI,SSL_CPS}. It directly optimizes loss functions of the same form as that used in self-training, \emph{i.e.}, $\ell_{\texttt{GEN}}(\mathbf{P}^{(n)}, {\mathcal{Y}}^{(n)})+\ell_{\texttt{REG}}(\mathbf{P}^{(n)})$, by gradient descent or alternating direction method~\cite{WSSS_ORL_2018_ECCV}, if we treat the final segmentation of image $\mathbf{X}^{(n)}$ as a special case of pseudo labels $\mathbf{P}^{(n)}$ as well. End-to-end training can bypass the expensive iterative inference in self-training, and even achieve superior results for some label-efficient segmentation problems, \emph{e.g.}, semi-supervised semantic segmentation~\cite{SSL_CPS}.

\vspace{-3mm}
\subsection{Functions of the Heuristic Priors}
As shown in Table~\ref{tab:overview}, the label-efficient image segmentation methods are designed based on the heuristic priors we summarized in Sec.~\ref{sec:introduction}. In this section, we provide a systematic study on the functions of these priors in dealing with weak supervision of different types as shown in Table~\ref{tab:priors}, to facilitate readers to understand each specific method. First, we figure out which types of weak supervision a specific prior is applicable to deal with. For example, \textbf{cross-label constraint} describes the relation between fine-grained labels and coarse-grained labels, thus mostly it is applicable for inexact supervision; \textbf{cross-pixel similarity} indicates that pixels with highly-
similar cues, such as color, brightness and texture, probably belong to the same semantic region in an image. As this prior is derived from principles of perceptual grouping, it is applicable for weak supervision of all types; \textbf{cross-view consistency} and \textbf{cross-image relation} are also very general, which can be applicable for weak supervision of all types. Then, we summarize the functions of the priors in the training process, which might be varying for weak supervision of different types. For example, \textbf{cross-pixel similarity} is usually used to generate pseudo labels for no supervision, since it can guide pixel grouping in a unsupervised manner. Its function becomes pseudo label regularization for weak supervision of other types; \textbf{cross-image relation} serves as the basis to generate pseudo labels for incomplete supervision, and it is used to regularize pseudo labels for weak supervision of other types. We also list some common loss functions in Table~\ref{tab:priors} to show how to mathematically apply these priors to deal with weak supervision of different types. The notations of the loss functions follow those defined in Sec.~\ref{sec:problem_definition} and Sec.~\ref{sec:methodology_overview}. In addition, we define: $p(\cdot|\cdot)$ is a conditional probability measuring the mapping between the pseudo label $\mathbf{P}^{(n)}$ and the given label $\mathcal{Y}^{(n)}$; $[\cdot]$ is a indicator; $<\cdot,\cdot>$ is the inner product of two vectors; $\mathtt{max}_d(\cdot)$ selects the maximal values along the $d$-th channel of a tensor; The entry at pixel location $i$ on a map, \emph{e.g.}, $\mathbf{Z}$, is $\mathbf{z}_i$; $\mathbf{L}^{(n,m)}$ denotes the mask map for a given box-level label $\mathbf{b}^{(n,m)}$, \emph{i.e.}, $l_{i}^{(n,m)}=1$ if $i\in\mathcal{I}^{(n)}_{\mathbf{b}^{(n,m)}}$ and $l_{i}^{(n,m)}=0$ otherwise; $\mathbf{X}^{(n,a)}$ and $\mathbf{X}^{(n,b)}$ are two views of $\mathbf{X}^{(n)}$ obtained by different image transforms, including but not limited to geometric transforms and photometric transforms. Accordingly, the dense feature representations of $\mathbf{X}^{(n,a)}$ and $\mathbf{X}^{(n,b)}$ are $\mathbf{Z}^{(n,a)}$ and $\mathbf{Z}^{(n,b)}$, respectively; $\tau$ is a temperature parameter.

\begin{table*}[t]
\caption{Systematic study on the functions of the heuristic priors in dealing with weak supervision of different types. ``-'' means ``has not been proposed yet''.
}
\vspace{-3mm}
\label{tab:priors}
\centering
\resizebox{\textwidth}{!}{
\begin{tabular}{c|c|c|c}
\toprule
Prior            & \makecell[c]{Applicable \\ supervision }        & Function in training            & Common loss function           \\ \midrule
 \midrule
 \makecell[c]{Cross-label \\ constraint} & Inexact & {\makecell[c]{Pseudo label generation}} &  \makecell[c]{$\ell_{\texttt{GEN}}=-\sum_{n\in\mathcal{N}}\sum_{c\in\mathcal{C}^{(n)}}\log p(c|g_{\mathbf{\omega}}(\mathbf{Z}^{(n)}))$ \\ $\ell_{\texttt{GEN}}=-\sum_{n\in\mathcal{N}} \sum_{m=1}^{M^{(n)}} \sum_{d\in{\{x,y\}}} <\mathtt{max}_{d}(\mathbf{L}^{(n,m)}), \log (\mathtt{max}_{d}(\mathbf{P}^{(n,m)}))> $}\\
 \midrule

 \multirow{6}{*}{\makecell[c]{Cross-pixel \\similarity}}
 & No& Pseudo label generation & $\ell_{\texttt{GEN}} = -\sum_{n\in\mathcal{N}}\sum_{i,j}<\mathbf{z}^{(n)}_i, \mathbf{z}^{(n)}_j>
 <\mathbf{p}^{(n)}_i, \mathbf{p}^{(n)}_j>$\\
 & \multirow{1}{*}{\makecell[c]{Inexact}} & \multirow{1}{*}{\makecell[c]{Pseudo label regularization}}  & $\ell_{\texttt{REG}}=-\sum_{n\in\mathcal{N}}\sum_{i,j}[\mathbf{p}^{(n)}_i=\mathbf{p}^{(n)}_j]\log(\exp(-\|\mathbf{z}^{(n)}_i-\mathbf{z}^{(n)}_j\|))$\\
 & & & $+[\mathbf{p}^{(n)}_i\neq\mathbf{p}^{(n)}_j]\log(1-\exp(-\|\mathbf{z}^{(n)}_i-\mathbf{z}^{(n)}_j\|))$\\
 & Incomplete & Pseudo label regularization &  $
\ell_{\texttt{REG}}=-\sum_{n\in\mathcal{N} \backslash \mathcal{N}_l}\sum_{i,j}\log[\mathbf{p}^{(n)}_i=\mathbf{p}^{(n)}_j]\frac{e^{<\mathbf{z}^{(n)}_i,\mathbf{z}^{(n)}_j>/\tau}}{e^{<\mathbf{z}^{(n)}_i,\mathbf{z}^{(n)}_j>/\tau}+\sum_{\mathbf{z}_-}e^{<\mathbf{z}^{(n)}_i,\mathbf{z}_->/\tau}}
 $ \\
 & Inaccurate & Pseudo label regularization & -\\
 \midrule

 \multirow{8}{*}{\makecell[c]{Cross-view \\consistency}}
 & No& Pseudo label regularization& $\ell_{\texttt{REG}}=-\sum_{n\in\mathcal{N}}\sum_i\log\frac{e^{<\mathbf{z}^{(n,a)}_i,\mathbf{z}^{(n,b)}_i>/\tau}}{e^{<\mathbf{z}^{(n,a)}_i,\mathbf{z}^{(n,b)}_i>/\tau}+\sum_{\mathbf{z}_-}e^{<\mathbf{z}^{(n,a)}_i,\mathbf{z}_->/\tau}}$\\
 & Inexact & Pseudo label regularization & $\ell_{\texttt{REG}}=\sum_{n\in\mathcal{N}}\sum_i\|g_{\mathbf{\omega}}(\mathbf{z}^{(n,a)}_i)- g_{\mathbf{\omega}}(\mathbf{z}^{(n,b)}_i)\|$ \\
 & Incomplete&  Pseudo label regularization & $\ell_{\texttt{REG}}=-\sum_{n\in\mathcal{N}}\sum_i\log\frac{e^{<\mathbf{z}^{(n,a)}_i,\mathbf{z}^{(n,b)}_i>/\tau}}{e^{<\mathbf{z}^{(n,a)}_i,\mathbf{z}^{(n,b)}_i>/\tau}+\sum_{\mathbf{z}_-}e^{<\mathbf{z}^{(n,a)}_i,\mathbf{z}_->/\tau}}$ \\
 & Inaccurate & Pseudo label regularization & $\ell_{\texttt{REG}}=\sum_{n\in\mathcal{N}}\sum_i g_{\mathbf{\omega}}(\mathbf{z}^{(n,a)}_i)\log\frac{g_{\mathbf{\omega}}(\mathbf{z}^{(n,a)}_i)}{g_{\mathbf{\omega}}(\mathbf{z}^{(n,b)}_i)}$  \\
 \midrule

 \multirow{6}{*}{\makecell[c]{Cross-image \\relation}}
 & No& Pseudo label generation& $\ell_{\texttt{GEN}} = -\sum_{m,n\in\mathcal{N}}\sum_{i,j}<\mathbf{z}^{(m)}_i, \mathbf{z}^{(n)}_j>
 <\mathbf{p}^{(m)}_i, \mathbf{p}^{(n)}_j>$\\
 & Inexact & Pseudo label regularization & $\ell_{\texttt{REG}}=-\sum_{n,m\in\mathcal{N}}\sum_{i,j}[\mathbf{p}^{(n)}_i=\mathbf{p}^{(m)}_j]\log\frac{e^{<\mathbf{z}^{(n)}_i,\mathbf{z}^{(m)}_j>/\tau}}{e^{<\mathbf{z}^{(n)}_i,\mathbf{z}^{(m)}_j>/\tau}+\sum_{\mathbf{z}_-}e^{<\mathbf{z}^{(n)}_i,\mathbf{z}_->/\tau}}$ \\
 & Incomplete & Pseudo label generation & $\ell_{\texttt{GEN}}=-\sum_{n\in\mathcal{N}_l}\sum_i\log p(\mathbf{y}_i^{(n)}|g_{\mathbf{\omega}}(\mathbf{Z}^{(n)}))-\sum_{n\in\mathcal{N} \backslash \mathcal{N}_l}\sum_i\log p(\mathbf{p}_i^{(n)}|g_{\mathbf{\omega}}(\mathbf{Z}^{(n)}))$\\
 & Inaccurate & Pseudo label regularization & -\\
  \bottomrule
\end{tabular}

}
\vspace{-5mm}
\end{table*}

\vspace{-3mm}
\subsection{Datasets and Evaluation Metrics}
\vspace{-1mm}
Label-efficient image segmentation generally follows the evaluation metrics and datasets used for fully-supervised image segmentation. For example, label-efficient semantic segmentation methods are usually evaluated on
PASCAL VOC~\cite{pascal} as well as Cityscapes~\cite{Cityscapes_2016_CVPR} with the metric of mean Intersection-over-Union (mIoU)~\cite{YYDS_FCN_CVPR_2015} or per-pixel accuracy (Acc)~\cite{YYDS_FCN_CVPR_2015} sometimes. A special case is domain adaptive semantic segmentation, which requires two datasets from different domains. Thus, its training is performed on a synthetic dataset, such as GTA5~\cite{GTA5_2016_ECCV} and SYNTHIA~\cite{UDA_SYNTHIA} and its evaluation is conducted on Cityscapes; Label-efficient instance segmentation methods are generally evaluated on PASCAL VOC~\cite{pascal} and COCO~\cite{COCO} with the metric of average precision (AP)~\cite{COCO} or average recall (AR) sometimes; Label-efficient panoptic segmentation methods are evaluated on PASCAL VOC~\cite{pascal} and COCO-Stuff~\cite{COCO-Stuff} with the metric of panoptic quality (PQ)~\cite{YYDS_Panoptic_CVPR_2019}. We summarize the evaluation metrics and commonly used datasets for label-efficient image segmentation in Table~\ref{tab:dataset_eval}.
\begin{table}[t]
	\centering
	\renewcommand\arraystretch{1.0}
	\caption{Datasets and evaluation metrics used for label-efficient deep image segmentation.}
 \vspace{-3mm}
	\label{tab:dataset_eval}
	\footnotesize
	\resizebox{\linewidth}{!}{
		\begin{tabular}{c|c|c}
			\toprule
			\textbf{Problem} & \textbf{Datasets} & \textbf{\makecell[c]{Evaluation \\ metrics}} \\
			\midrule\midrule
			\multirow{3}{*}{\makecell[c]{Unsupervised \\ semantic segmentation}} & Cityscapes~\cite{Cityscapes_2016_CVPR}   & \multirow{3}{*}{\makecell[c]{mIoU,  Acc}} \\
			&  PASCAL VOC~\cite{pascal}   & \\
			&  COCO-Stuff~\cite{COCO-Stuff}   & \\
			\hline

            \multirow{2}{*}{\makecell[c]{Unsupervised \\ instance segmentation}}& PASCAL VOC~\cite{pascal} &  \multirow{2}{*}{\makecell[c]{AP, AR}}  \\
            &   COCO~\cite{COCO} & \\
			\hline
	
			\multirow{2}{*}{\makecell[c]{Semantic segmentation \\with image-level supervision}} & \multirow{2}{*}{\makecell[c]{PASCAL VOC~\cite{pascal}}}  &  \multirow{2}{*}{\makecell[c]{mIoU}}  \\
           & \\
            \hline

		    \makecell[c]{Semantic segmentation \\with box-level supervision}& \makecell[c]{PASCAL VOC~\cite{pascal}}  &  \makecell[c]{mIoU}  \\
			\hline

			\makecell[c]{Semantic segmentation \\with scribble-level supervision}& \makecell[c]{PASCAL VOC~\cite{pascal}}  &  \makecell[c]{mIoU}  \\
			\hline

            \multirow{2}{*}{\makecell[c]{Instance segmentation \\ with image-level supervision}}& PASCAL VOC~\cite{pascal} &   \multirow{2}{*}{AP} \\
            &   COCO~\cite{COCO} &  \\
			\hline
			
			\multirow{2}{*}{\makecell[c]{Instance segmentation \\ with box-level supervision}}& PASCAL VOC~\cite{pascal} &   \multirow{2}{*}{AP} \\
			& COCO~\cite{COCO}&  \\
			\hline
			
			\multirow{2}{*}{\makecell[c]{Panoptic segmentation \\ with image-level supervision}} & PASCAL VOC~\cite{pascal}  & \multirow{2}{*}{PQ}    \\  &
            COCO-Stuff~\cite{COCO-Stuff} &  \\
			\hline
			\multirow{2}{*}{\makecell[c]{Panoptic segmentation \\ with box-level supervision}} & PASCAL VOC~\cite{pascal}  & \multirow{2}{*}{PQ}    \\  &
            COCO-Stuff~\cite{COCO-Stuff} &  \\
			\hline
			
            \multirow{2}{*}{\makecell[c]{Semi-supervised \\semantic segmentation}} & PASCAL VOC~\cite{pascal} & \multirow{2}{*}{mIoU} \\
            & Cityscapes~\cite{Cityscapes_2016_CVPR} \\
			\hline
			
			\multirow{2}{*}{\makecell[c]{Domain adaptive \\ semantic segmentation}}& GTA5~\cite{GTA5_2016_ECCV} $\rightarrow$ Cityscapes~\cite{Cityscapes_2016_CVPR} & \multirow{2}{*}{\makecell[c]{mIoU}} \\
			& SYNTHIA~\cite{UDA_SYNTHIA} $\rightarrow$ Cityscapes~\cite{Cityscapes_2016_CVPR}  \\
			\hline
			
            \makecell[c]{Semi-supervised \\ instance segmentation} & \makecell[c]{COCO~\cite{COCO}}
			&  \makecell[c]{AP} \\
			\hline
			
			\makecell[c]{Semantic segmentation\\ with noisy supervision} &\makecell[c]{PASCAL VOC~\cite{pascal}}  &  \makecell[c]{mIoU}   \\
			\bottomrule
		\end{tabular}}
  \vspace{-5mm}
	
\end{table}

\vspace{-2mm}
\section{No Supervision}
\vspace{-1mm}
\label{sec:no supervision}
\subsection{Unsupervised Semantic Segmentation}
Semantic segmentation with no supervision, \emph{i.e.}, label-free semantic segmentation, is also known as unsupervised semantic segmentation in literature~\cite{con_maskcontrast,clu_iic,clu_ac,clu_picie}. In the early days, unsupervised semantic segmentation was achieved by performing clustering algorithms, such as K-means and Graph Cut~\cite{YYDS_NCUT_TPAMI00}, on hand-crafted image features to partition images to multiple segments with high self-similarities. Very recently, with the rapid development of unsupervised feature representation learning, especially popularized by MoCo~\cite{con_moco}, SimCLR~\cite{con_simclr}, BYOL~\cite{con_boyl}, \emph{etc},  unsupervised semantic segmentation is promoted by unsupervised dense feature representation learning, which learns a dense feature map $\mathbf{Z}=f_{\bm{\theta}}(\mathbf{X})$ for an given image $\mathbf{X}$ without any labels by a deep network $f_{\bm{\theta}}$ parameterized by $\bm{\theta}$, where $\mathbf{z}_i$ is the feature representation at spatial location $i$. A well-learned dense feature map holds the property that pixels from the same semantic region (object/stuff) have similar feature representations and those from different semantic regions have distinctive feature representations. With the well-learned dense feature maps, segmentation can be directly facilitated as training a good segmentation model based on them becomes an easy task.

Since no supervision is provided, the key to addressing unsupervised semantic segmentation is how to get supervision signals. Current methods attempted to generate dense self-supervision signals according to some heuristic priors, such as \textbf{cross-pixel similarity}, \textbf{cross-view consistency} and \textbf{cross-image relation}, as summarized in ~\cref{tab:overview}. Next, we review these methods according to the priors used to generate self-supervision signals.

\vspace{-2mm}
\subsubsection{Cross-pixel Similarity as Dense Self-supervision}
As \textbf{Cross-pixel similarity} is derived from principles of perceptual grouping, almost all unsupervised semantic segmentation methods utilized it to generate dense self-supervision. In this section, we only review the method solely relied on this prior. 

Hwang~\emph{et al.}~\cite{clu_segsort} claimed their SegSort as the first unsupervised deep learning approach for semantic segmentation. 
They first generated dense self-supervision, \emph{i.e.}, pseudo segments, by clustering with the help of a contour detector~\cite{clu_c1,clu_c2}. They then extracted prototypes for each pseudo segment, which is the average of the pixel-wise representations inside the segment. The training objective of SegSort is to pull the feature representations of pixels within a pseudo segment towards the prototype of this pseudo segment and push them away from the prototypes of other pseudo segments.

\vspace{-2mm}
\subsubsection{Cross-view Consistency as Dense Self-supervision}
\textbf{Cross-view consistency}, referring to the same object show consistency in different views, is another commonly-used prior in unsupervised semantic segmentation. 
This prior is widely used in contrastive learning based~\cite{con_moco,con_simclr} and Siamese structure based~\cite{con_boyl,con_swav,con_SimSiam} unsupervised representation learning, which have achieved great successes, and inspired unsupervised dense representation learning.

\noindent \emph{3.1.2.1~~~~Contrastive learning for cross-view consistency}\\
In contrastive learning, given an image $\mathbf{X}$, two views of the images are first generated, where one view is taken as a query $\mathbf{q}$, the other is a positive key $\mathbf{k}_+$ for the query. The optimization goal of contrastive learning is minimizing the contrastive loss:
\begin{align}\label{eq:contrastive}
    \ell(\mathbf{X}) = -\log\frac{\exp(\mathbf{q}\cdot\mathbf{k}_+/\tau)}{\sum_{\mathbf{k}\in\mathcal{K}}\exp(\mathbf{q}\cdot\mathbf{k}/\tau)},
\end{align}
where $\mathcal{K}$ is a key set consisting of both the positive key $\mathbf{k}_+$ and the negative keys from other images and $\tau$ is a temperature parameter.~\cref{eq:contrastive} is also called InfoNCE loss. It can be observed that two cruxes in contrastive learning are 1) how to generate different views and 2) how to determine positive/negative pairs.

\begin{figure}
    \centering
    \includegraphics[width=\linewidth]{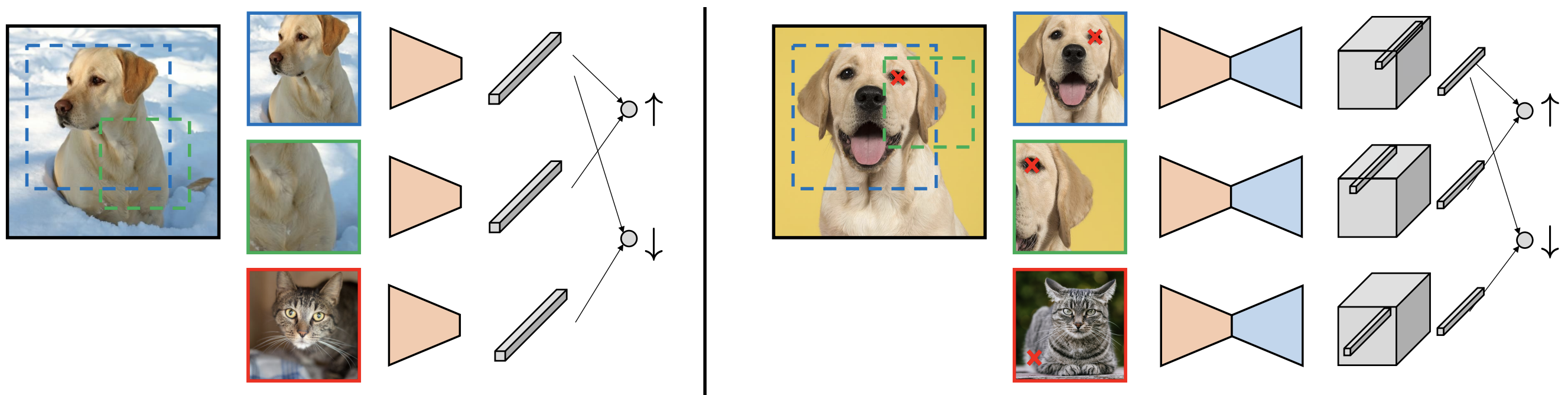}
    \vspace{-5mm}
    \caption{The illustration of VADeR (image from~\cite{con_vader}). The left is the image-level contrastive learning and the right is VADeR (pixel-wise contrastive learning).}
    \vspace{-6mm}
    \label{fig:vader}
\end{figure}

Pinheiro \emph{et al.}~\cite{con_vader} extended contrastive learning into dense representation learning for the first time. They proposed View-Agnostic Dense Representation (VADeR)~\cite{con_vader}, a pixel-wise contrastive learning method.~\cref{fig:vader} shows the comparison between VADeR and image-level contrastive learning. Following MoCo~\cite{con_moco}, the authors of VADeR 1) employed data augmentation, include geometric transforms, such as scaling, skewing, rotation and flipping, and photometric transforms, such as contrast changing and colour saturation, to generate two views $\mathbf{X}^{a},\mathbf{X}^{b}$ of one image $\mathbf{X}$, where one view is used for queries and the other is used for keys, 2) adopted two networks $f_{\bm{\theta}},f_{\bar{\bm{\theta}}}$ with the same architecture to compute feature representations of queries and keys, respectively, where the parameters $\bm{\theta}$ of the first network are trainable and the parameters $\bar{\bm{\theta}}$ of the other are obtained by the moving average of $\bm{\theta}$, and 3) maintained a memory bank $\mathcal{M}$ to store negative keys. They defined a positive pair as the feature representations at the same pixel $i$ from two different views, \emph{i.e.}, $\mathbf{z}_i^{a},\mathbf{z}_i^{b}$, and a negative pair as the feature representations at pixels from different images. Then, a pixel-wise contrastive loss was applied to learn the feature representation at each pixel $i$, as shown in Table~\ref{tab:priors}. Follow-up pixel-wise contrastive learning methods mainly attempted to improve the criterion to determine positive/negative pairs~\cite{con_densecl,con_pixpro}. 

Following the spirit of VADeR~\cite{con_vader}, Gansbeke \emph{et al.}~\cite{con_maskcontrast} proposed {MaskContrast} for unsupervised semantic segmentation, which combines SegSort~\cite{clu_segsort} and contrastive learning. They also generated two views (a query view and a key view) for each image by data augmentation, but they introduced prototypes into contrastive learning as keys. Each prototype is the mean pixel representation inside an object mask proposal (similar to the prototype in Segsort~\cite{clu_segsort}), generated by unsupervised saliency detection. They applied pixel-wise contrastive learning to pulling each pixel representation from the query view towards the prototype of its corresponding mask proposal from the key view (positive key) and pushing away it from the prototypes of other mask proposals (negative keys). It is intriguing to see their objective is also similar to Segsort~\cite{clu_segsort}. Ouali \emph{et al.}~\cite{clu_ac} proposed a pixel-wise contrastive learning method for unsupervised semantic segmentation, which is also similar to VADeR~\cite{con_vader}. But they adopted a different strategy to generate different views of an image rather than geometric and photometric transforms. They utilized different orderings over the input images using various forms of masked convolutions to construct different views of the image. Wang \emph{et al.}~\cite{con_cp2} proposed to copy and paste foreground image crops onto different background images to generate positive pairs. This copy-paste operation presents a more intuitive dense objective, \emph{i.e.},
foreground-background segmentation, which enables jointly training a backbone and a segmentation
head in the unsupervised manner.

\begin{figure}
    \centering
    \begin{overpic}[width=\linewidth]{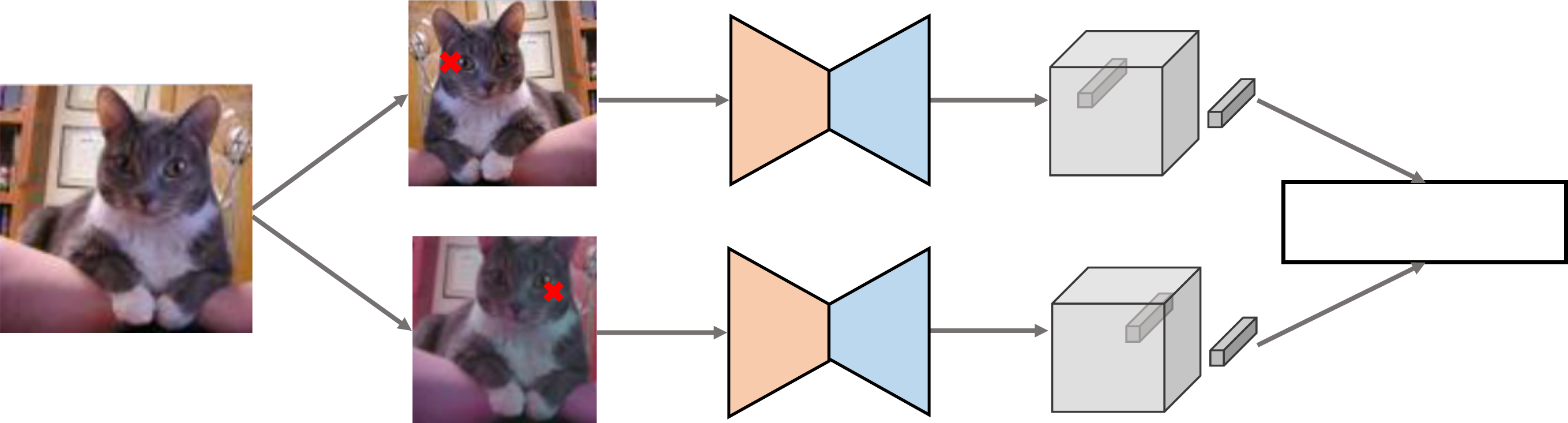}
	\put(82.5,12){\footnotesize{Consistency}}
	\end{overpic}
    \vspace{-5mm}
    \caption{Siamese structure based unsupervised dense representation learning.}
    \vspace{-5mm}
    \label{fig:siamese}
\end{figure}

\noindent \emph{3.1.2.2~~~~Siamese structure for cross-view consistency}\\
Siamese structure based unsupervised representation learning also generates two views of an image, but it maximizes the consistency between the two views without negative samples, as shown in~\cref{fig:siamese}. Generally, the representations of one view are updated online, while the gradient flow on the other is stopped~\cite{con_SimSiam}. In addition, cross-view consistency is usually represented by cluster assignment relation across the two views~\cite{con_swav}.

Ji \emph{et al.}~\cite{clu_iic} explored Siamese structure based dense representation learning for unsupervised segmentation. They proposed Invariant Information Clustering (IIC), which maximizes the mutual information between adjacent pixels from different views to ensure cross-view consistency. The mutual information was calculated based on clustering, \emph{i.e.}, a joint distribution of two adjacent pixels from the two views, respectively, defined on their similarity distribution over a series of cluster centroids. Cho \emph{et al.}~\cite{clu_picie} proposed another strategy to measure cluster assignment relation in Siamese structure based dense representation learning. Their method is called PiCIE (Pixel-level feature Clustering using Invariance and Equivariance), which performs clustering on two different views separately. Then the representation of a pixel should be closest to its corresponding cluster centroid from the same view as well as the corresponding cluster centroid from the other view.

\vspace{-2mm}
\subsubsection{Cross-image Relation as Dense Self-supervision}
Pixels from objects of the same category across images have semantic relations. However, since no supervision is available, how to mine semantically-similar pixels across images is a bottleneck. To tackle this problem, current methods usually used the features learned from unsupervised pretraining, such like MoCo~\cite{con_moco} and BYOL~\cite{con_boyl}, as the basis to build the \textbf{cross-image relation}, then refined the features.

Zhang \emph{et al.}~\cite{con_look} proposed a pixel-wise contrastive learning method by \textbf{implicitly} involving cross-image relation for unsupervised semantic segmentation. They performed clustering on features of all training images learned by unsupervised pretraining, which is able to assign a pseudo label to each pixel of all training images according to cluster assignment. The pseudo labels were used for selection of positive/negative pairs in contrastive learning.

Hamilton \emph{et al.}~\cite{con_stego} proposed STEGO, a dense representation learning method by \textbf{explicitly} mining cross-image relation according to feature correspondences for unsupervised semantic segmentation. They trained a segmentation model in a self-supervised manner by preserving the feature correspondences outputted from the network backbone and those outputted from the segmentation head, both within the same image and across image collections.
\vspace{-3mm}
\subsection{Unsupervised Instance Segmentation}
Unsupervised instance segmentation is a more challenging problem, which has not been explored until very recently. To our best knowledge, FreeSOLO~\cite{UIS_FreeSOLO_CVPR2022} is the first work to study this problem. FreeSOLO unified the segmentation process in SOLO~\cite{YYDS_SOLO_ECCV_2020} and dense self-supervised learning in DenseCL~\cite{con_densecl} as a ``query-key'' attention design. It made use of the dense correspondences involved in the self-supervised pretrained model (DenseCL) to generate some class-agnostic coarse masks for each unlabeled image by \textbf{cross-pixel similarity}. These coarse masks are then utilized to train a SOLO segmentor with a weakly-supervised~\cite{WSIS_Boxinst_CVPR_2021} and self-training manner, leading to better segmentation performance. Besides, FreeSOLO can be treated as a strong self-supervised pretext task. By finetuning on limited fully annotated images, it can achieve apparent improvements against DenseCL.

\vspace{-3mm}
\subsection{Discussion}
Unsupervised segmentation has become a promising direction recently, riding the wave of unsupervised dense representation learning. The experimental results in~\cite{con_cp2} showed that fine-tuning based on the dense representations learned by unsupervised segmentation leads to superior performance than fully-supervised segmentation model. \emph{E.g.}, the fine-tuning result of DeepLabV3~\cite{deeplabv3} pre-trained by CP$^2$~\cite{con_cp2} achieved $77.6\%$ mean Intersection-over-Union (mIoU) on the Pascal VOC 2012 dataset~\cite{pascal}, which is better than the result of fully-supervised DeepLabV3~\cite{deeplabv3} ($76.0\%$ mIoU). This encouraging result evidenced that the dense representations learned in the unsupervised manner are well structurized and can facilitate segmentation. 

Nevertheless, the exploration of dense representation learning for unsupervised segmentation is still in the preliminary stage. Different to image-level representation learning, dense representation learning requires some regional priors to indicate the relation between pixels, \emph{i.e.}, whether they belong to the same semantic region or not. This is essentially the same goal of unsupervised segmentation. Consequently, unsupervised segmentation suffers from a chicken and egg situation. How to introduce more accurate regional priors initially or how to refine them during learning is worth exploring in future.

\vspace{-3mm}
\section{Inexact Supervision}
\label{sec:coarse supervision}

As shown in~\cref{fig:weak_label} and~\cref{tab:math_def}, inexact supervision can be image-level (only category labels are provided for each training image), box-level (besides category labels, object bounding boxes are also annotated for each training image) or scribble-level (a subset of pixels in each training image are annotated). Segmentation with inexact supervision is also usually called weakly-supervised segmentation in literature. Although in a broader sense, this term can also refer to segmentation with other types of weak supervision, \emph{e.g.}, incomplete supervision, we use it specifically to refer to segmentation with inexact supervision following the precious literature in this section.

\vspace{-3mm}
\subsection{Segmentation with Image-level Supervision}

\subsubsection{Semantic segmentation with image-level supervision}
In this section, we review the methods to perform semantic segmentation with image-level supervision. The difficulty lies in this problem is the large supervision gap between image-level supervision and pixel-level dense prediction. The former is used for training image classification models, while the latter is required to delineate object/stuff masks. 

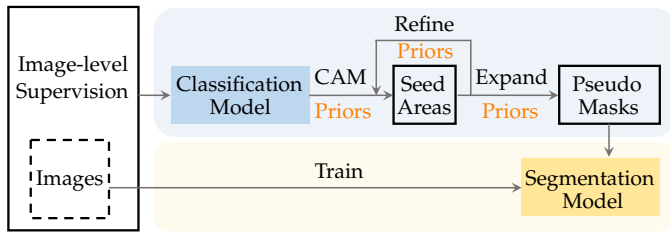
\begin{figure}[t]
	\centering
	\begin{overpic}[width=\linewidth]{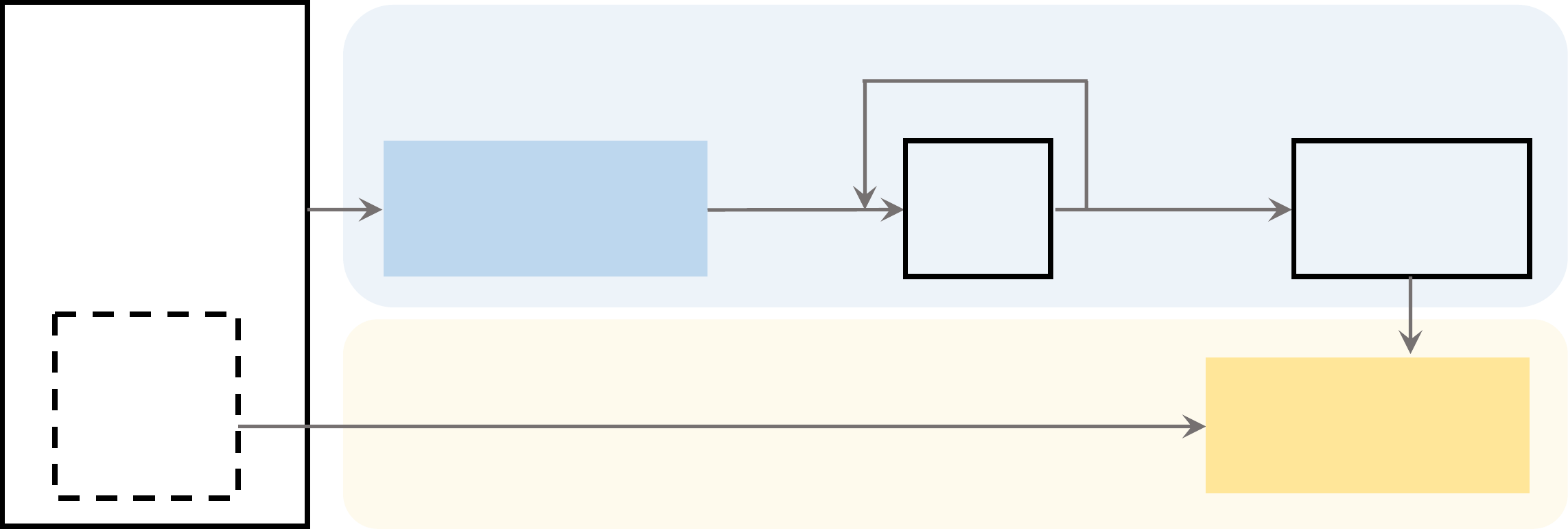}
	\put(4.3,7){\footnotesize{Images}}
	\put(1.6,24){\footnotesize{Image-level}}
	\put(1.3,20){\footnotesize{Supervision}}
	
	\put(25.3,21){\footnotesize{Classification}}
	\put(30.3,17.5){\footnotesize{Model}}
	\put(58.9,21){\footnotesize{Seed}}
	\put(58.3,17.5){\footnotesize{Areas}}
	\put(84.5,21){\footnotesize{Pseudo}}
	\put(85.4,17.5){\footnotesize{Masks}}
	
	\put(46,22){\footnotesize{CAM}}
	\put(46,17){\footnotesize{\textcolor{orange}{Priors}}}
	\put(70,22){\footnotesize{Expand}}
	\put(71,17){\footnotesize{\textcolor{orange}{Priors}}}
	\put(57.9,30){\footnotesize{Refine}}
	\put(58.3,26){\footnotesize{\textcolor{orange}{Priors}}}
	
	\put(46, 8){\footnotesize{Train}}
	\put(77.5, 7){\footnotesize{Segmentation}}
	\put(83, 3.5){\footnotesize{Model}}

	\end{overpic}
        \vspace{-5mm}
	\caption{The mainstream pipeline for semantic segmentation with image-level supervision.}
        \vspace{-5mm}
	\label{fig:WSSS_F1_Pipeline}
\end{figure}

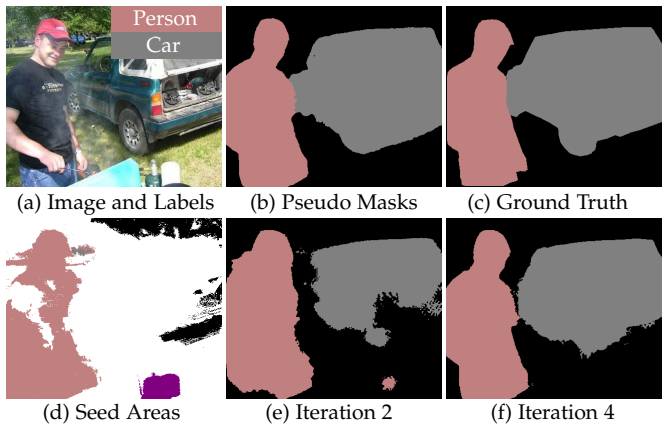
\begin{figure}[t]
	\centering
	\begin{overpic}[width=\columnwidth]{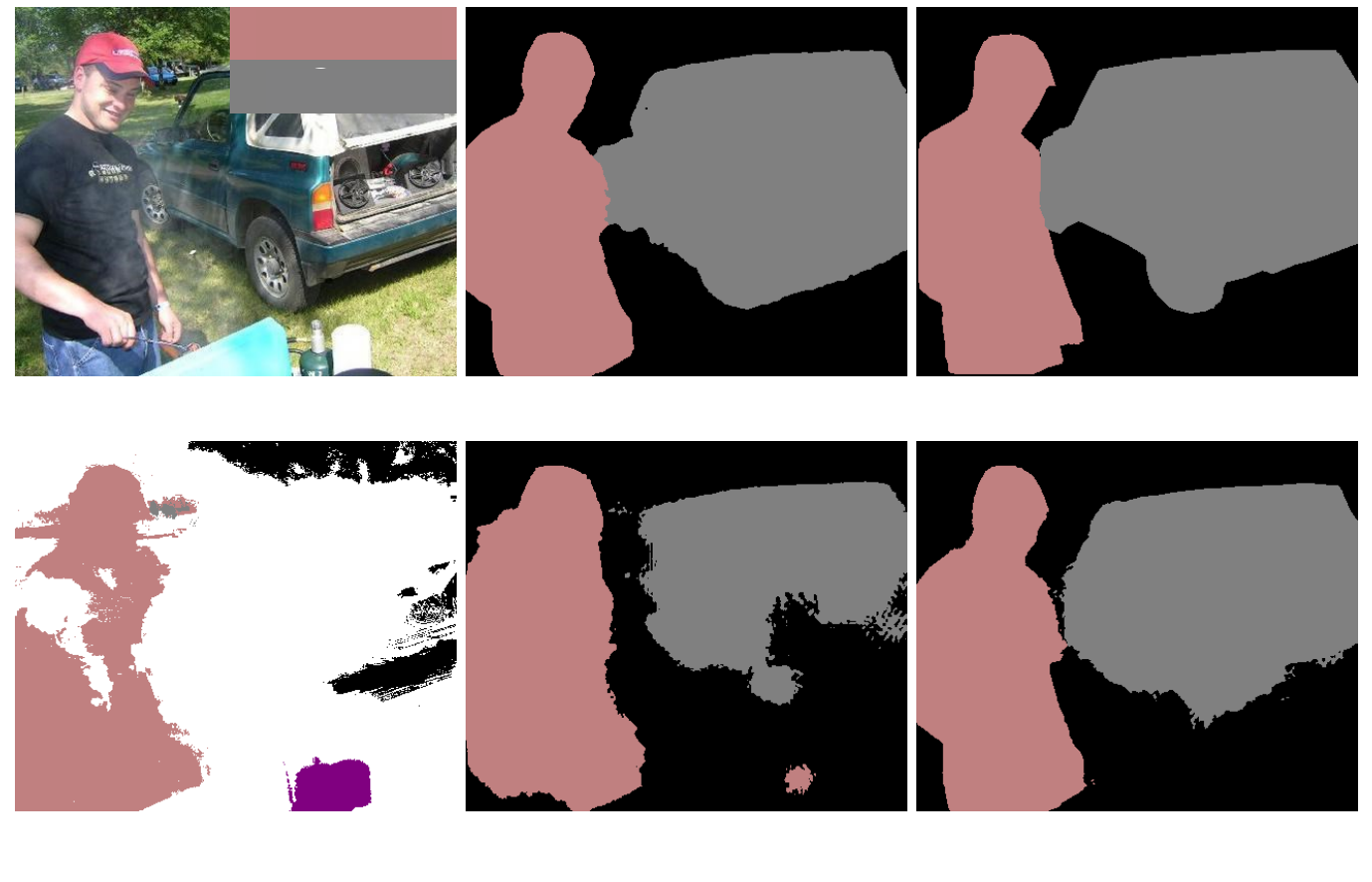}
	\put(20, 61){\footnotesize{\textcolor{white}{Person}}}
	\put(22, 57){\footnotesize{\textcolor{white}{Car}}}
	\put(6.5, 2){\footnotesize{(d) Seed Areas}}
	\put(2.6, 33.5){\footnotesize{(a) Image and Labels}}
	\put(37.5, 33.5){\footnotesize{(b) Pseudo Masks}}
	\put(70, 33.5){\footnotesize{(c) Ground Truth}}
	\put(39.8, 2){\footnotesize{(e) Iteration 2}}
	\put(73.5, 2){\footnotesize{(f) Iteration 4}}
	\end{overpic}
        \vspace{-6mm}
	\caption{Pseudo mask generation from seed areas. (Image from \cite{WSSS_IAL_2020_IJCV}.)}
        \vspace{-6mm}
	\label{fig:WSSS_F4_RDC}
\end{figure}

To bridge the supervision gap, as shown in~\cref{fig:WSSS_F1_Pipeline}, researchers followed a two-stage pipeline, where pseudo masks (dense labels) are generated for each training image based on a classification model trained with image-level supervision in the first stage, then a semantic segmentation model is trained based on the pseudo masks (An illustration for pseudo mask generation from seed areas is shown in~\cref{fig:WSSS_F4_RDC}). Since the pseudo masks are inevitably noisy, training segmentation models from pseudo masks is equivalently the problem of segmentation with noisy supervision. Thus, we review the methods for the second stage in~\cref{sec:noisy supervision} and we mainly review the methods for the first stage in this section.

The goal of the first stage is to generate high-quality pseudo masks, which consists of two subsequent steps: 1) In the first step, some seed areas are obtained in each training image based on the information derived from the classification model. This step is usually achieved by computing class activation maps (CAMs)~\cite{WSSS_CAM_2016_CVPR, WSSS_Grad_CAM_ICCV_2017, WSSS_Grad_CAM++_WACV_2018} of the classification model, so that the seed areas can cover discriminative semantic regions in each image. 2) Then, in the second step, pseudo masks (dense labels) are generated by propagating the semantic information from the seed areas to the whole image. This pseudo mask generation process is usually iterative, involving self-training the segmentation model. Existing methods made efforts in either refining seed areas to make them more accurate and complete or generating more reliable pseudo masks on the basis of seed areas, by involving the common priors we summarized. Next, we review the efforts have been made in \textbf{seed area refinement} and \textbf{pseudo mask generation} based on different priors.

\noindent \emph{4.1.1.1~~~~Seed area refinement by cross-label constraint}\\
\noindent The class activation maps (CAMs)~\cite{WSSS_CAM_2016_CVPR, WSSS_Grad_CAM_ICCV_2017, WSSS_Grad_CAM++_WACV_2018} serve as the \emph{de facto} tools to generate seed areas based on classification models, which are adopted in all semantic segmentation methods with image-level supervision. CAMs essentially make use of the prior of \textbf{cross-label constraint} to locate the seed areas in an image from the information provided by a classification model. However, the seed areas captured by CAMs suffer from two limitations: 1) Incompleteness: a CAM usually fails to cover the entire semantic region of 
the target class; 2) Redundancy: a CAM may overlap the regions of other classes. To address these issues, researchers designed several strategies to improve CAMs, producing CAM-like maps for seed area refinement, including: 1) \textbf{expanding by ensemble}~\cite{WSSS_OAA_2019_ICCV, WSSS_RDC_2018_CVPR, WSSS_USI_2019_CVPR}, 2) \textbf{re-finding by erasing}~\cite{WSSS_AE_2017_CVPR, WSSS_ACL_2018_CVPR, WSSS_ECSNet_2021_ICCV}, 3) \textbf{discovering by optimization}~\cite{WSSS_Advcam_2021_CVPR, WSSS_BottleNeck_2021_NEURIPS, WSSS_SCE_2020_CVPR} and 4) \textbf{reasoning by decoupling}~\cite{WSSS_CONTA_2020_NEURIPS,WSSS_CDA_2021_ICCV} 

{\noindent\textbf{Expanding by ensemble.}} Since a CAM usually cannot cover the entire semantic region of the target class, an intuitive strategy is to expand the seed area by an ensemble of different CAMs. Wei \emph{et al.}~\cite{WSSS_RDC_2018_CVPR} proposed to enlarge the seed area by an ensemble of CAMs computed using multiple dilated convolutional (MDC) blocks of different dilation rates. Different from~\cite{WSSS_RDC_2018_CVPR} that formed a fixed combination of different CAMs, Lee \emph{et al.}~\cite{WSSS_USI_2019_CVPR} proposed to generate a variety of CAMs by using random combinations of hidden units in the classification model. This was realized by applying spatial dropout~\cite{WSSS_dropout_JMLR_2014} to the feature maps of the classification model. Jiang \emph{et al.}~\cite{WSSS_OAA_2019_ICCV} pointed out an interesting observation that a CAM is prone to shifting to different regions of the target class during the training progress~\cite{WSSS_OAA_2019_ICCV}. Motivated by this concept, they proposed an online accumulation method, which acquires the CAM-like map by accumulating the CAMs at different training phases.

{\noindent\textbf{Re-finding by erasing.}} Another intuitive strategy is erasing the current CAM, then enforcing the classification model to re-find other regions to form a new CAM. The ensemble of the new and old CAMs can expand the seed area to cover a more complete semantic region of the target class.
Wei \emph{et al.}~\cite{WSSS_AE_2017_CVPR} proposed a pioneer ``erasing'' framework, which iteratively erases the current CAM then discovers another one. Since the semantic regions covered by the current CAM were erased, the classification network was encouraged to discover other related semantic regions for maintaining the classification prediction. Finally, the CAMs discovered at all iterations were assembled to obtain the final CAM-like map which is likely to cover entire object regions. Zhang \emph{et al.}~\cite{WSSS_ACL_2018_CVPR} improved the ``erasing'' framework by introducing Adversarial Complementary Learning (ACoL). They produced two parallel classifiers in a single network, which were initialized differently, so that the CAMs produced by each classifier can encourage the counterpart classifier to discover complementary
semantic regions. ACoL fused the CAMs from both of the two classifiers as the final CAM-like map. Rather than only aggregating the CAMs obtained from different steps, Sun \emph{et al.} \cite{WSSS_ECSNet_2021_ICCV} suggested that the interaction between CAMs may provide additional supervision to learn more pixel-level information. On the basis of this assumption, they proposed an Erased CAM Supervision Net (ECS-Net) to sample reliable pixels from the erased CAM for generating pixel-level pseudo labels to supervise the generation of a new CAM.

{\noindent\textbf{Discovering by optimization.}}  Instead of fusing different CAMs, one can also discover seed areas by encouraging the classification model to see larger regions during optimization. Lee \emph{et al.}~\cite{WSSS_Advcam_2021_CVPR} applied an anti-adversarial manner to perturb images along pixel gradients which are regarding to the classification of the target class. This manipulation forces larger semantic regions to participate in classification and produces a CAM-like map to identify a more complete region of the target object. From the perspective of the information bottleneck principle~\cite{WSSS_information_bottleneck_related_ICLR_2018,WSSS_information_bottleneck_related2_ITW_2015}, Wang \emph{et al.}~\cite{WSSS_BottleNeck_2021_NEURIPS} explained why a CAM is unable to cover the entire semantic region of the target class. According to the information bottleneck theory, information regarding the input image is compressed through sequential network layers and only the task-relevant information can pass the final layer of the network. In the classification task, the most relevant information often exists in the discriminative part of an object or stuff, and thus, the trained classification model is prone to ignoring the non-discriminative parts. Moreover, Chang \emph{et al.}~\cite{WSSS_SCE_2020_CVPR} pointed out the core reason why a CAM only covers a discriminative sub-region of the target class is the objective function to train classification networks does not require the networks to ``see'' the entire region of the target class. In light of this opinion, they introduced an additional self-supervised sub-category classification task to enforce the networks to also pay attention to the non-discriminative sub-regions of the target class.

{\noindent\textbf{Reasoning by decoupling.}} The reason why a CAM may overlap the regions of non-target classes might be there are so called co-occurrence classes, \emph{e.g.}, ``horse'' and ``person'' often co-occur with each other. The co-occurrence classes mislead the classification model. Zhang \emph{et al.}~\cite{WSSS_CONTA_2020_NEURIPS} addressed this issue by causal inference~\cite{WSSS_causal_inference_2019_Bio}. They analyzed and decoupled the causality between an image and the confounder set, \emph{i.e.}, the set of co-occurrence classes, to prevent seed areas from being expounded to redundant regions. Moreover, in order to avoid background interference, Su \emph{et al.}~\cite{WSSS_CDA_2021_ICCV} used copy-paste operation to decouple the relation among the confounder set. They pasted a foreground image onto different images, so that the classes from the foreground image are free from a stereotypical contextual relationship with the corresponding background, encouraging seed areas to focus more on the foreground regions.

\noindent \emph{4.1.1.2~~~~Seed area refinement by cross-pixel similarity}\\
\noindent The ``erasing'' framework may mislead seed areas to gradually expand to regions of incorrect semantics. To address this issue, some recent methods made use of the prior of \textbf{cross-pixel similarity} to guide seed area expanding. This can be instantiated by involving saliency maps~\cite{WSSS_SD_CVPR_2017} to ensure expanding to regions of similar semantics.

Hou \emph{et al.}~\cite{WSSS_SE_2018_NEURIPS} proposed Self-Erasing Network (SeeNet), which is the first attempt to use saliency maps~\cite{WSSS_SD_CVPR_2017} to guide seed area refinement.
Saliency maps were then widely used in follow-up works. One example is~\cite{WSSS_DRS_2021_AAAI}, which incorporated saliency maps as a soft guidance to control seed area expanding. Xu \emph{et al.}~\cite{WSSS_LAT_2021_ICCV} proposed a cross-task affinity learning framework to joint learn classification task, saliency detection task and segmentation task. In particular, the authors pointed out the saliency detection task and the segmentation task possess similar structured semantics, which motivated them to learn cross-task affinity maps from the saliency and segmentation representations to refine seed areas.

\noindent \emph{4.1.1.3~~~~Seed area refinement by cross-view consistency}\\ 
\noindent Some researchers proposed to use \textbf{cross-view consistency} to improve the quality of seed areas, since \textbf{cross-view consistency} can encourage the semantic consistency between CAMs obtained from different spatial perturbations of the same image~\cite{WSSS_SEAM_2020_CVPR, WSSS_CP_2021_ICCV}.

Wang \emph{et al.}~\cite{WSSS_SEAM_2020_CVPR} designed a Siamese network for seed area refinement. The Siamese network contains two branches with different data augmentations, where one branch adds additional affine transformations to each input image compared with the counterpart. Based on the Siamese network, the authors encouraged the CAMs computed from the two branches to keep consistent. Following the spirit of ~\cite{WSSS_SEAM_2020_CVPR}, Zhang \emph{et al.}~\cite{WSSS_CP_2021_ICCV} considered the essence of seed area expanding as an increase in information and they proved that the information of the ensemble of CAMs generated from a pair of images with complementary parts, named CP pair, is always greater than or equal to one individual CAM. Based on this thought, the authors proposed a CP Network to reduce the gap between the ensemble of CAMs generated by the CP pair and the initial CAM. The CP Network delivers a CAM which is more informative and can cover more complete semantic regions. More recently, Jiang~\emph{et al.}~\cite{WSSS_L2G_2022_CVPR} proposed a novel augmentation technique to construct multiple views that enforces local-to-global consistencies among CAMs computed from a bag of views with different resolutions.

\noindent \emph{4.1.1.4~~~~Seed area refinement by cross-image relation}\\ 
\noindent \textbf{Cross-image relation} can be used to strengthen the robustness of seed area generation by encouraging pixel-wise interactions among different images with semantic co-occurrence.  

Sun \emph{et al.}~\cite{WSSS_MCIS_2020_ECCV} proposed two neural co-attentions to complimentarily capture cross-image semantic similarities and differences between each pair of images with semantic co-occurrence. One is co-attention which aims to help CAMs to cover complementary parts of objects belonging to the same category, the other is contrastive co-attention which was designed to help CAMs to discriminate semantics of objects belonging to different categories. Li \emph{et al.}~\cite{WSSS_Group_2021_AAAI} proposed group-wise semantic mining (GWSM) to capture \textbf{cross-image relation} among a group of images rather than a pair of images via a graph neural network (GNN)~\cite{WSSS_GNN_2008_TNN}. The authors plugged the GNN into the classification model to propagate pixel-level semantic correspondences both within the same image and across images, progressively driving CAMs to cover more complete object regions. Very recently, Zhou \emph{et al.}~\cite{WSSS_RSC_2022_CVPR} introduced a memory bank consisting of dataset-level prototypical region feature embeddings to modulate learned CAMs by \textbf{cross-image relation}. 

\noindent \emph{4.1.1.5~~~~Pseudo mask generation by cross-pixel similarity}\\
\noindent Pseudo masks are usually generated in an iterative manner, \emph{i.e.}, starting from the seed areas, alternatively expounding the current pseudo masks (initialized by seed areas) and then using a segmentation model learned by self-training to update the pseudo masks. The prior of \textbf{cross-pixel similarity} is widely used in pseudo mask expounding, where the key is how to measure the similarity (affinity) among pixels, so that the expounding process can be regularized. The affinity can be based on either \textbf{low-level features} (\emph{e.g.,} color and texture)~\cite{WSSS_SEC_2016_ECCV,WSSS_RRM_2020_AAAI,WSSS_DSRG_2018_CVPR,WSSS_MCOF_2018_CVPR} or \textbf{high-level semantics}~\cite{WSSS_PSA_2018_CVPR,WSSS_IAL_2020_IJCV,WSSS_AALR_2021_MM}.

{\noindent\textbf{Affinity learning with low-level cues.}} Kolesnikov \emph{et al.}~\cite{WSSS_SEC_2016_ECCV} proposed three principles for training a segmentation model to generate pseudo masks from seed areas: 1) Training only with the pseudo labels with high confidences; 2) Updated pseudo labels should be consistent with the given image-level labels; and 3) Constraining updated pseudo masks to respect object boundaries. These three principles were widely adopted in follow-up works. Kolesnikov \emph{et al.} achieved the third one by measuring the pixel-level affinity in terms of low-level image cues, \emph{e.g.}, color and spatial location. Huang \emph{et al.}~\cite{WSSS_DSRG_2018_CVPR} followed the three principles. They adopted a seeded region growing (SRG) strategy~\cite{WSSS_DRG_TPAMI_1994}, which expounds pseudo masks to neighborhood pixels with high confidence. Wang \emph{et al.}~\cite{WSSS_MCOF_2018_CVPR} utilized saliency map to guide pseudo mask expounding. Zhang \emph{et al.}~\cite{WSSS_RRM_2020_AAAI} also followed the three principles and proposed an end-to-end reliable region mining (RRM) framework, jointly performing classification and segmentation. They introduced a dense energy loss~\cite{WSSS_NC_2018_CVPR} to propagate semantic information from seed areas to the remaining unlabeled regions by leveraging low-level image cues.

{\noindent\textbf{Affinity learning with high-level learned features.}} 
The affinity can be also measured by the similarity between learned high-level features.  
Ahn \emph{et al.}~\cite{WSSS_PSA_2018_CVPR} proposed AffinityNet to learn a pixel-level feature extractor that is supervised by semantic labels of seed areas.
The trained AffinityNet was used to build a pixel-to-pixel semantic affinity matrix, which was further applied in random walk~\cite{WSSS_RWR_2006_TPAMI} to generate pseudo masks. Ru \emph{et al.}~\cite{WSSS_afa_2022_CVPR} directly made use of the attentions learned from a Transformer to form the pixel-to-pixel semantic affinity matrix. Wang \emph{et al.}~\cite{WSSS_IAL_2020_IJCV} built an end-to-end iterative affinity learning framework (IAL), which is similar to the previously introduced RRM framework~\cite{WSSS_RRM_2020_AAAI}. The difference is the pairwise affinity matrix in~\cite{WSSS_IAL_2020_IJCV} was built on learned high-level features rather than low-level image cues.~\cref{fig:WSSS_F4_RDC} shows one example of the pseudo masks progressively generated from seed areas by IAL. Zhang \emph{et al.}~\cite{WSSS_AALR_2021_MM} pointed out that treating all seed areas equally may result in over-fitting to some erroneous seeds. To address this issue, the authors introduced an adaptive affinity loss, where adaptive weights were adopted to measure the reliability of the pixel-wise affinities.

\noindent \emph{4.1.1.6~~~~Pseudo mask generation by cross-image relation}\\
\noindent Affinity learning can be also benefited from \textbf{cross-image relation}.
Fan \emph{et al.}~\cite{WSSS_CIAN_2020_AAAI} built a cross-image affinity module (CIAN) for pseudo mask generation from pairs of images with semantic co-occurrence. In each pair of images, one image was taken as a query image, and the other was a reference image. The feature map of the query image was adjusted by the feature map of the reference image according to
the pixel-wise affinities between the two of them, which leaded to more complete and accurate pseudo masks.

\vspace{-2mm}
\subsubsection{Instance segmentation with image-level supervision}

\begin{figure}[!t]
    \centering
	\begin{overpic}[width=\linewidth]{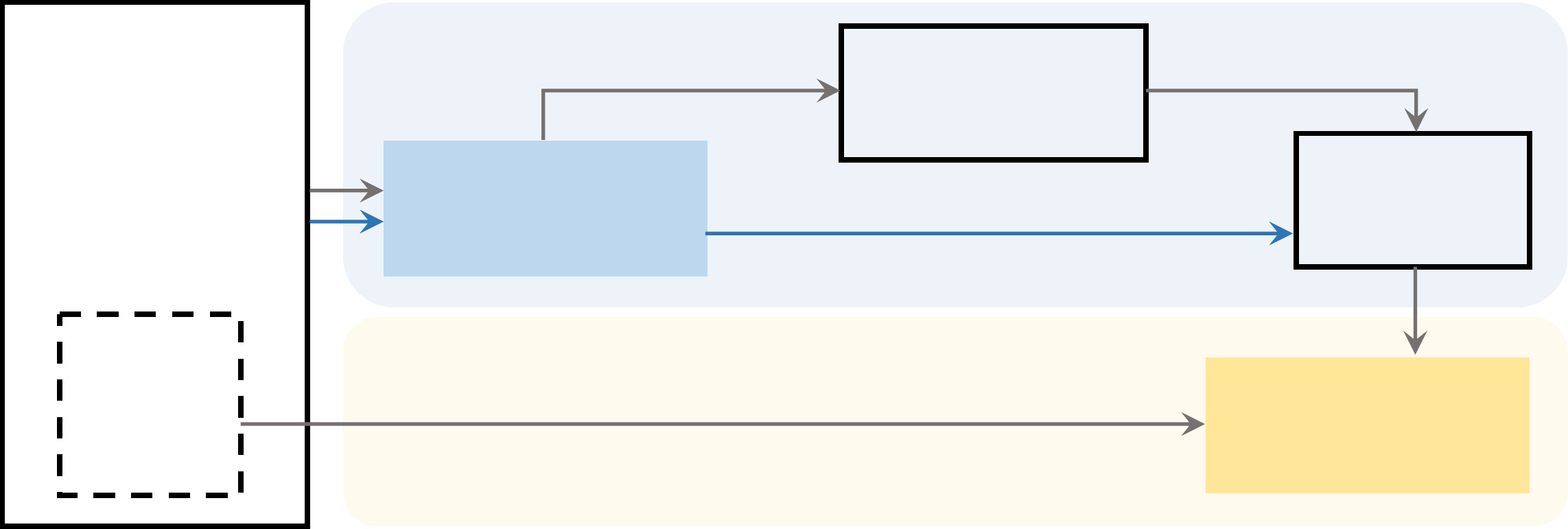}
	\put(4.3,7){\footnotesize{Images}}
	\put(1.6,24){\footnotesize{Image-level}}
	\put(1.3,20){\footnotesize{Supervision}}
	
	\put(25.3,21){\footnotesize{Classification}}
	\put(30.3,17.5){\footnotesize{Model}}
	\put(59.5,29){\footnotesize{Seed}}
	\put(58.8,25.5){\footnotesize{Areas}}
	\put(84.5,21.5){\footnotesize{Pseudo}}
	\put(85.4,18){\footnotesize{Masks}}
	
	\put(24.5,29){\footnotesize{CAM + Peak Locate}}
	\put(44.5,25){\footnotesize{\textcolor{orange}{Priors}}}
	\put(73.5,29){\footnotesize{Expand}}
	\put(73.5,25){\footnotesize{\textcolor{orange}{Priors}}}
	\put(49,20){\footnotesize{End-to-end Generate}}
	\put(59,16){\footnotesize{\textcolor{orange}{Priors}}}

	\put(46, 8){\footnotesize{Train}}
	\put(77.5, 7){\footnotesize{Segmentation}}
	\put(83, 3.5){\footnotesize{Model}}

	\end{overpic}
 \vspace{-5mm}
    \caption{The mainstream pipeline for instance segmentation with image-level supervision.}
    \vspace{-3mm}
    \label{fig:psis_arch1}
\end{figure}
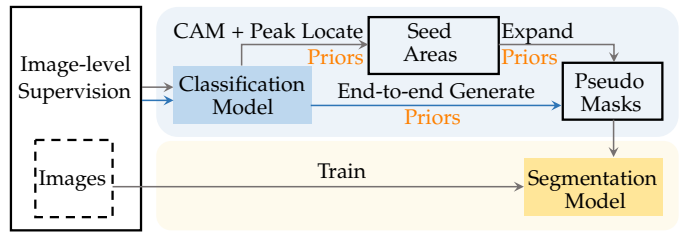

In this section, we review instance segmentation methods with image-level supervision. As shown in~\cref{fig:psis_arch1}, similar to the strategies used for semantic segmentation with image-level supervision, the methods for instance segmentation with image-level supervision also first generated pseudo masks, then trained the segmentation model. But since instance segmentation further requires locating each object, pseudo masks should be instance-level rather than category-level. Instance-level pseudo masks can be obtained by 1) expounding instance-level seed areas by \textbf{self-training} according to \textbf{cross-pixel similarity} (the gray flow line in~\cref{fig:psis_arch1} ) or 2) \textbf{end-to-end training} according to \textbf{cross-label constraint} (the blue flow line in~\cref{fig:psis_arch1} )). 

\begin{figure}[t]
	\centering
	\begin{overpic}[width=\linewidth]{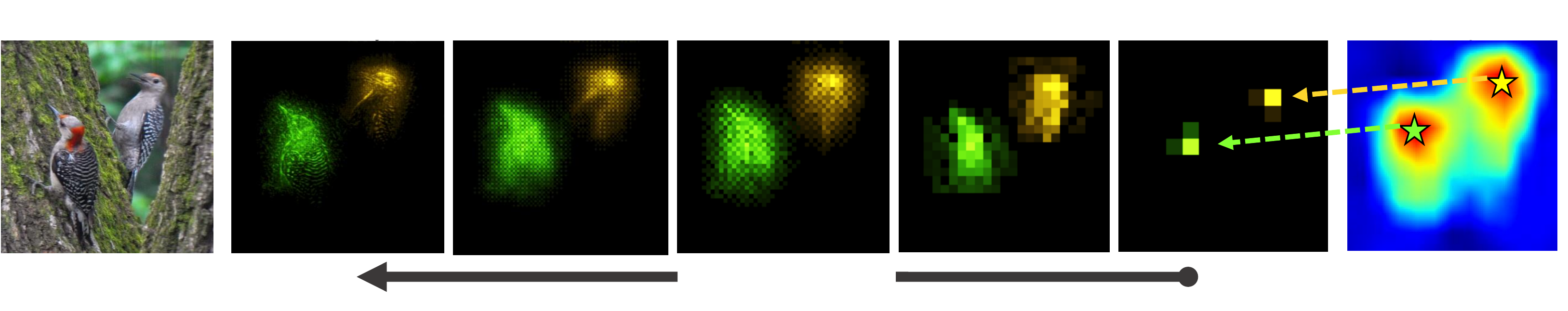}
	\put(4.5,1){\tiny{Image}}
	\put(48,2){\tiny{Peak}}
	\put(46,0){\tiny{Backprop}}
	\put(82.5,2.0){\tiny{Class Response Map}}
	\put(89.0,0.){\tiny{(for bird)}}
	\put(15.5,18){\tiny{Bottom layer}}
	\put(30,18){\tiny{Conv Block1}}
	\put(44,18){\tiny{Conv Block2}}
	\put(58,18){\tiny{Conv Block3}}
	\put(72,18){\tiny{Conv Block4}}
	\put(88,18){\tiny{Top layer}}
	\end{overpic}
        \vspace{-8mm}
	\caption{\vspace{-2pt}Illustration of instance-level seed area generation by peak back-propagation. (Image from \cite{WSIS_PRM_CVPR_2018}.)}
    \vspace{-6mm}
	\label{fig_prm}
\end{figure}
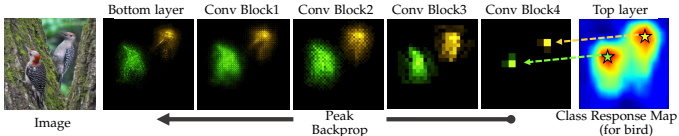

\noindent \emph{4.1.2.1~~~~Instance-level seed area generation}\\
\noindent Intuitively, instance-level seed areas can be obtained from category-level seed areas by peak locating~\cite{WSIS_PRM_CVPR_2018}. PRM~\cite{WSIS_PRM_CVPR_2018} is the first work to solve this task through introducing peak response maps. High-confidence responses (peaks) in seed areas provided by the classification model for a specific class imply the possible locations of instances belonging to the class. The peaks of seed areas were gradually merged into a few ones by a learned kernel, each of which was assumed to correspond to one instance. As shown in~\cref{fig_prm}, the peak response map was obtained by back-propagating semantic information from the peaks to the whole image. GrabCut~\cite{YYDS_Grabcut_TOG_2004} was employed to locate the boundary for each instance on the peak response map. 

\noindent \emph{4.1.2.2~~~~Instance-level pseudo mask generation}\\
\noindent \textbf{Expounding by self-training.} Generating instance-level pseudo masks from instance-level seed areas usually involves a self-training process. WISE~\cite{WSIS_Wheremask_BMVC_2019} and IAM~\cite{WSIS_IAM_CVPR_2019} are two self-training based works built on PRM~\cite{WSIS_PRM_CVPR_2018}. WISE selected the local maxima of the output of PRM as the pseudo labels to train an instance segmentation model. The authors of IAM~\cite{WSIS_IAM_CVPR_2019} pointed out that PRM can only identify the most discriminative part of an instance. In IAM, they generated instance-level pseudo masks by expounding the peak response maps through a differentiable filling module. IRNet~\cite{WSIS_IRNet_CVPR_2019} generated instance-level pseudo masks by learning a class-agnostic instance map and pairwise semantic affinities simultaneously from high-confidence seeds. The latter was used to identify object boundaries which enabled the propagation from the former to form accurate instance-level pseudo masks.

\noindent\textbf{Generating by end-to-end training.} Unlike the above self-training based methods which are composed of multiple offline stages, there are also end-to-end training based methods which directly transform image-level labels to instance-level pseudo masks according to \textbf{cross-label constraint}. Ge \emph{et al.}~\cite{WSIS_Labelpenet_ICCV_2019} proposed Label-PEnet to explored to transfer image-level labels to pixel-level labels in an online and coarse-to-fine manner. They designed a cascaded pipeline which was composed of four parallel modules, \emph{i.e.}, classification, object detection, instance refinement, and instance segmentation. These modules shared the same backbone and were trained with a curriculum learning strategy, which generalized labels from image-level supervision to pixel-level pseudo masks gradually with increasing accuracy. Hwang \emph{et al.}~\cite{WSIS_CL_WACV_2021} introduced a simple yet efficient community learning framework, WSIS-CL, which formed a positive feedback loop between object detection and instance mask generation. It adopted the most popular method OICR~\cite{WSOD_OICR_CVPR_2017} in weakly supervised object detection to generate object proposals and supervised the instance segmentation model by the combination of multi-level CAMs from spatial pyramid pooling~\cite{YYDS_SPP_TPAMI_2015} for high-confidence proposals. PDSL~\cite{WSIS_PDSL_ICCV_2021} performed weakly supervised object detection and self-supervised instance segmentation in parallel for the same proposal obtained by selective search~\cite{YYDS_SelectiveSearch_IJCV13}. The results of these two branches are 
constrained by conducting correlation learning to keep consistent predictions. Zhou \emph{et al.}~\cite{response_2000_2022_ECCV} presented a simple way to expand the vocabulary of detectors and instance segmentors to tens of thousands of concepts by training classifiers on image classification data.

\vspace{-2mm}
\subsubsection{Panoptic segmentation with image-level supervision}
Panoptic segmentation with image-level supervision has not been widely explored, probably because it is very challenging.  As far as we know,  Shen \emph{et al.}~\cite{Shen_WSPS_2021_CVPR} is the only work to address this problem. They proposed a joint thing-and-stuff mining (JTSM) framework, where mask-of-interest pooling was designed to form fixed-size pixel-accurate feature
representations for segments from arbitrary categories. The unified feature representations for both things and stuff enable connecting pixel-level pseudo labels to image-level labels by multiple instance learning, according to \textbf{cross-label constraint}. The pseudo masks were refined by Grabcut~\cite{YYDS_Grabcut_TOG_2004}, according to \textbf{cross-pixel similarity}, and used for self-training the panoptic segmentation model.

\vspace{-2mm}
\subsection{Segmentation with Box-level Supervision}

\subsubsection{Semantic segmentation with box-level supervision}

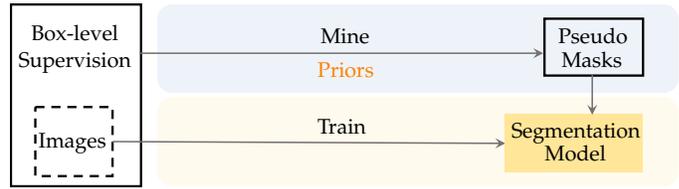
\begin{figure}[t]
	\centering
	\begin{overpic}[width=\linewidth]{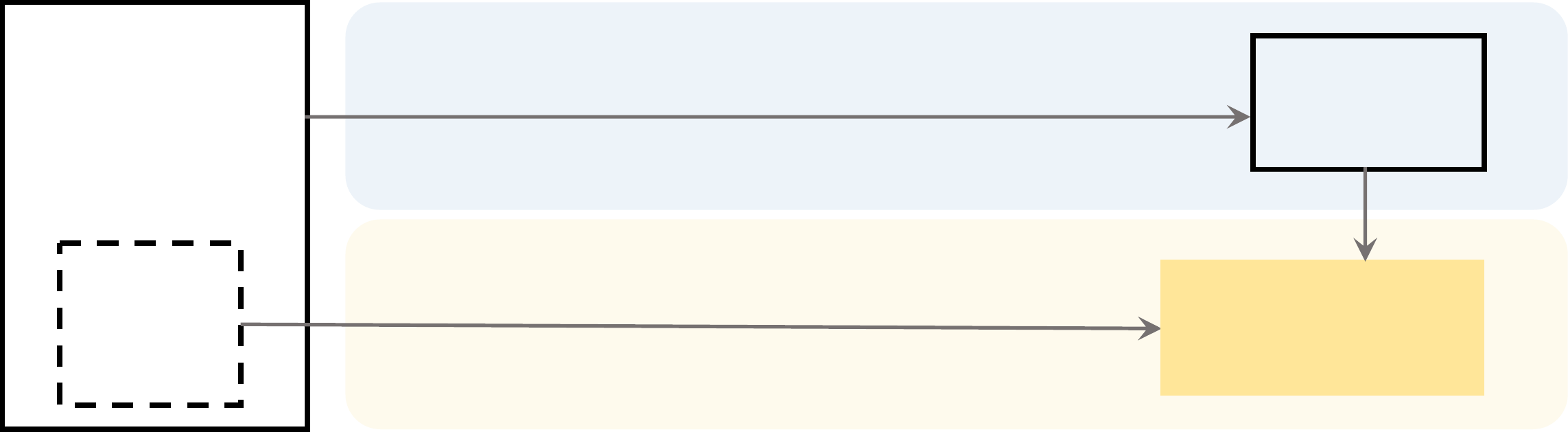}
	\put(4.3,6){\footnotesize{Images}}
	\put(3.2,22){\footnotesize{Box-level}}
	\put(1.3,18){\footnotesize{Supervision}}
	
	\put(75,7.5){\footnotesize{Segmentation}}
	\put(80,4){\footnotesize{Model}}
	\put(82,21.5){\footnotesize{Pseudo}}
	\put(82.5,18){\footnotesize{Masks}}
	
	\put(46.5,21.5){\footnotesize{Mine}}
	\put(46,16.5){\footnotesize{\textcolor{orange}{Priors}}}
	\put(46,8){\footnotesize{Train}}

	\end{overpic}
 \vspace{-5mm}
	\caption{The mainstream pipeline for semantic segmentation with box-level supervision.}
 \vspace{-3mm}
	\label{fig:WSSS_F5_Pipeline1}
\end{figure}

In this section, we review the methods for semantic segmentation with box-level supervision. The box-level supervision serves as a more powerful alternative to the image-level supervision, which naturally narrows down the search space for locating objects. The core challenge in semantic segmentation with box-level supervision shifts to distinguish between foreground objects and background regions inside the annotated bounding boxes. Since annotated bounding boxes play a similar role as CAM-like maps, as shown in ~\cref{fig:WSSS_F5_Pipeline1}, semantic segmentation with box-level supervision also consists of two steps: 1) Mining pseudo masks from the annotated bounding boxes according to \textbf{cross-pixel similarity}; 2) Training a segmentation model based on the pseudo masks.

The first attempt for this task was made by Dai \emph{et al.}~\cite{WSSS_BoxSup_2015_ICCV}. They presented a method to alternatively update pseudo masks and the segmentation model. Specifically, the authors first adopted MCG~\cite{WSSS_MCG_2014_CVPR}, an unsupervised region proposal method, to generate around 2,000 candidate segments per image. Then they repeatedly performed the following three steps: 1) Use a segmentation model to predict the semantic labels for each candidate segment; 2) For each annotated bounding box, from the candidate segments which are predicted as the same semantic label as that of the bounding box, select the one with the largest overlapping region as the pseudo mask for the bounding box; 3) Update the segmentation model by pseudo masks.

Since MCG~\cite{WSSS_MCG_2017_TPAMI} generates candidate segments without considering box-level supervision, the reliability of the pseudo masks generated by MCG is limited. Instead, most recent methods~\cite{WSSS_BDC_2019_CVPR,WSSS_Box2Seg_2020_ECCV,WSSS_BAP_2021_CVPR} regarded the box-level supervision as a noisy starting point to mine the pseudo masks of foreground objects instead.

Song \emph{et al.}~\cite{WSSS_BDC_2019_CVPR} proposed to calculate the filling rate of each class in annotated bounding boxes as a stable guidance to guide segmentation model training. Similarly, Kulharia \emph{et al.}~\cite{WSSS_Box2Seg_2020_ECCV} computed the filling rate at each spatial position inside each bounding box based on a CAM-like map, which can reduce erroneous interests on background regions. Instead of discovering foreground objects in bounding boxes~\cite{WSSS_BDC_2019_CVPR,WSSS_Box2Seg_2020_ECCV}, Oh \emph{et al.}~\cite{WSSS_BAP_2021_CVPR} tried to obtain pseudo masks by removing background regions from annotated bounding boxes. They hypothesized that small patches in background regions from an image are perceptually consistent, which gave a criterion to remove the the background pixels inside each annotated bounding box.

\vspace{-2mm}
\subsubsection{Instance segmentation with box-level supervision}

\begin{figure}[t]
    \centering
    \begin{overpic}[width=\linewidth]{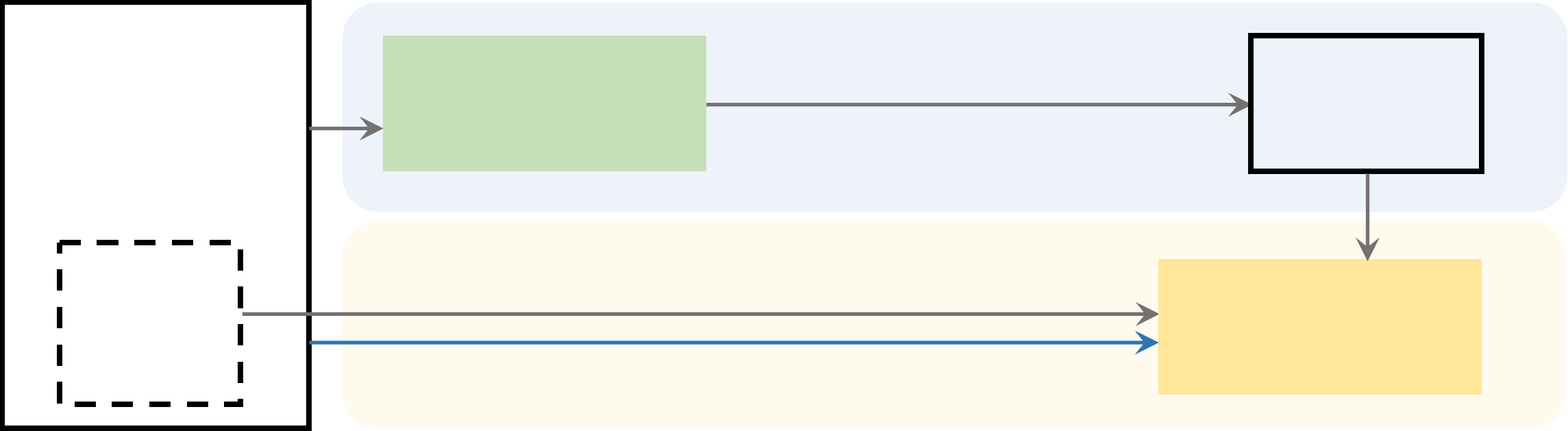}
	\put(4.3,6){\footnotesize{Images}}
	\put(3.2,22){\footnotesize{Box-level}}
	\put(1.3,18){\footnotesize{Supervision}}
	
	\put(75,7.5){\footnotesize{Segmentation}}
	\put(80,4){\footnotesize{Model}}
	\put(82,21.5){\footnotesize{Pseudo}}
	\put(82.5,18){\footnotesize{Masks}}
	\put(28,22){\footnotesize{Detection}}
	\put(30,18.5){\footnotesize{Model}}
	
	\put(57.5,22.5){\footnotesize{Mine}}
	\put(57,17){\footnotesize{\textcolor{orange}{Priors}}}
	\put(43,8){\footnotesize{Train}}
	\put(31,2.5){\footnotesize{End-to-end Train (\textcolor{orange}{Priors})}}

	\end{overpic}
 \vspace{-5mm}
    \caption{The mainstream pipeline for instance segmentation with box-level supervision.}
    \vspace{-5mm}
    \label{fig:psis_arch3}
\end{figure}
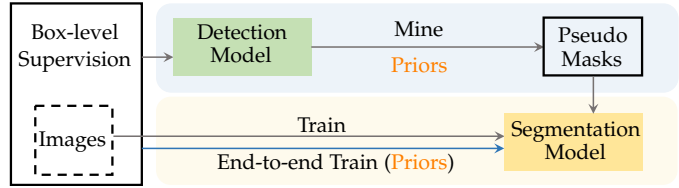

In this section, we review the instance segmentation methods with box-level supervision. 
Instance segmentation with box-level supervision is an easier problem than instance segmentation with image-level supervision, since annotated bounding boxes already provide instance locations for training. The remained difficulty in training is how to perform foreground/background segmentation within a given annotated bounding box. As depicted in~\cref{fig:psis_arch3}, this can be addressed by 1) generating pseudo masks from the annotated bounding box according to \textbf{cross-label constraint}, then performing \textbf{self-training} (the gray flow line in~\cref{fig:psis_arch3}), or 2) \textbf{end-to-end training} with a loss function which directly formulates \textbf{cross-label constraint} (the blue flow line in~\cref{fig:psis_arch3}).

\noindent \emph{4.2.2.1~~~~Mask prediction by self-training}\\
\noindent SDI~\cite{WSIS_SDI_CVPR_2017} is the first deep learning based method to address this problem. For each given annotated box, SDI used the whole box region or the initial segment produced by Grabcut~\cite{YYDS_Grabcut_TOG_2004} within the box as the pseudo mask, and then performed self-training to iteratively refine the pseudo mask and finally deliver mask prediction. Lee \emph{et al.} proposed BBAM~\cite{WSIS_BBAM_CVPR_2021} to employed high-level semantic information from an object detector to produce pseudo masks. They tried to find some learnable areas within a box from which an object detector can predict almost the same detection and classification results as those obtained from the whole box region. Intuitively, these areas represent discriminative parts of an object. The areas of different object proposals corresponding to a given annotated box were combined as its pseudo mask. BoxCaSeg~\cite{WSIS_BoxCaSeg_CVPR_2021} enhanced the ability of foreground/background separation by introducing extra knowledge from a saliency segmentation dataset, leading to more precise pseudo masks.

\noindent \emph{4.2.2.2~~~~Mask prediction by end-to-end training}\\
\noindent BBTP~\cite{WSIS_BBTP_NIPS_2019} and BoxInst~\cite{WSIS_Boxinst_CVPR_2021} are two end-to-end training based instance segmentation methods with box-level supervision. In these two methods, a projection loss was deigned to directly formulate \textbf{cross-label constraint}, which guarantees the consistency between a given annotated box and the projection of a predicted mask along its four sides. Nevertheless, this projection loss cannot impose any constraint to the shape of the predicted mask, which may lead to trivial solutions, such as an all-one rectangle. To solve this issue, extra pairwise loss functions were also provided in BBTP and Boxinst, which defined \textbf{cross-pixel similarity} based on spatial location and color, respectively.

\vspace{-2mm}
\subsubsection{Panoptic segmentation with box-level supervision}
Panoptic segmentation from box-level supervision remains a challenging problem. The only work that attempted to address this problem is WPS~\cite{Li_WSPS_ECCV_2018}. In this work, background stuff was annotated by image-level labels and foreground instances were annotated by box-level labels. WPS~\cite{Li_WSPS_ECCV_2018} first used Grad-CAM~\cite{WSSS_Grad_CAM_ICCV_2017} to obtain the heatmap of foreground and background categories, according to \textbf{cross-label constraint}, then utilized Grabcut~\cite{YYDS_Grabcut_TOG_2004} to locate the pseudo mask of each foreground instance from the heat maps, according to \textbf{cross-pixel similarity}.

\vspace{-3mm}
\subsection{Segmentation with Scribble-level Supervision}
\vspace{-1mm}

\begin{figure}[t]
	\centering
	\begin{overpic}[width=\linewidth]{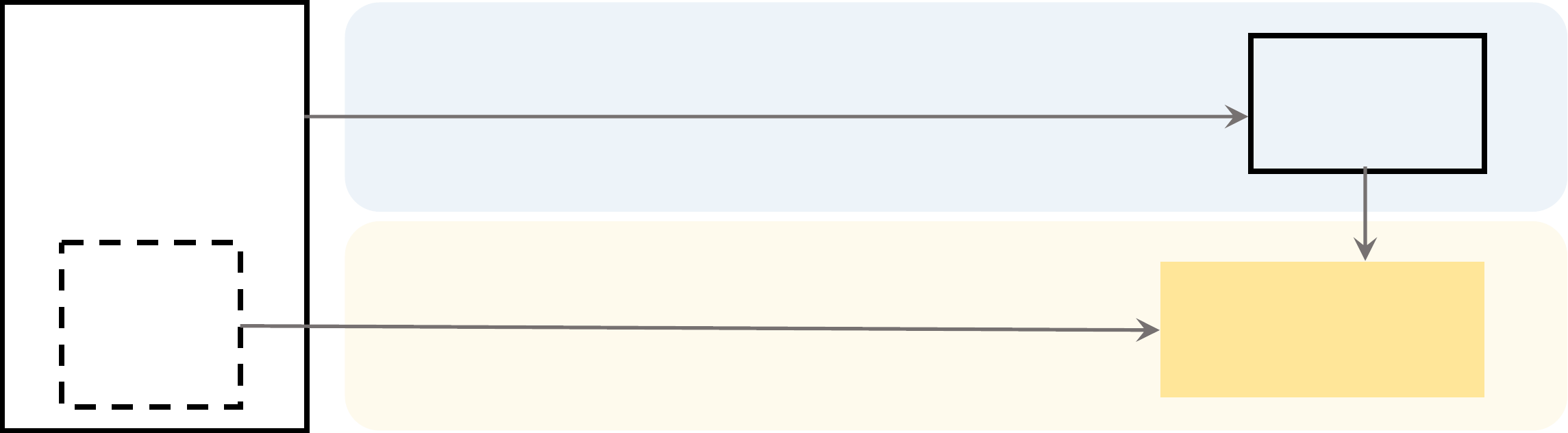}
	\put(4.3,6){\footnotesize{Images}}
	\put(0.3,22){\footnotesize{Scribble-level}}
	\put(1.3,18){\footnotesize{Supervision}}
	
	\put(75,7.5){\footnotesize{Segmentation}}
	\put(80,4){\footnotesize{Model}}
	\put(82,21.5){\footnotesize{Pseudo}}
	\put(82.5,18){\footnotesize{Masks}}
	
	\put(43,21.5){\footnotesize{Propagate}}
	\put(46,16.5){\footnotesize{\textcolor{orange}{Priors}}}
	\put(46,8){\footnotesize{Train}}

	\end{overpic}
 \vspace{-5mm}
	\caption{The mainstream pipeline for semantic segmentation with srcibble-level supervision.}
 \vspace{-5mm}
	\label{fig:WSSS_F5_Pipeline2}
\end{figure}

In this section, we review the methods for scribble-based (semantic) segmentation, where annotations are provided with only a small fraction of pixels, usually as the form of hand-drawn scribbles. The hand-drawn scribbles can be thought as a kind of seed areas.~\cref{fig:WSSS_F5_Pipeline2} shows the mainstream pipeline of semantic segmentation with scribble-level supervision. The crux to address this problem is how to propagate semantic information from the sparse scribbles to all other unlabeled pixels. Current methods achieved this by making use of the internal prior of images, \emph{i.e.}, \textbf{cross-pixel similarity}.

As the first attempt, Di \emph{et al.}~\cite{WSSS_ScribbleSup_2016_CVPR} propagated information from scribbles to unlabeled pixels via a graph model. The graph model was built on the superpixels~\cite{YYDS_superpixel_2003} of an image, where the nodes were superpixels and the edges represented similarities between adjacent nodes, measured by low-level appearance cues, such as color and texture. They jointly optimized the graph model and a segmentation model by an alternated scheme: Fix the segmentation model, the multi-label graph cuts solver~\cite{WSSS_graph_cut_2004_TPAMI} was adopted to assign semantic labels for each unlabeled node in the graph model to form pseudo masks; Based on the pseudo masks, they re-trained the segmentation model. In a similar vein, Xu \emph{et al.}~\cite{WSSS_SSI_2021_ICCV} utilized multi-level semantic features rather than low-level features in the graph model for inferring pseudo masks. Paul \emph{et al.}~\cite{WSSS_RW_2017_CVPR} generated pseudo masks from scribble via a label propagator. The label propagator is a differentiable model based on random walk~\cite{WSSS_RWR_2006_TPAMI} for semantic propagation, which enabled end-to-end joint training with a segmentation model. It was further incorporated with a learnable boundary predictor~\cite{WSSS_HN_B_ICCV_2015} to explicitly constrain the spatial propagation to ``walk'' inside object regions.

The aforementioned methods all require an extra model to get pseudo masks. There are also a few methods which directly optimize the segmentation model by designing a loss function to connect scribbles and dense predictions. Tang \emph{et al.}~\cite{WSSS_NC_2018_CVPR} designed a new loss function, where scribbles were used as partial per-pixel supervision and normalized cut~\cite{YYDS_NCUT_TPAMI00} was adopted to softly ensure the consistency of all pixels. Tang \emph{et al.}~\cite{WSSS_ORL_2018_ECCV} further proposed an extended version which also incorporated CRF into the loss function.

\vspace{-2mm}
\subsection{Discussion}
Segmentation with inexact supervision reduces the requirement for the quality of training images with full dense labels. As summarized in this section, the main pipeline to address this problem is stage-wise: 1) Generate pseudo masks from seed areas by either propagation or mining (\emph{ref.} ~\cref{tab:overview}); 2) Train the segmentation model based on the pseudo masks (self-training). The state-of-the-art results of image segmentation with inexact supervision are comparable to the result of segmentation with full dense supervision, as shown in~\cref{tab:Table_WSSS_performance},~\cref{tab:Table_WSSS_box_performance} and~\cref{tab:Table_WSSS_scribble_performance} in the Appendix. However, CAM based seed areas might be significantly inaccurate for small objects and objects with holes. For these challenging cases, another pipeline, \emph{i.e.}, the end-to-end pipeline to directly link dense predictions to inexact labels (\emph{ref.} ~\cref{tab:overview}), might be an solution and worth further exploring.


\vspace{-2mm}
\section{Incomplete Supervision}
\label{sec:incomplete supervision}
As shown in~\cref{fig:weak_label} and~\cref{tab:math_def}, incomplete supervision can be categorized into 1) semi supervision, 2) domain-specific supervision and 3) partially supervision. Accordingly, segmentation with these three types of weak supervision are called semi-supervised segmentation, domain-adaptive segmentation and partially-supervised segmentation, respectively.

\vspace{-2mm}
\subsection{Semi-supervised Segmentation}
\subsubsection{Semi-supervised semantic segmentation}

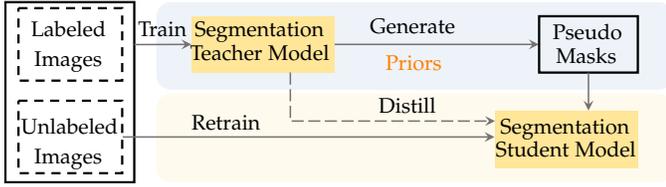
\begin{figure}[t]
	\centering
    \begin{overpic}[width=\linewidth]{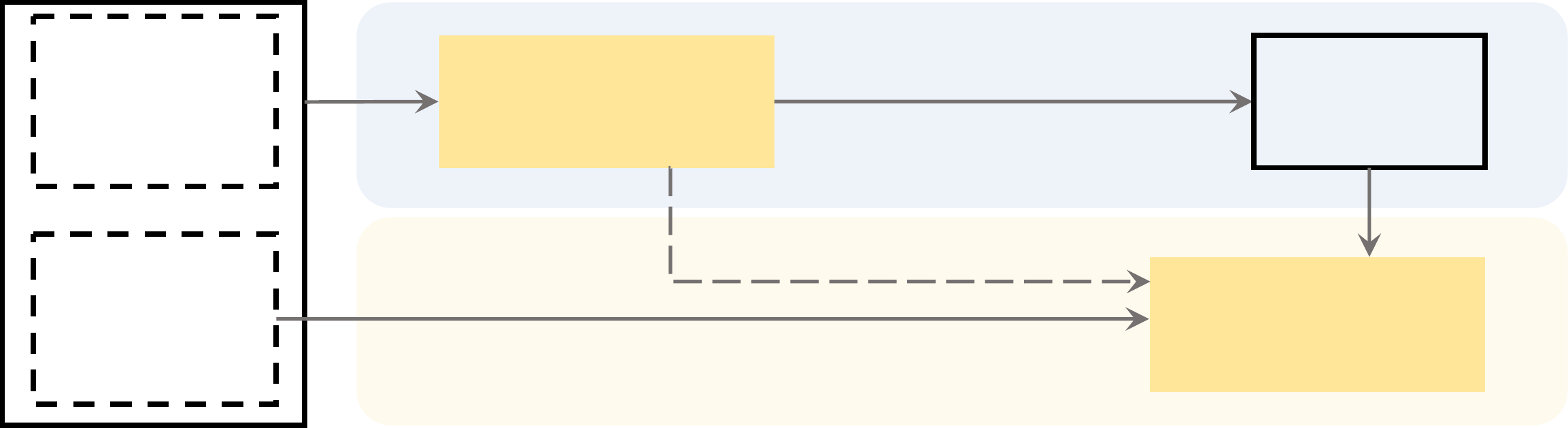}
	\put(4, 22){\footnotesize{Labeled}}
	\put(4.5,17.5){\footnotesize{Images}}
	\put(2.5,7.5){\footnotesize{Unlabeled}}
	\put(4.5,3){\footnotesize{Images}}
	
	\put(74.2,7.5){\footnotesize{Segmentation}}
	\put(73.5,4){\footnotesize{Student Model}}
	\put(82,21.5){\footnotesize{Pseudo}}
	\put(82.5,18){\footnotesize{Masks}}
	\put(28.5,22){\footnotesize{Segmentation}}
	\put(27.8,18.5){\footnotesize{Teacher Model}}
	\put(20,22){\footnotesize{Train}}
	
	\put(54.7,22.5){\footnotesize{Generate}}
	\put(57,17){\footnotesize{\textcolor{orange}{Priors}}}
	\put(28,8){\footnotesize{Retrain}}
	
	\put(56,10.5){\footnotesize{Distill}}

	\end{overpic}
 \vspace{-5mm}
	\caption{The mainstream pipeline for semi-supervised semantic segmentation.}
 \vspace{-5mm}
	\label{fig:SSL_Framework}
\end{figure}

In this section, we review the methods for semi-supervised semantic segmentation, where only a small fraction of training images is provided pixel-level annotations and the rest training images are not annotated.  The objective of semi-supervised semantic segmentation is involving the large number of unlabeled training images into training to improve segmentation performance. A common adopted framework for semi-supervised semantic segmentation is \emph{self-training} shown in ~\cref{fig:SSL_Framework}, which applies the segmentation model trained on labeled training images (teacher model) to unlabeled training images to generate pseudo dense labels (masks), then retrains the segmentation model with the pseudo dense labels (student model). The pseudo dense labels are inevitably noisy, thus current semi-supervised semantic segmentation methods either 1) \textbf{refined} the pseudo dense labels to improve their reliability implicitly according to \textbf{cross-image relation} or 2) \textbf{regularized} them by introducing extra supervisions explicitly according to \textbf{cross-view consistency}. 

\noindent \emph{5.1.1.1~~~~Pseudo label refinement for self-training}\\ 
Intuitively, the reliability of pseudo dense labels can be determined by their confidences provided by the segmentation model. Existing methods improved the reliability of pseudo dense labels by refining them with iterative self-training or by neglecting those with less confidences. 

Hung \emph{et al.}~\cite{SSL_Hung2018} made use of a discriminator network to generate reliable confidence maps for unlabeled images. The discriminator network was trained with labeled images with the ability to determine whether the input is from ground-truth dense labels or predictions from the segmentation model. Ke \emph{et al.}~\cite{SSL_SelfTraining} proposed a three-stage self-training framework to refine pseudo labels in a stage-wise manner. They modified the segmentation model by adding an auxiliary branch which was the duplicate of the last two blocks of the original model. The last two blocks of the original branch and the auxiliary branch were trained by the unlabeled data with the pseudo labels and the labeled data, respectively. Since the auxiliary branch was trained by only the labeled data, it can generate more reliable pseudo labels for the next stage. This stage-wise self-training framework iteratively improved the reliability of pseudo labels, and thus leaded to performance improvements.

He \emph{et al.} \cite{SSL_DARS} proposed a quality-control policy for self-training, where a labeling ratio $\alpha$ was introduced to control the quality of pseudo labels so that only $\alpha\%$ of pixels in an unlabeled image retain corresponding pseudo labels. As the labeling ratio was determined by a category-specific confidence threshold, He's method can address the problem of long-tailed data distribution in semi-supervised semantic segmentation.

\noindent \emph{5.1.1.2~~~~Pseudo label regularization by cross-view consistency}\\
Pseudo label regularization can benefit from unsupervised dense representation learning (\cref{sec:no supervision}), since they both aim at training segmentation models on unlabeled images. Thus, the Siamese structure and contrastive learning are also used in semi-supervised semantic segmentation to ensure \textbf{cross-view consistency} among pseudo dense labels of the same image under different views.  

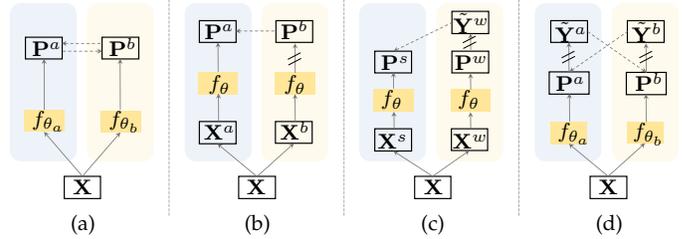
\begin{figure}[t]\footnotesize%
\centering
\begin{overpic}[width=\linewidth]{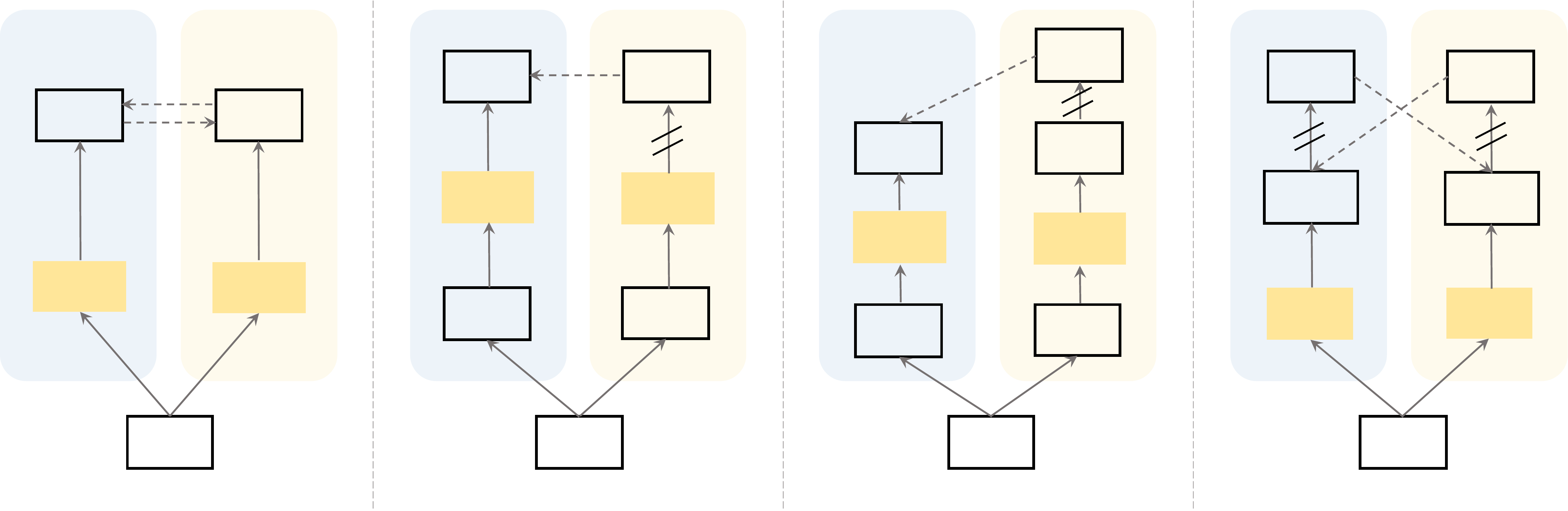}
\put(9.1,-2.){(a)}
\put(35.1,-2.){(b)}
\put(61.5,-2.){(c)}
\put(87.6,-2.){(d)}

\put(9.4, 3.3){$\mathbf{X}$}
\put(35.4, 3.3){$\mathbf{X}$}
\put(61.9, 3.3){$\mathbf{X}$}
\put(88., 3.3){$\mathbf{X}$}

\put(3.3, 13.7){$f_{\theta_{a}}$}
\put(14.8, 13.7){$f_{\theta_{b}}$}
\put(3.3, 24.){$\mathbf{P}^a$}
\put(14.8, 24.){$\mathbf{P}^b$}

\put(29., 11.5){$\mathbf{X}^a$}
\put(40.4, 11.5){$\mathbf{X}^b$}
\put(29.7, 19.){$f_{\theta}$}
\put(41., 19.){$f_{\theta}$}
\put(29.1, 26.6){$\mathbf{P}^a$}
\put(40.5, 26.6){$\mathbf{P}^b$}

\put(55.0, 10.2){$\mathbf{X}^s$}
\put(66.2, 10.2){$\mathbf{X}^w$}
\put(55.5, 16.7){$f_{\theta}$}
\put(67., 16.7){$f_{\theta}$}
\put(55.5, 21.9){$\mathbf{P}^s$}
\put(66.5, 21.9){$\mathbf{P}^w$}
\put(66.3, 27.5){$\tilde{\mathbf{Y}}^w$}

\put(81.9, 11.8){$f_{\theta_{a}}$}
\put(93.2, 11.8){$f_{\theta_{b}}$}
\put(81.6, 18.8){$\mathbf{P}^a$}
\put(93.5, 18.8){$\mathbf{P}^b$}
\put(81.5, 26.2){$\tilde{\mathbf{Y}}^a$}
\put(93., 26.2){$\tilde{\mathbf{Y}}^b$}

\end{overpic}
\vspace{-5mm}
\caption{Semi-supervised semantic segmentation by Siamese structures. 
(a) GCT~\cite{SSL_GCT},
(b) CutMix-Seg~\cite{SSL_CutMix-Seg},
(c) PseudoSeg~\cite{SSL_PseudoSeg}
and (d) CPS~\cite{SSL_CPS}.
  `$\rightarrow$' means forward operation and `$\dashrightarrow$' means loss supervision.
`$//$' on `$\rightarrow$' means stop-gradient.}
\vspace{-5mm}
\label{fig:semi_structure}
\end{figure}

{\noindent\textbf{Siamese structure based.}} ~\cref{fig:semi_structure} illustrates several typical Siamese structures for pseudo label regularization in semi-supervised semantic segmentation. GCT~\cite{SSL_GCT} utilized two segmentation networks that shared the same architecture but were initialized differently to compute two segmentation probability maps from two different views of an unlabeled image, respectively. The pair of segmentation probability maps were kept consistent as the extra supervision for training. CutMix-Seg~\cite{SSL_CutMix-Seg} also utilized two segmentation networks with the same architecture, but the parameters of one network were the moving average of the other's. The two segmentation probability maps outputted from the two networks were kept consistent for training. PseudoSeg~\cite{SSL_PseudoSeg} used the pseudo dense labels generated from a view with weak augmentation to supervise the pseudo dense labels generated from a view with strong augmentation.  CPS~\cite{SSL_CPS} followed the strategy to utilize two differently-initialized segmentation networks with the same architecture and enforced the consistency between pseudo dense labels outputted from them. Their experimental results showed that their method can achieve better segmentation performance.

{\noindent\textbf{Contrastive learning based.}}
Zhong \emph{et al.}~\cite{SSL_PCCSeg} applied pixel-wise contrastive learning to facilitating the feature learning of intermediate layers. For a query pixel, they investigated several sampling strategies to select negative keys (pixels) in pixel-wise contrastive learning, including 1) Uniform: pixels at different locations in one image or from different images are negative keys; 2) Different image: pixels from different images are negative keys; 3) Uniform + Pseudo Label: pixels at different locations in one image or from different images with low confident pseudo labels are negative keys; 4) Different image + Pseudo Label: pixels from from different images with low confident pseudo labels are negative keys. Lai \emph{et al.}~\cite{SSL_CAC} proposed a novel strategy to generate different views of an unlabeled image by considering contextual information. For an unlabeled image, two different patches were randomly cropped with an overlapping region. The feature maps of the two patches were computed by an encoder and a non-linear projector. Then contrastive learning was applied to ensuring that the feature representations of the overlapping region computed under different contexts are consistent. The feature representations at the same pixel were taken as a positive pair while the feature representations in the remaining regions formed the negative samples.

\vspace{-2mm}
\subsubsection{Semi-supervised Instance segmentation}

\begin{figure}
    \centering
    \begin{overpic}[width=\linewidth]{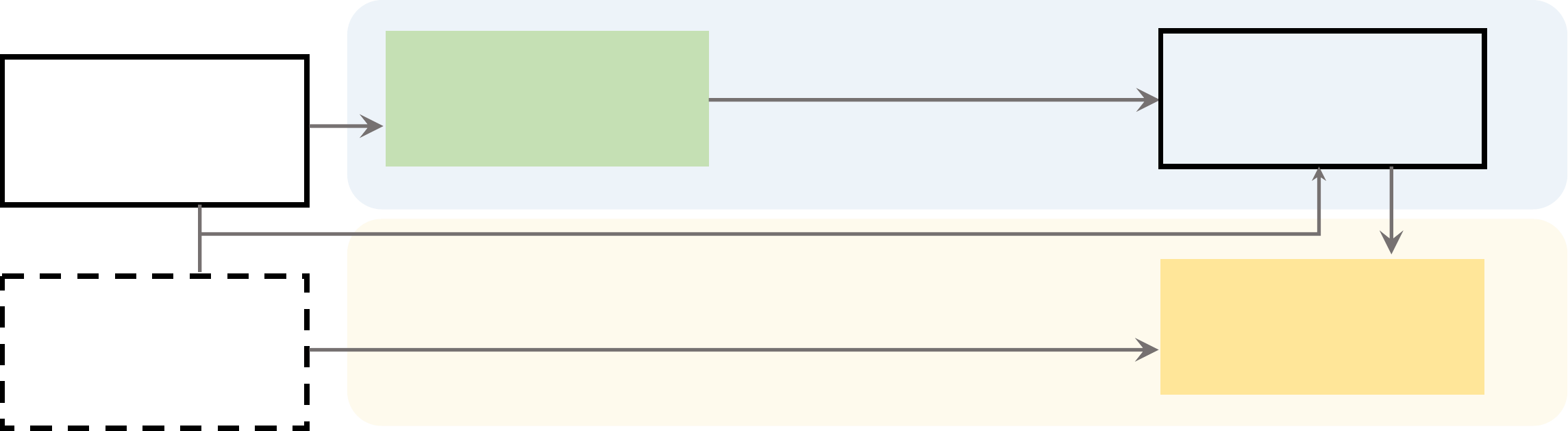}
	\put(1.2,21.5){\footnotesize{Images with}}
	\put(3,18.5){\footnotesize{Box-level}}
	\put(1.6,15.5){\footnotesize{Supervision}}
	\put(1.2,7.2){\footnotesize{Images with}}
	\put(2.2,4.2){\footnotesize{Mask-level}}
	\put(1.6,1.2){\footnotesize{Supervision}}
	
	\put(74.5,7.5){\footnotesize{Segmentation}}
	\put(79.5,4){\footnotesize{Model}}
	\put(77,22){\footnotesize{Auxiliary}}
	\put(75.5,18.5){\footnotesize{Information}}
	\put(27.7,22){\footnotesize{Detection}}
	\put(30.2,18.5){\footnotesize{Model}}
	
	\put(56.1,22.5){\footnotesize{Export}}
	\put(56.3,18){\footnotesize{\textcolor{orange}{Priors}}}
	\put(53,6.5){\footnotesize{Train}}
	\put(40,14){\footnotesize{Generate}}
	\put(42,9.5){\footnotesize{\textcolor{orange}{Priors}}}
	\put(90,13.5){\footnotesize{Assist}}
	
	\put(0.5,11){\footnotesize{\emph{base}}}
	\put(0.5,25.3){\footnotesize{\emph{base + novel}}}

	\end{overpic}
 \vspace{-5mm}
    \caption{The mainstream pipeline for semi-supervised instance segmentation.}
    \vspace{-5mm}
    \label{fig:psis_arch}
\end{figure}
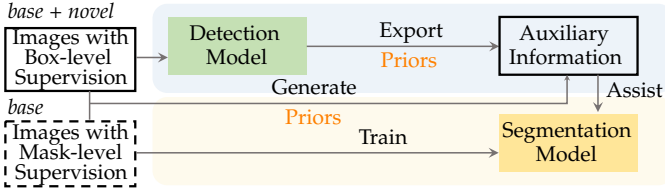

The setting of semi-supervised instance segmentation is slightly different to the standard one: In the literature, under this setting, object categories are divided into two disjoint splits: \emph{base} and \emph{novel}, where both of the two splits are provided by weak box-level annotations, but only the \emph{base} categories are provided by per-pixel annotations. Thus, semi-supervised instance segmentation is also called partially-supervised instance segmentation. Formally, following the definition in Table~\ref{tab:math_def}, the training set of partially-supervised instance segmentation is
\begin{align}
    \mathcal{T}=&\{(\mathbf{X}^{(n)},\mathbf{Y}^{(n)})|n\in\mathcal{N}_l\}\bigcup\{(\mathbf{X}^{(n)},\mathcal{B}^{(n)})|n\in\mathcal{N} \backslash \mathcal{N}_l\}, \nonumber \\ &\texttt{\textbf{s.t.}} \forall n\in\mathcal{N}_l, m\in\mathcal{N} \backslash\mathcal{N}_l, \mathcal{C}^{(n)}\bigcap\mathcal{C}^{(m)}=\emptyset.
\end{align}
Intuitively, the difficulty of this task lies in the supervision gap between box-level annotations and pixel-level dense predictions on the \emph{novel} categories.
As shown in~\cref{fig:psis_arch}, existing methods mainly follow a detection-then-segmentation pipeline, \emph{e.g.}, Mask R-CNN~\cite{maskrcnn}, and explore how to extract auxiliary information from a detection model by utilizing the priors we have summarized to assist the learning of a segmentation model on the \emph{novel} categories. 

\noindent \emph{5.1.2.1~~~~Auxiliary information from cross-label constraint} 

From the box-level annotations, two types of auxiliary information for segmentation model training can be extracted from the prior of \textbf{cross-label constraint}. One is the connection between box category labels and segmentation masks, as explored in CAMs~\cite{WSSS_CAM_2016_CVPR, WSSS_Grad_CAM_ICCV_2017}; The other is the connection between box location labels and segmentation masks~\cite{PSIS_Opmask_ICCV_2021}, since the segmentation mask for an object is tightly enclosed by its bounding box.

Mask\textsuperscript{X} RCNN~\cite{PSIS_MaskXRCNN_CVPR_2018}, built upon Mask RCNN, is the first partially-supervised instance segmentation method. Mask\textsuperscript{X} RCNN was motivated by the label connection between box-level classification and per-pixel classification (segmentation) within the box. It learned a category-agnostic function to transfer parameters of the detection head to the parameters of the segmentation head. Once this function was learned on base categories, it can be used to generate the parameters of the mask head for novel categories. In OPMask~\cite{PSIS_Opmask_ICCV_2021}, Biertimpel \emph{et al.} produced a CAM-like map within each box from a box-level annotation, where each channel represented an object mask prior (OMP) for a specified category. This CAM-like map was then applied to enhance the features for training the segmentation head.

\noindent \emph{5.1.2.2~~~~Auxiliary information from cross-pixel similarity}\\
\noindent
An important goal of the partially-supervised setting is to explore class-agnostic commonalities between base and novel categories, which can be utilized to improve the feature discrimination ability for novel categories. Exploiting the prior of cross-pixel similarity from low-level (color, texture) or high-level (semantic relationship, affinity) information is a good strategy to approach this goal.

Zhou \emph{et al.} proposed Shapeprop~\cite{PSIS_Shapeprop_CVPR_2020} to produce class-agnostic shape activation maps, \emph{i.e.}, more structured and finer CAMs, as the auxiliary commonality information. 
They employed multiple instance learning to locate a salient region within each given bounding box in a class-agnostic manner for all categories.
Then they designed a saliency propagation module to expand the salient region to cover the whole object within each given bounding box, forming the shape activation map.
In CPMask~\cite{PSIS_CPMask_ECCV_2020}, Fan \emph{et al.} explored class-agnostic shape cues, which were extracted by boundary prediction and non-local attention based pixel-to-pixel affinity learning.

Although the authors of Shapeprop~\cite{PSIS_Shapeprop_CVPR_2020} and CPMask~\cite{PSIS_CPMask_ECCV_2020} claimed their auxiliary commonality information is class-agnostic, they extracted the information only from base categories, which leaded to a misalignment problem of the features between base and novel categories. Wang \emph{et al.} presented ContraskMask~\cite{PSIS_ContrastMask_CVPR_2022} which was built upon OPMask and addressed this issue by introducing an extra unified pixel-level contrastive learning framework. In this framework, all images were used to train an extra class-agnostic encoder through a unified pixel-level contrastive loss and an elaborated query-keys sampling strategy according to \textbf{cross-pixel similarity}. The encoder provided aligned and distinctive encoded features for all categories, facilitating the segmentation on novel categories.

\noindent \emph{5.1.2.3~~~~Auxiliary information from cross-image relation} 

Kuo \emph{et al.} proposed ShapeMask~\cite{PSIS_Shapemask_ICCV_2019} to address partially-supervised instance segmentation by exploring common shape priors from \textbf{cross-image relations}, since objects from similar categories in different images should have similar coarse shapes. The shape priors, obtained by performing clustering on mask-level annotations of all training images from base categories, can be linearly assembled and then generalized to diverse categories to assist the segmentation head to progressively refine predicted segmentation masks.

\noindent \emph{5.1.2.4~~~~Auxiliary information from a larger segmentation model} 

In Deep-MAC~\cite{PSIS_DeepMAC_ICCV_2021}, Birodkar \emph{et al.} investigated partially-supervised instance segmentation from a new perspective, \emph{i.e.}, the capacity of the segmentation head. By finding that a much stronger segmentation head can smooth over the gap caused by the missing supervision of novel categories, Deep-MAC replaced the original segmentation head in ~\cite{maskrcnn} with a much stronger model, \emph{e.g.}, Hourglass-100, leading to significant performance improvement.

\vspace{-2mm}
\subsection{Domain-adaptive Segmentation}
\subsubsection{Domain Adaptive Semantic Segmentation}

\begin{figure}[t]
	\centering
    \begin{overpic}[width=\linewidth]{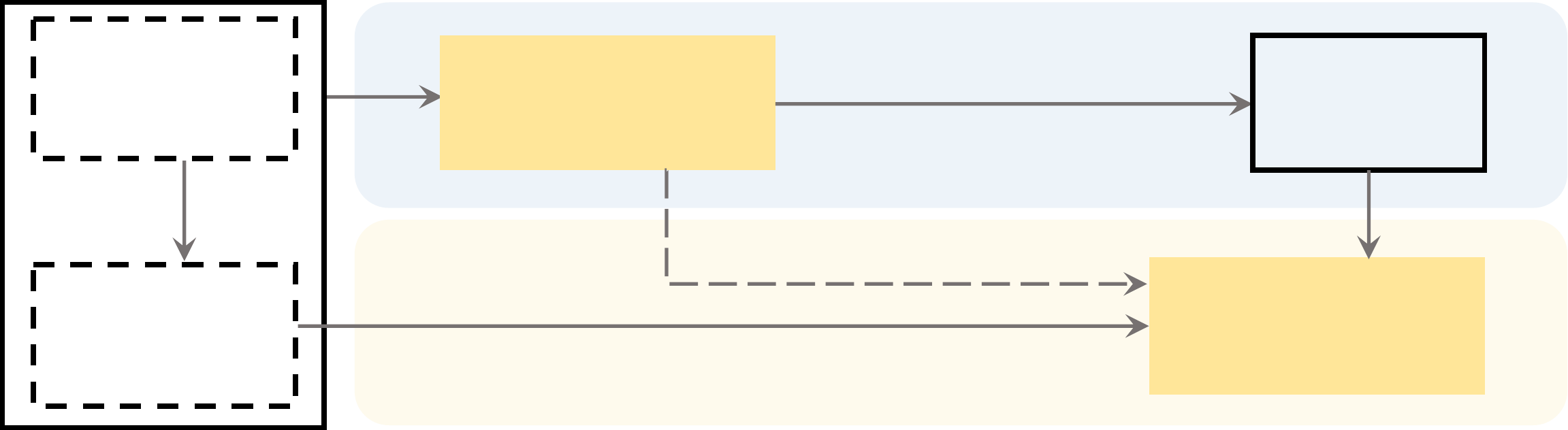}
	\put(5.7,22.7){\footnotesize{Source}}
	\put(5.5,19.2){\footnotesize{Images}}
	\put(6.2,7){\footnotesize{Target}}
	\put(5.5,3.5){\footnotesize{Images}}
	\put(0,13.4){\footnotesize{Transfer}}
	\put(12,13.4){\footnotesize{\textcolor{orange}{Priors}}}
	
	\put(74.2,7.5){\footnotesize{Segmentation}}
	\put(73.5,4){\footnotesize{Student Model}}
	\put(82,21.5){\footnotesize{Pseudo}}
	\put(82.5,18){\footnotesize{Masks}}
	\put(28.8,22){\footnotesize{Segmentation}}
	\put(28.1,18.5){\footnotesize{Teacher Model}}
	\put(20.6,22){\footnotesize{Train}}
	
	\put(54.7,22.5){\footnotesize{Generate}}
	\put(57,17){\footnotesize{\textcolor{orange}{Priors}}}
	\put(28,8){\footnotesize{Retrain}}
	
	\put(56,10.5){\footnotesize{Distill}}

	\end{overpic}
 \vspace{-5mm}
	\caption{The mainstream pipeline for domain adaptive semantic segmentation.}
 \vspace{-5mm}
	\label{fig:UDA_Framework}
\end{figure}
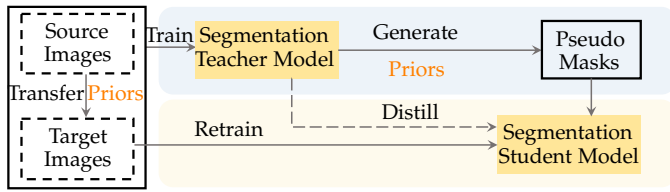

Here, we focus on the scenario where only images on the source domain have pixel-wise annotations and there exists a domain gap between the source domain and the target domain. The goal is to train a segmentation model using source domain data which can be generalized to the target domain. Domain adaptive semantic segmentation is essentially similar to semi-supervised semantic segmentation, where the only difference is whether there is a domain gap between the labeled images and unlabeled images. Thus, as shown in~\cref{fig:UDA_Framework}, the mainstream pipeline of domain adaptive semantic segmentation contains one extra step, compared with the mainstream pipeline of semi-supervised semantic segmentation (\cref{fig:SSL_Framework}): Narrowing down the domain gap. This extra step can be achieved by \textbf{adversarial learning} to map both the source and target domains into the same space~\cite{UDA_AdaptSeg, UDA_I2IAdapt, UDA_BDL, UDA_DPL}, augmentation based \textbf{domain mixing}~\cite{UDA_DACS,UDA_DAFormer} or \textbf{pseudo mask quality improvement} on the target domain~\cite{UDA_CBST, UDA_Kim, UDA_ProDA, UDA_TPLD, UDA_BAPANet}. 

\noindent \emph{5.2.1.1~~~~Adaptation by adversarial learning}\\
\noindent Adversarial learning is used to align source domain images and target domain images in either the image space or the feature space, which is usually achieved by Generative Adversarial Networks (GANs)~\cite{YYDS_GAN_NIPS14}.

Murez \emph{et al.}~\cite{UDA_I2IAdapt} proposed an unpaired image-to-image translation framework to find a joint latent embedding space, where domain-agnostic feature representations can be extracted. To preserve core information and eliminate the structured noise in a specific domain, the authors reconstructed each image by an identity loss and classify whether the features in the latent space are generated from source or target domain by an adversarial loss. Tsai \emph{et al.}~\cite{UDA_AdaptSeg} trained a discriminator to determine whether an output of the segmentation model was from the source domain or the target domain. By fooling the discriminator, the gap between the two domains was shorten. Vu \emph{et al.}~\cite{UDA_advent} reduced the domain gap by additionally encouraging the alignment of entropy distributions between predictions on source images and target images.


Li \emph{et al.}~\cite{UDA_BDL} directly transferred the source images to the target domain by GANs. They proposed a bidirectional learning framework which consisted of an image-to-image translation subnetwork and a segmentation adaptation subnetwork. In the forward direction, the translation subnetwork was trained with an adversarial loss to translate source domain images to the target domain, and the adaptive segmentation subnetwork was trained on the translated source domain images with corresponding ground-truth dense labels as well as the target domain images with no labels. The backward direction ensured cross-view (domain) consistency by the GAN loss, reconstruction loss and perceptual loss. Based on~\cite{UDA_BDL}, Cheng \emph{et al.}~\cite{UDA_DPL} proposed DPL, which made use of two complementary and interactive bidirectional learning frameworks to improve the translation quality from the source domain to the target domain.

\noindent \emph{5.2.1.2~~~~Adaptation by domain mixing}\\
\noindent Another strategy to narrow down the domain gap is mixing images from different domains by mix-up based copy-paste~\cite{YYDS_MIXUP_ICLR18,YYDS_CUTMIX_ICCV19}.

Tranheden \emph{et al.}~\cite{UDA_DACS} mixed the source and target domain images with corresponding ground-truth dense labels and pseudo dense labels respectively by pasting pixels of certain categories from a source domain image to a target domain image. The segmentation model was then trained on these mixed images with mixed ground-truth dense labels and pseudo dense labels. Based on~\cite{UDA_DACS}, Hoyer \emph{et al.}~\cite{UDA_DAFormer} introduced the transformer architecture into domain adaptive semantic segmentation instead of the traditional Deeplab architecture~\cite{deeplabv3} and significantly improved the state-of-the-art performance.

\noindent \emph{5.2.1.3~~~~Adaptation by pseudo mask quality improvement}\\
\noindent Due to the domain gap, the pseudo masks generated on target domain images are usually very noisy. Their quality should be improved before being used for training the student segmentation model. This pseudo mask quality improvement process usually involves some priors, such as \textbf{cross-pixel similarity} and \textbf{cross-view consistency}.

Zou \emph{et al.}~\cite{UDA_CBST} firstly brought the framework of self-training to domain adaptive semantic segmentation. In order to generate high-quality pseudo labels, pixels with higher confidence scores were more likely to be selected to train the student segmentation model. To get rid of negative influence of large classes, the class-wise confidence was normalized. Shin \emph{et al.}~\cite{UDA_TPLD} generated pseudo masks by label propagation from pseudo labels with high confidences, which was based on the assumption that nearby pixels tend to be similar.

Zhang \emph{et al.}~\cite{UDA_ProDA} made use of representative prototypes, \emph{i.e.}, the feature centroid of each class, to denoise the pseudo masks. The prototypes were initialized according to the generated pseudo labels for target domain images. Then they were dynamically updated by the moving average of corresponding cluster centroids in the current mini-batch. Pseudo mask denoising was performed according to prototype assignment, \emph{i.e.}, the probability of assigning a pseudo class label to a pixel was adjusted according to the similarity between the features of the pixel and the prototype of the class. In order to guarantee the robustness of prototype assignment, the prototype assignments for each pixel under weak and strong augmentations were enforced to be consistent.


\vspace{-2mm}
\subsection{Discussion}
Segmentation with incomplete supervision reduces the requirement for the quantity of training images with full dense labels. As summarized in this section, the strategies to address this problem include two main directions (\emph{ref.} ~\cref{tab:overview}): 1) Transfer the segmentation model trained on labeled data to unlabeled data; 2) Generate dense self-supervision on unlabeled data, similar to the strategies used for unsupervised segmentation. As shown in~\cref{tab:semi_res} in the Appendix, the result of the state-of-the-art semi-supervised semantic segmentation method ($75.9\%$ mIoU) is comparable to the result of fully-supervised semantic segmentation model ($76.0\%$ mIoU). However, when there is a large distribution shift between labeled data and unlabeled data, \emph{e.g.}, the domain gap in domain-adaptive semantic segmentation and the non-overlapping between based and novel categories in partially-supervised instance segmentation, segmentation with incomplete supervision suffers from more severe performance degradation (\emph{ref.}~\cref{tab:uda_res} and~\cref{tab:pis_res} in the Appendix). Consequently, how to design more effective strategies to dead with the distribution shift to enable fully making use of unlabeled data in training needs further exploration.

\vspace{-3mm}
\section{Inaccurate Supervision}
\label{sec:noisy supervision}

\vspace{-1mm}
\subsection{Semantic segmentation from Noisy supervision}
\label{sec:semantic segmentation from noisy labels}

Label inaccuracy is commonly existed in segmentation annotations. Given that pseudo masks inevitably involve somewhat noises, training segmentation models from pseudo masks is essentially a noisy semantic segmentation problem, as pointed out in~\cite{WSSS_AEL_2022_CVPR}. 

Liu \emph{et al.}~\cite{WSSS_AEL_2022_CVPR} observed a phenomenon that the segmentation models tend to memorize
the errors in annotations as training proceeds. To prevent over-fitting to the errors, they designed an adaptive early stop mechanism and enforced multi-scale \textbf{cross-view consistency} to boost robustness against annotation errors. Li \emph{et al.}~\cite{WSSS_UE_2021_AAAI} proposed to discover noisy labels via uncertainty estimation~\cite{WSSS_UDWN_2017_NIPS}, which was realized by calculating the pixel-wise variance among prediction maps under different scales, according to \textbf{cross-view consistency}. Larsson \emph{et al.}~\cite{WSSS_cross-season_2019_CVPR} proposed to boost segmentation performance with inaccurate labels by \textbf{cross-view consistency}, which is realized by enforcing label consistency among a series of 2D-2D point matches between two views obtained under different seasons.

\vspace{-3mm}
\section{Conclusion and Discussion}
\label{sec:conclusion and discussion}

Label-efficient image segmentation has become an active topic in computer vision, as it paves the way to addressing real world applications, where per-pixel annotations are notoriously difficult to obtain. As summarized in this survey, a large number of label-efficient image segmentation methods have been proposed in recent years, which addressed segmentation with weak supervision of different types, \emph{i.e.}, no supervision, inexact supervision, incomplete supervision and inaccurate supervision. As described in this survey, these methods are highly related, not only because the problems they were designed to address are related, \emph{e.g.}, segmentation with inaccurate supervision can be a sub-problem of segmentation with inexact supervision, but also because they made use of similar strategies to bridge the supervision gaps between dense prediction and weak supervision. Experimental results showed that these label-efficient image segmentation methods have achieved considerable progress. However, there is large room for improvement to approach the upper bound performance under the fully-supervised setting, \emph{i.e.}, using full dense labels for training. Next, we discuss the challenges need to be resolved and share our opinions about future prospects.

\vspace{-2mm}
\subsection{Challenges}

\noindent\textbf{1) The supervision gap} 
\\As we argued in this paper, the main challenge of label-efficient image segmentation is the supervision gap between dense predictions and incomplete labels. Although a lot of strategies have been proposed to tackle this challenge, as summarized in this survey, how to bridge the supervision gap is still an unresolved open issue. In addition, existing label-efficient image segmentation models are limited in their ability to scale to large numbers of object classes. To address these challenge, more efforts need to be made, including adopting more powerful network backbones and introducing extra supervision from other modalities, such as text supervision.

\noindent\textbf{2) The openness issue} 
\\The label-efficient segmentation problem is closely related to open-domain (or open-vocabulary) recognition, where new concepts can be described by texts, few examples, \textit{etc}. In such scenarios, an important issue lies in dealing with the openness of recognition, in particular, how to design a pipeline for adding new concepts into an existing recognition system? Simply relying on text guidance (\textit{e.g.}, leveraging text embeddings from a pre-trained model) may be insufficient, yet searching and distilling knowledge from web data is a promising solution.

\noindent\textbf{3) Granularity vs. consistency} 
\\Label-efficient image segmentation aims to cover more visual concepts. However, as the number of concepts goes up, there is a tradeoff between recognition granularity and consistency. That said, when fine-grained classes and/or object parts are added to the dictionary, it is possible that the algorithm may not produce consistent recognition results, \textit{e.g.}, when the object is small, the algorithm may choose to predict coarse-grained labels and/or parts -- it is good to adjust the evaluation protocol for such scenarios.

\vspace{-2mm}
\subsection{Potential Directions}


\noindent\textbf{1) Zero-shot segmentation with text supervision}\\
There are a large number of images with a wide variety of text supervision available on the Internet, which enables learning large-scale models, such as CLIP~\cite{clip}, to bridge visual representations and text supervision. Such pre-trained models shed light on a new way to perform zero-shot image classification~\cite{coop} as well as semantic segmentation, \emph{i.e.}, learning segmentation models for unseen categories~\cite{zs3,cagnet,ldss,pmzs3,groupvit,zegformer}. Generally, these works generalized segmentation models to unseen categories by aligning visual features to the text embedding of the corresponding semantic class. This is a promising direction worth exploring, as it has the potential of working in an open domain and generalizing to the scenario with an unlimited number of categories. However, existing `zero-shot' segmentation methods mostly relied on having seen the queried or similar class(es) in the pre-training stage, which is not \textit{really} `zero-shot'. We look forward to new evaluation metrics and/or benchmarks that focus on the open domain issue itself without caring too much about whether the test is `zero-shot'.

\noindent\textbf{2) Label-efficient segmentation by vision transformers}\\
The existing label-efficient segmentation methods are mainly based on convolutional neural networks. While recent years have witnessed a revolution in computer vision brought by the transformer module~\cite{YYDS_Attention_NIPS17}. The emergence of vision transformer (ViT)~\cite{YYDS_VIT_ICLR21} and its variants~\cite{YYDS_MSVIT_ICCV21,YYDS_T2T_ICCV21,YYDS_TIT_NIPS21,YYDS_GG_NIPS21} made breakthroughs in various vision tasks, such as segmentation~\cite{YYDS_SETR_ICCV21,YYDS_SEGMENTOR_ICCV21,YYDS_SEGFORMER_NIPS21,YYDS_HRFormer_2021_NIPS}. In addition, it has been observed that the self-attention maps of vision transformers pre-trained by advanced unsupervised representation learning methods, such as DINO~\cite{YYDS_DINO_ICCV21}, BEiT~\cite{YYDS_BEIT_ICLR22}, MAE~\cite{YYDS_MAE_CVPR22} and iBoT~\cite{YYDS_IBOT_ICLR22}, contain rich information about the segmentation of an image, which provides a potential way to generate reliable pseudo dense labels without any supervision. We believe that exploring the usage of the unsupervised self-attention maps of vision transformers in pseudo dense label generation for label-efficient segmentation is an interesting and open future direction. Some recent studies have shown that the ability of vision transformers in modeling long-range dependency can benefit label-efficient segmentation, \emph{e.g.}, MCTformer~\cite{WSSS_MCT_2022_CVPR} replaced CNNs with ViTs to capture more complete seed areas for higher accuracy in weakly-supervised semantic segmentation, and CLIMS~\cite{WSSS_CLIMS_2022_CVPR} inherited extra knowledge from CLIP~\cite{clip} to build a cross language-image mapping that benefits weakly-supervised semantic segmentation. In addition, STEGO~\cite{con_stego}/DAformer~\cite{UDA_DAFormer} validated a similar story on unsupervised/transfer semantic segmentation, respectively. We expect that, when the emerging properties from self-supervised pre-training are integrated, the weakly-supervised learning algorithms can be enhanced.

\noindent\textbf{3) Unexplored label-efficient segmentation problems}\\
As shown in~\cref{fig:taxonomy}, there are some unexplored label-efficient segmentation problems, such as instance segmentation from noisy labels and panoptic segmentation from incomplete labels. The reason why these problems are not explored yet might be that there lack of proper datasets for evaluation or sufficiently sophisticated models to achieve reasonable results. With the development of label-efficient segmentation techniques, these research gaps will be filled in the future.

\noindent\textbf{4) Mixing various annotations for lifelong learning}\\
In real-world applications, we do not expect computer vision algorithms to work in a fixed domain or dataset, but assume that new classes with various types of annotations can appear at any time. This setting is known as lifelong learning where multiple difficulties are combined, including incremental learning, domain transfer, and imperfect labels. We look forward to integrating the settings surveyed in this paper into a unified framework in which various annotations appear as different types of queries. A benchmark is strongly required for this research direction.

\ifCLASSOPTIONcompsoc
  \vspace{-2mm}
  \section*{Acknowledgments}
\else
  \vspace{-2mm}
  \section*{Acknowledgment}
\fi

This work was supported by NSFC 62176159, Natural Science Foundation of Shanghai 21ZR1432200 and Shanghai Municipal Science and Technology Major Project 2021SHZDZX0102.
\vspace{-2mm}


\ifCLASSOPTIONcaptionsoff
  \newpage
\fi

\bibliographystyle{IEEEtran}
\bibliography{refs_v2}


\clearpage
\appendices  

\section{Result Summary}
We list the benchmark results of label-efficient image segmentation methods in this section, including methods for semantic segmentation with image-level supervision (Table~\ref{tab:Table_WSSS_performance}), instance segmentation with image-level supervision (Table~\ref{tab:sota_imagelevel_instanceseg}), semantic segmentation with box-level supervision (Table~\ref{tab:Table_WSSS_box_performance}), instance segmentation with box-supervision (Table~\ref{tab:sota_boxlevel_instanceseg}), semantic segmentation with scribble-level supervision (Table~\ref{tab:Table_WSSS_scribble_performance}), semi-supervised semantic segmentation (Table~\ref{tab:semi_res}), semi-supervised (partially-supervised) instance segmentation (Table~\ref{tab:pis_res}) and domain-adaptive semantic segmentation (Table~\ref{tab:uda_res}). 

\begin{table}[h]
\caption{Results (mIoU) of semantic segmentation methods with image-level supervision on PASCAL VOC~\cite{pascal} \textit{val} and \textit{test}. $^\dag$: using saliency maps. }
\centering
\small
\resizebox{1.0\linewidth}{!}{
\begin{tabular*}{9.5cm}{@{\extracolsep{\fill}}lccc}
\toprule
Method            & Backbone           & Val            & Test           \\ \midrule
  Wang \emph{et al.} (CVPR18) \cite{WSSS_MCOF_2018_CVPR}$^\dag$ &  ResNet101 &60.3& 61.2 \\
  SeeNet (NeurIPS18) \cite{WSSS_SE_2018_NEURIPS}$^\dag$ & ResNet38 &63.1 & 62.8 \\
 Sun \emph{et al.} (ECCV20) \cite{WSSS_MCIS_2020_ECCV}$^\dag$   &          ResNet101        &        66.2          &      66.9        \\
 Li \emph{et al.} (AAAI21) \cite{WSSS_Group_2021_AAAI}$^\dag$   &  ResNet101 &    68.2       &    68.5           \\
 Yao \emph{et al.} (CVPR21) \cite{WSSS_NSROM_2021_CVPR}$^\dag$ & ResNet101  &68.3&68.5 \\
 Xu \emph{et al.} (ICCV21) \cite{WSSS_LAT_2021_ICCV}$^\dag$ & ResNet38  &69.0&68.6 \\ 
 Kim \emph{et al.} (AAAI21) \cite{WSSS_DRS_2021_AAAI}$^\dag$ & ResNet101  &70.4 & 70.7 \\
 Jiang \emph{et al.} (CVPR22) \cite{WSSS_L2G_2022_CVPR}$^\dag$ & ResNet101  &72.1&71.7 \\ 
 Zhou \emph{et al.} (CVPR22) \cite{WSSS_RSC_2022_CVPR}$^\dag$ & ResNet101  &72.2&72.8 \\ \midrule
 Kolesnikov \emph{et al.} (ECCV16) \cite{WSSS_SEC_2016_ECCV} & VGG16 &50.7 & 51.7 \\
 Wei \emph{et al.} (CVPR17) \cite{WSSS_AE_2017_CVPR} &  VGG16 &55.0& 55.7\\
   Wei \emph{et al.} (CVPR18) \cite{WSSS_RDC_2018_CVPR} & VGG16 &60.4 & 60.8 \\
   Huang \emph{et al.} (CVPR18) \cite{WSSS_DSRG_2018_CVPR} & ResNet101 &61.4 & 63.2 \\
  Ahn \emph{et al.} (CVPR18) \cite{WSSS_PSA_2018_CVPR} & ResNet38 &61.7 & 63.7 \\
 RRM  (AAAI20) \cite{WSSS_RRM_2020_AAAI} & ResNet38 &62.6 & 62.9 \\
 Araslanov \emph{et al.} (CVPR20) \cite{WSSS_single_stage_2020_CVPR}    &          ResNet38         &      62.7        &   64.3       \\
  Fan \emph{et al.} (CVPR2020) \cite{WSSS_ICD_2020_CVPR}   &          ResNet101        &        64.1          &      64.3       \\
  Zhang \emph{et al.} (ACMMM21) \cite{WSSS_AALR_2021_MM} &  ResNet38 &63.9& 64.8 \\
  CIAN (AAAI20)  \cite{WSSS_CIAN_2020_AAAI}    &         ResNet101     &  64.3         &  65.3 \\
  Lee \emph{et al.} (CVPR18) \cite{WSSS_USI_2019_CVPR} & ResNet101 &64.9 & 65.3 \\
  Wang \emph{et al.} (IJCV20) \cite{WSSS_IAL_2020_IJCV} & ResNet38 &64.3 & 65.4 \\
  OAA (ICCV2019) \cite{WSSS_OAA_2019_ICCV} & ResNet38 &63.9 & 65.6 \\
  SEAM (CVPR20) \cite{WSSS_SEAM_2020_CVPR} & ResNet38 & 64.5 & 65.7 \\

   Ru \emph{et al.} (CVPR22) \cite{WSSS_afa_2022_CVPR} & MiT-B1 &66.0 & 66.3 \\
Chang \emph{et al.} (CVPR20) \cite{WSSS_SCE_2020_CVPR}  &            ResNet101         &      66.1        &    65.9\\ 
 CONTA (NeurIPS20) \cite{WSSS_CONTA_2020_NEURIPS}   &  ResNet38 &    66.1  &    66.7            \\ 
  Su \emph{et al.} (ICCV21) \cite{WSSS_CDA_2021_ICCV} & ResNet38  &66.1&66.8 \\
  ECS-Net (ICCV21) \cite{WSSS_ECSNet_2021_ICCV} & ResNet38  &66.6 & 67.6 \\
 Lee \emph{et al.} (CVPR21) \cite{WSSS_Advcam_2021_CVPR} & ResNet101  &68.1 &68.0 \\
 Zhang \emph{et al.} (ICCV21) \cite{WSSS_CP_2021_ICCV} & ResNet38  & 67.8&68.5 \\ 
 Wang \emph{et al.} (NeurIPS21) \cite{WSSS_BottleNeck_2021_NEURIPS} & ResNet38&68.3 & 68.6 \\
Xie \emph{et al.} (CVPR22) \cite{ WSSS_CLIMS_2022_CVPR} & ResNet50&70.4 & 70.0 \\
 Xu \emph{et al.} (CVPR22) \cite{WSSS_MCT_2022_CVPR} & ResNet38&71.9 & 71.6 \\
  \bottomrule
\end{tabular*}
}
\label{tab:Table_WSSS_performance}
\end{table}

\begin{table}[h]
\caption{Results of instance segmentation methods with image-level supervision on PASCAL VOC~\cite{pascal}. }
\centering
\small
\resizebox{1.0\linewidth}{!}{
\begin{tabular}{lcccc}
\toprule
Method            & Backbone           & mAP$_{25}$    & mAP$_{50}$        & mAP$_{75}$           \\ \midrule
  PRM (CVPR18) \cite{WSIS_PRM_CVPR_2018} & ResNet50 &44.3 & 26.8 & 9.0 \\
 IAM (CVPR19) \cite{WSIS_IAM_CVPR_2019}   & ResNet50        &        45.9     & 28.8     &      11.9        \\
 WISE (BMVC19) \cite{WSIS_Wheremask_BMVC_2019}   &  ResNet50 &    49.2       &   41.7 & 23.7           \\
 IRNet (CVPR19) \cite{WSIS_IRNet_CVPR_2019} & ResNet50  &-&46.7&- \\
 Label-PEnet (ICCV19) \cite{WSIS_Labelpenet_ICCV_2019} & VGG16  &49.1&30.2&12.9 \\ 
 WSIS-CL (WACV21) \cite{WSIS_CL_WACV_2021} & ResNet50  &57.0&35.7&5.8 \\ 
 PDSL (ICCV21) \cite{WSIS_PDSL_ICCV_2021} & ResNet50-WS  &59.3 & 49.6 &12.7 \\
  \bottomrule
\end{tabular}
}
\label{tab:sota_imagelevel_instanceseg}
\end{table}

\begin{table}[h]
\caption{Results of semantic segmentation methods with box-level supervision in terms of mIoU on the PASCAL VOC 2012~\cite{pascal} \textit{val} and \textit{test} sets.} 
\centering
\small
\resizebox{1.0\linewidth}{!}{
\begin{tabular*}{9.5cm}{@{\extracolsep{\fill}}lccc}
\toprule
Method            & Backbone          & Val            & Test           \\ \midrule
 Dai \emph{et al.} (ICCV15) \cite{WSSS_BoxSup_2015_ICCV} & VGG16 & 62.0&64.6 \\
 Song \emph{et al.} (CVPR19) \cite{WSSS_BDC_2019_CVPR} & ResNet101  & 70.2& - \\
 Kulharia \emph{et al.} (ECCV20) \cite{WSSS_Box2Seg_2020_ECCV} & ResNet101  & 76.4& - \\
 Oh \emph{et al.} (CVPR21) \cite{WSSS_BAP_2021_CVPR} & ResNet101  & 74.6&76.1 \\
  \bottomrule
\end{tabular*}
}
\label{tab:Table_WSSS_box_performance}
\end{table}

\begin{table}[h]
\caption{Results of instance segmentation methods with box-level supervision on COCO~\cite{COCO}. }
\centering
\small
\resizebox{1.0\linewidth}{!}{
\begin{tabular*}{9.5cm}{@{\extracolsep{\fill}}lcccc}
\toprule
Method            & Backbone           & mAP    & AP$_{50}$        & AP$_{75}$           \\ \midrule
 COCO \emph{val}:  & & & \\
 BBTP (NeurIPS19) \cite{WSIS_BBTP_NIPS_2019} & ResNet101 &21.1 & 45.5 & 17.2 \\
 BBAM (CVPR21) \cite{WSIS_BBAM_CVPR_2021}   & ResNet101        &        26.0     & 50.0     &      23.9        \\
 Boxinst (CVPR21) \cite{WSIS_Boxinst_CVPR_2021}   &  ResNet101 &    31.6       &   54.0 & 31.9           \\
 \midrule
 COCO \emph{test-dev}:  & & & \\
 BBAM (CVPR21) \cite{WSIS_BBAM_CVPR_2021}   & ResNet101        &        25.7     & 50.0     &      23.3        \\
 Boxinst (CVPR21) \cite{WSIS_Boxinst_CVPR_2021}   &  ResNet101 &    32.5       &   55.3 & 33.0           \\
 BoxCaSeg (CVPR21) \cite{WSIS_BoxCaSeg_CVPR_2021} & ResNet101  &30.9&54.3&30.8 \\
  \bottomrule
\end{tabular*}
}
\label{tab:sota_boxlevel_instanceseg}
\end{table}

\begin{table}[h]
\caption{Results (mIoU) of semantic segmentation methods with scribble-level supervision on PASCAL VOC~\cite{pascal} \textit{val}. }
\centering
\small
\resizebox{1.0\linewidth}{!}{

\begin{tabular*}{9.5cm}{@{\extracolsep{\fill}}lcc}
\toprule
Method            & Backbone            & Val            \\ \midrule
 Paul \emph{et al.} (CVPR17) \cite{WSSS_RW_2017_CVPR} & ResNet101  & 61.4 \\
 Di \emph{et al.} (CVPR16) \cite{WSSS_ScribbleSup_2016_CVPR} & VGG16  & 63.1 \\
 Tang \emph{et al.} (CVPR18) \cite{WSSS_ORL_2018_ECCV} & ResNet101  & 72.8 \\
 Tang \emph{et al.} (ECCV18) \cite{WSSS_NC_2018_CVPR} & ResNet101  & 73.0\\
 Xu \emph{et al.} (ICCV21) \cite{WSSS_SSI_2021_ICCV} & ResNet101  & 74.9 \\

  \bottomrule
\end{tabular*}
}
\label{tab:Table_WSSS_scribble_performance}
\end{table}

\begin{table}[h]
\caption{Results (mIoU) of semi-supervised semantic segmentation methods using labeled training data of different proportions ($1/2,1/4,1/8,1/16$) on PASCAL VOC \textit{val}. }
\centering
\small
\resizebox{1.0\linewidth}{!}{
\begin{tabular*}{9.5cm}{@{\extracolsep{\fill}}lccccc}

\toprule
 Method & 1/2 & 1/4 & 1/8 & 1/16 \\ 
 \midrule
  
  AdvSemSeg (BMVC18)~\cite{SSL_Hung2018} & 65.3 & 60.0 & 47.6 & 39.7 \\
  MT (NeurIPS17)~\cite{SSL_MT} & 69.2 & 63.0 & 55.8 & 48.7 \\
  GCT (ECCV20)~\cite{SSL_GCT} & 70.7 & 64.7 & 55.0 & 46.0 \\
  VAT (CoRR17)~\cite{SSL_VAT} & 63.3 & 56.9 & 49.4 & 36.9 \\
  CutMix-Seg (BMVC20)\cite{SSL_CutMix-Seg} & 69.8 & 68.4 & 63.2 & 55.6 \\
  PseudoSeg (ICLR21)~\cite{SSL_PseudoSeg} & 72.4 & 69.1 & 65.5 & 57.6 \\
  CPS (CVPR21)~\cite{SSL_CPS} & 75.9 & 71.7 & 67.4 & 64.0 \\
  PC$^{2}$Seg (ICCV21)~\cite{SSL_PCCSeg} & 73.1 & 69.8 & 66.3 & 57.0 \\

  \bottomrule
\end{tabular*}
}
\label{tab:semi_res}
\end{table}

\begin{table*}[h]
    \newcommand{\CC}[1]{\cellcolor{gray!#1}}
    \centering
    \caption{
        ``\emph{nonvoc}$\to$\emph{voc}'' denotes that categories in \emph{nonvoc} and \emph{voc} are the \emph{base} and \emph{novel} categories, respectively, and vice versa. \emph{voc} contains 20 classes existed in both COCO and VOC. \emph{nonvoc} contains 60 classes existed only in COCO, not in VOC. $1 \times$ represents for 12 epochs and \emph{130k} is a customized schedule only used in OPMask~\cite{PSIS_Opmask_ICCV_2021}. ``Layers'' indicates the number of Conv blocks adopted in the mask head to perform mask prediction. Generally, a heavier mask head leads to better performance, which has been demonstrated in~\cite{PSIS_DeepMAC_ICCV_2021}.
    }
    \renewcommand{\arraystretch}{1.0}
    \resizebox{\linewidth}{!}{
    \begin{tabular}{llcc|cccccc|cccccc}
    \toprule
    & & & & \multicolumn{6}{c}{\emph{nonvoc}$\rightarrow$\emph{voc}} & \multicolumn{6}{c}{\emph{voc}$\rightarrow$\emph{nonvoc}} \\
    \makecell[c]{Method} & Backbone & Schedule & Layers & mAP & AP$_{50}$ & AP$_{75}$ & AP$_S$ & AP$_M$ & AP$_L$ & mAP & AP$_{50}$ & AP$_{75}$ & AP$_S$ & AP$_M$ & AP$_L$ \\
    \midrule

    Mask$^X$ R-CNN~\cite{PSIS_MaskXRCNN_CVPR_2018} & ResNet50& $1\times$ & 4           & 28.9 & 52.2 & 28.6 & 12.1 & 29.0 & 40.6 & 23.7 & 43.1 & 23.5 & 12.4 & 27.6 & 32.9 \\
    Mask GrabCut~\cite{YYDS_Grabcut_TOG_2004} &ResNet50 & $1\times$ & -               & 19.5 & 46.2 & 14.2 & 4.7  & 15.9 & 32.0 & 19.5 & 39.2 & 17.0 & 6.5  & 20.9 & 34.3 \\
    CPMask~\cite{PSIS_CPMask_ECCV_2020} &  ResNet50 &$1\times$ & 4                      & - & - & - & - & - & - & 28.8 & 46.1 & 30.6 & 12.4 & 33.1 & 43.4 \\
    ShapeProp~\cite{PSIS_Shapeprop_CVPR_2020} &ResNet50 & $1\times$ & 4                 & 34.4 & 59.6 & 35.2 & 13.5 & 32.9 & 48.6 & 30.4 & 51.2 & 31.8 & 14.3 & 34.2 & 44.7 \\
    ContrastMask~\cite{PSIS_ContrastMask_CVPR_2022} & ResNet50 &$1\times$ & 4                         &35.1 & 60.8 & 35.7 & 17.2 & 34.7 & 47.7 &30.9 & 50.3 & 32.9 & 15.2 & 34.6 & 44.3 \\
    \hline
    OPMask~\cite{PSIS_Opmask_ICCV_2021}      &   ResNet50    & $130k$ &  7       & 36.5 & 62.5 & 37.4 & 17.3 & 34.8 & 49.8 & 31.9 & 52.2 & 33.7 & 16.3 & 35.2 & 46.5 \\
    ContrastMask~\cite{PSIS_ContrastMask_CVPR_2022} & ResNet50 &$3\times$ & 4                         & 37.0 & 63.0 & 38.6 & 18.3& 36.4 & 50.2 & 32.9 & 52.5 & 35.4 & 16.6 & 37.1 & 47.3 \\

    \hline

    Mask GrabCut~\cite{YYDS_Grabcut_TOG_2004} &ResNet101& $1\times$ & -               & 19.6 & 46.1 & 14.3 & 5.1  & 16.0 & 32.4 & 19.7 & 39.7 & 17.0 & 6.4  & 21.2 & 35.8 \\
    Mask$^X$ R-CNN~\cite{PSIS_MaskXRCNN_CVPR_2018} & ResNet101&$1\times$ & 4           & 29.5 & 52.4 & 29.7 & 13.4 & 30.2 & 41.0 & 23.8 & 42.9 & 23.5 & 12.7 & 28.1 & 33.5 \\
    ShapeMask~\cite{PSIS_Shapemask_ICCV_2019}   & ResNet101&$1\times$ & 8              & 33.3 & 56.9 & 34.3 & 17.1 & 38.1 & 45.4 & 30.2 & 49.3 & 31.5 & 16.1 & 38.2 & 28.4 \\
    ShapeProp~\cite{PSIS_Shapeprop_CVPR_2020} &ResNet101 &$1\times$ & 4                 & 35.5 & 60.5 & 36.7 & 15.6 & 33.8 & 50.3 & 31.9 & 52.1 & 33.7 & 14.2 & 35.9 & 46.5 \\
    ContrastMask~\cite{PSIS_ContrastMask_CVPR_2022} & ResNet101&$1\times$ & 4                         & 36.6 & 62.2 & 37.7 & 17.5 & 36.5 & 50.1 & 32.4 & 52.1 & 34.8 & 15.2 & 36.7 & 47.3 \\
    \hline
    ShapeMask*~\cite{PSIS_Shapemask_ICCV_2019}  & ResNet101 &$3\times$ & 8     & 35.7 & 60.3 & 36.6 & 18.3 & 40.5 & 47.3 & 33.2 & 53.1 & 35.0 & 18.3 & 40.2 & 43.3 \\
    CPMask~\cite{PSIS_CPMask_ECCV_2020}  & ResNet101 &$3\times$ & 4                      & 36.8 & 60.5 & 38.6 & 17.6 & 37.1 & 51.5 & 34.0 & 53.7 & 36.5 & 18.5 & 38.9 & 47.4 \\
    OPMask~\cite{PSIS_Opmask_ICCV_2021}     &   ResNet101     & $130k$    &  7       & 37.1 & 62.5 & 38.4 & 16.9 & 36.0 & 50.5 & 33.2 & 53.5 & 35.2 & 17.2 & 37.1 & 46.9 \\
    ContrastMask~\cite{PSIS_ContrastMask_CVPR_2022} &ResNet101 &$3\times$ & 4                         &38.4 & 64.5 & 39.8 & 18.4 & 38.1 & 52.6 & 34.3 & 54.7 & 36.6 & 17.5 & 38.4 & 50.0\\
    deep-MAC~\cite{PSIS_DeepMAC_ICCV_2021} &SpineNet143 &$3\times$ & 52                         &41.0 & 68.2 & 43.1 & 22.0 & 40.0 & 55.9 & 38.7 & 62.5 & 41.0 & 22.3 & 43.0 & 55.9\\
    \bottomrule
    \end{tabular}
    }

    \label{tab:pis_res}
\end{table*}

\begin{table*}[t]
\caption{Results of domain adaptive semantic segmentation methods in terms of mIoU on GTA5~\cite{GTA5_2016_ECCV} (source) $\rightarrow$ Cityscapes~\cite{Cityscapes_2016_CVPR} (target). }
\centering
\small
\resizebox{1.0\linewidth}{!}{
\begin{tabular}{lccccccccccccccccccc|c}

\toprule
 & Road & S.walk & Build. & Wall & Fence & Pole & Tr.Light & Sign & Veget. & Terrain & Sky & Person & Rider & Car & Truck & Bus & Train & M.bike & Bike & mIoU          \\ \midrule
  
  I2IAdapt (CVPR18)~\cite{UDA_I2IAdapt} & 85.8 & 37.5 & 80.2 & 23.3 & 16.1 & 23.0 & 14.5 & 9.8 & 79.2 & 36.5 & 76.4 & 53.4 & 7.4 & 82.8 & 19.1 & 15.7 & 2.8 & 13.4 & 1.7 & 35.7 \\
  
  AdaptSeg (CVPR18)~\cite{UDA_AdaptSeg} & 86.5 & 36.0 & 79.9 & 23.4 & 23.3 & 23.9 & 35.2 & 14.8 & 83.4 & 33.3 & 75.6 & 58.5 & 27.6 & 73.7 & 32.5 & 35.4 & 3.9 & 30.1 & 28.1 & 42.4 \\
  
  CBST (ECCV18)~\cite{UDA_CBST} & 91.8 &   53.5 & 80.5 & 32.7 & 21.0 & 34.0 & 28.9 & 20.4 & 83.9 & 34.2 & 80.9 & 53.1 & 24.0 & 82.7 & 30.3 & 35.9 & 16.0 & 25.9 & 42.8 & 45.9 \\
 
  BDL (CVPR19)~\cite{UDA_BDL} & 91.0 &    44.7 & 84.2 & 34.6 & 27.6 & 30.2 & 36.0 & 36.0 & 85.0 & 43.6 & 83.0 & 58.6 & 31.6 & 83.3 & 35.3 & 49.7 & 3.3 & 28.8 & 35.6 & 48.5 \\
  
  Kim \emph{et al.} (CVPR20)~\cite{UDA_Kim} & 92.9 & 55.0 & 85.3 & 34.2 & 31.1 & 34.9 & 40.7 & 34.0 & 85.2 & 40.1 & 87.1 & 61.0 & 31.1 & 82.5 & 32.3 & 42.9 & 0.3 & 36.4 & 46.1 & 50.2 \\
  
  TPLD (ECCV20)~\cite{UDA_TPLD} & 94.2 & 60.5 & 82.8 & 36.6 & 16.6 & 39.3 & 29.0 & 25.5 & 85.6 & 44.9 & 84.4 & 60.6 & 27.4 & 84.1 & 37.0 & 47.0 & 31.2 & 36.1 & 50.3 & 51.2 \\
  
  DACS (WACV21)~\cite{UDA_DACS} & 89.9 &  39.7 & 87.9 & 30.7 & 39.5 & 38.5 & 46.4 & 52.8 & 88.0 & 44.0 & 88.8 & 67.2 & 35.8 & 84.5 & 45.7 & 50.2 & 0.0 & 27.3 & 34.0 & 52.1 \\
 
  ProDA (CVPR21)~\cite{UDA_ProDA} & 87.8 & 56.0 & 79.7 & 46.3 & 44.8 & 45.6 & 53.5 & 53.5 & 88.6 & 45.2 & 82.1 & 70.7 & 39.2 & 88.8 & 45.5 & 59.4 & 1.0 & 48.9 & 56.4 & 57.5 \\
  
  DPL (ICCV21)~\cite{UDA_DPL} & 92.8 &    54.4 & 86.2 & 41.6 & 32.7 & 36.4 & 49.0 & 34.0 & 85.8 & 41.3 & 86.0 & 63.2 & 34.2 & 87.2 & 39.3 & 44.5 & 18.7 & 42.6 & 43.1 & 53.3 \\
  
  BAPA-Net (ICCV21)~\cite{UDA_BAPANet} & 94.4 & 61.0 & 88.0 & 26.8 & 39.9 & 38.3 & 46.1 & 55.3 & 87.8 & 46.1 & 89.4 & 68.8 & 40.0 & 90.2 & 60.4 & 59.0 & 0.00 & 45.1 & 54.2 & 57.4 \\
  
  DAFormer (CVPR22)~\cite{UDA_DAFormer} & 95.7 & 70.2 & 89.4 & 53.5 & 48.1 & 49.6 & 55.8 & 59.4 & 89.9 & 47.9 & 92.5 & 72.2 & 44.7 & 92.3 & 74.5 & 78.2 & 65.1 & 55.9 & 61.8 & 68.3 \\

  \bottomrule
\end{tabular}
}
\label{tab:uda_res}
\end{table*}

\end{document}